\documentclass{article}
\usepackage{iclr2027_conference,times}

\usepackage{amsmath,amsfonts,bm}

\def\eqref#1{equation~\ref{#1}}

\def\1{\bm{1}}

\DeclareMathAlphabet{\mathsfit}{\encodingdefault}{\sfdefault}{m}{sl}
\SetMathAlphabet{\mathsfit}{bold}{\encodingdefault}{\sfdefault}{bx}{n}

\usepackage{hyperref}
\usepackage{url}
\usepackage{booktabs}
\usepackage{multirow}
\usepackage{tabularx}
\usepackage{array}
\usepackage{graphicx}
\usepackage{subcaption}
\usepackage{bbm}
\usepackage{enumitem}
\usepackage{amssymb}
\usepackage{amsthm}
\usepackage{comment}
\usepackage{pifont}
\usepackage{wrapfig}
\usepackage{longtable}
\usepackage{ragged2e}
\usepackage{xcolor}
\usepackage[percent]{overpic}

\newcommand{\cmark}{\ding{51}}
\newcommand{\xmark}{\ding{55}}

\definecolor{myorange}{RGB}{222,159,131}
\definecolor{myblue}{RGB}{116,148,188}

\title{ConflictGuide: AutoResearch Improves When Competing Behaviors Are Made Visible}

\author{
Binqian Xu\textsuperscript{1},
Qiran Zou\textsuperscript{1},
Xiangbo Shu\textsuperscript{2},
Dianbo Liu\textsuperscript{1}
\\
\textsuperscript{1}National University of Singapore
\\
\textsuperscript{2}Nanjing University of Science and Technology
}

\iclrfinalcopy

\begin{document}

\maketitle

\lhead{}

\begin{abstract}
    When designing machine learning models, desirable properties are often in tension: improving one behavior can impair another, so task progress can depend on alleviating the conflict. LLM-based AutoResearch systems, which iteratively edit model code and retain edits based on scalar task-performance feedback, have largely ignored this trade-off. We find that scalar feedback supports broad exploration early in search, but it does not reveal how edits affect competing behaviors. In matched-budget experiments, introducing competing-behavior feedback as task gains diminish increases the share of proposals that improve both behaviors and sustains progress beyond scalar-only plateaus. Obtaining this feedback for a given model requires identifying its competing behaviors and designing probes to measure them. To make competing-behavior feedback actionable, we introduce \textbf{ConflictGuide}. Its reusable ConflictGuide-Skill combines a literature-grounded taxonomy with model-specific evidence to identify competing behaviors and specify probes for a code agent to implement as metrics. Evolution proceeds in two stages: Stage I explores with task feedback; Stage II uses probe feedback to steer proposals toward conflict alleviation and retains marginal-gain edits only when probes indicate sufficient alleviation. Across five diverse model families, ConflictGuide reduces task and conflict-related errors by up to 28\% and 14\%, respectively, relative to scalar-only AutoResearch, with gains extending to other code agents.
\end{abstract}

\section{Introduction}



Machine-learning models are often expected to exhibit more than one desirable behavior~\citep{sener2018multi}. Balancing these behaviors has long been a challenge: improving one can impair another~\citep{tsipras2019robustness}. In the 1990s, soft-margin support vector machines balanced margin width against the cost of margin violations~\citep{cortes1995support}. GANs face a different challenge: generating convincing samples without collapsing to a few modes~\citep{salimans2016improved}. Related challenges arise in adversarial classifiers, where gains in standard accuracy can reduce robustness~\citep{tsipras2019robustness}, and in uncertainty-aware classifiers, which must predict accurately on familiar inputs while expressing greater uncertainty as inputs move farther from the training data~\citep{liu2020simple}. The sources of tension vary across models: limited capacity, competing demands on shared representations, properties of the data, and training choices can all contribute~\citep{sener2018multi}. The resulting trade-offs need not be strictly zero-sum: advances in model design or training can improve the attainable balance~\citep{addepalli2022efficient}.

\begin{figure}[!t]
    \centering
    \includegraphics[width=\linewidth]{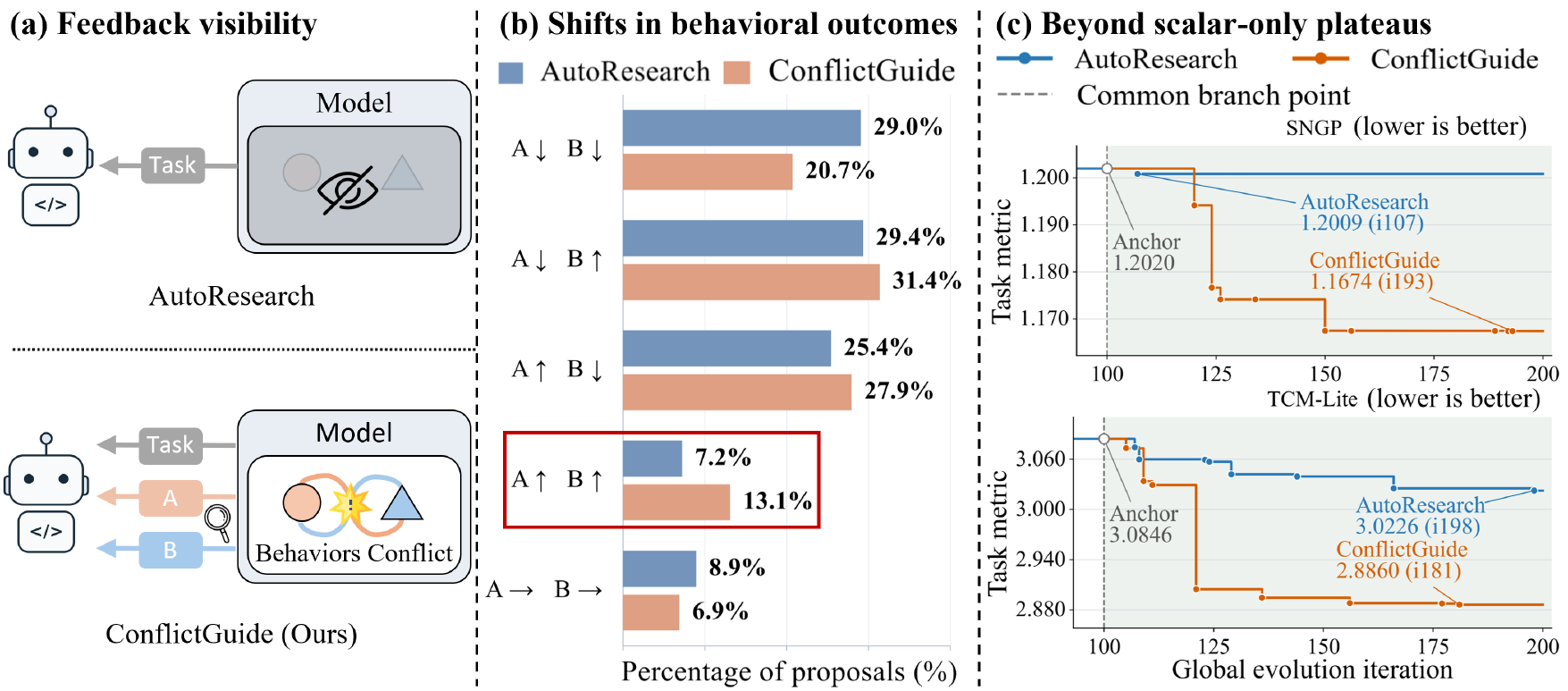}
    \caption{Comparison of AutoResearch (scalar-only feedback) and ConflictGuide (competing-behavior feedback).
(a) Feedback visible to the code agent.
(b) Proposal-level behavioral outcomes across models. ConflictGuide increases joint improvements from 7.2\% to 13.1\% and reduces joint degradations from 29.0\% to 20.7\%.
(c) From common branch points under matched continuation budgets, ConflictGuide continues to lower the task metrics, whereas AutoResearch plateaus.}
    \label{fig:conflict_feedback}
    \vspace{-20pt}
\end{figure}



Recent AutoResearch systems automate model development through iterative experimentation. In a typical loop, an LLM coding agent edits training code, evaluates each revision under a fixed training-time budget, and retains edits primarily based on a scalar task metric~\citep{karpathy2026autoresearch}. Some systems augment this loop with task-specific diagnostic feedback. Self-EvolveRec uses simulated-user critiques and model diagnostics to identify recommendation failures~\citep{kim2026self}, while EvoPINN analyzes training trajectories to guide changes to neural representations and training procedures~\citep{yin2026evopinn}. These systems use failure diagnosis and training-dynamics analysis, respectively, to identify task-specific performance bottlenecks and guide subsequent revisions. From a model-design perspective, however, unresolved tensions between desirable behaviors may limit performance, yet scalar task metrics do not reveal how code edits affect each behavior. Competing-behavior feedback can therefore complement these diagnostics by exposing such effects. To our knowledge, we provide the first explicit exploration of this feedback in AutoResearch.

How competing-behavior feedback shapes search may, however, depend on when it is introduced. We therefore conducted two matched-budget exploratory studies within a standard scalar-guided AutoResearch loop. On a spectral neural operator~\citep{qin2024toward}, feedback provided from the outset produced conflict-specific edits but concentrated the search on spectral compensation, whereas scalar-only feedback yielded a broader mix of representational and optimization changes (Appendix~\ref{app:exploratory}). We then tested delayed feedback from common scalar-only branch points reached as task gains diminished. The share of proposals improving both behaviors rose from 7.2\% to 13.1\%, while the share degrading both behaviors fell from 29.0\% to 20.7\% (Figure~\ref{fig:conflict_feedback}(b)). Delayed-feedback searches also continued to improve task performance, whereas matched scalar-only continuations made little further progress (Figure~\ref{fig:conflict_feedback}(c)). Together, these observations motivate a two-stage design in which scalar feedback first supports broad task-oriented exploration, after which competing-behavior feedback focuses subsequent refinement on unresolved conflicts.

\begin{figure}[t]
    \centering
    \includegraphics[width=0.9\linewidth]{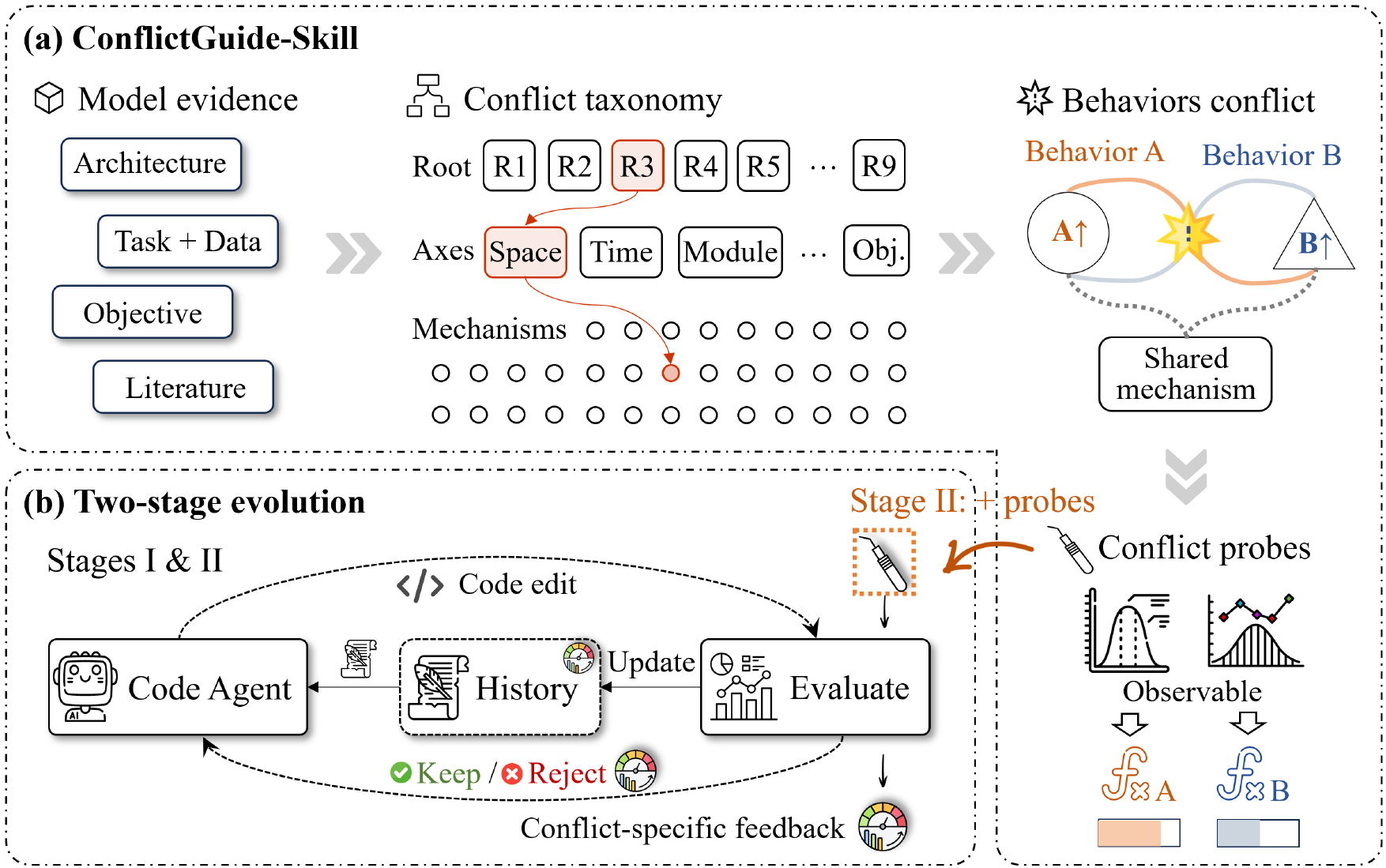}
    \caption{Overview of ConflictGuide.
(a) ConflictGuide-Skill uses a literature-grounded taxonomy to identify a model-specific conflict and operationalize its competing behaviors through qualified probes.
(b) Both stages share the evolution loop: Stage I uses scalar feedback for exploration; once task-metric gains diminish, Stage II adds conflict-specific feedback to guide proposals and retention.}
    \label{fig:overview}
    \vspace{-10pt}
\end{figure}

To obtain competing-behavior feedback across models, we construct a literature-grounded conflict taxonomy. It organizes conflicts by root cause, competing axis, and mechanism, with nine roots, six axes, and 110 mechanisms spanning model design, data, training, and evaluation (Appendix~\ref{app:taxonomy}). Built on this taxonomy, the reusable \textbf{ConflictGuide-Skill} uses evidence from a target model's architecture, task and data, training objective, and evaluation setting to identify candidate conflicts. For each supported conflict, it specifies a compact, non-redundant probe set that a code agent implements as metrics (Figure~\ref{fig:overview}(a)). Before evolution, probes are experimentally qualified, with supporting evidence and results available for inspection. Guided by the exploratory findings, \textbf{ConflictGuide} begins with scalar task feedback for broad exploration (Stage I). As task gains diminish, Stage II makes probe results visible to the code agent while recording them alongside code edits in the search history (Figure~\ref{fig:overview}(b)). These results reveal the conflict-related effects of code edits and guide subsequent proposals toward conflict alleviation. They also inform retention: edits with marginal positive task gains are kept only when the probes indicate sufficient alleviation.

Our main contributions are summarized as follows: 
\textbf{1) Competing-behavior feedback for AutoResearch.} To our knowledge, we are the first to introduce such feedback into AutoResearch to alleviate behavioral conflicts and enable further performance gains.
\textbf{2) ConflictGuide-Skill for conflict identification.} We introduce a reusable agent skill built on a literature-grounded Root--Axis--Mechanism taxonomy. Given a target model, it identifies model-specific conflicts and designs measurable probes for experimental qualification. \textbf{3) ConflictGuide for two-stage evolution.} 
We propose ConflictGuide, which uses scalar task feedback for initial exploration and then introduces skill-derived conflict feedback to guide proposals toward conflict-relevant model components and inform edit retention. 
\textbf{4) Evaluation across models and agents.} Across five different model families, ConflictGuide achieves up to 28\% lower task error and 14\% lower conflict-related error under matched budgets, improves beyond scalar-only search plateaus, and generalizes across code agents.

\section{Related Work}
\vspace{-5pt}


\paragraph{Feedback and Diagnosis in Automated Model Evolution.}

Automated model-evolution systems supplement task-performance feedback with diagnostic and structural signals. LLaMEA-SAGE derives mutation guidance from performance-relevant code structures \citep{van2026llamea}, while NOVA integrates history, verification diagnostics, and trajectory memory into an architecture gradient \citep{liu2026nova}. Other systems evaluate candidates against external criteria, such as machine-checkable physical requirements \citep{abueidda2026physics} or benchmarks for known alignment failures \citep{yueh2026automated}. A more targeted line diagnoses performance limits and intervention targets: Self-EvolveRec translates model weaknesses into directional guidance \citep{kim2026self}, GoalEvolve localizes multi-objective gaps to optimization stages \citep{liu2026goalevolve}, and EvoPINN conditions proposals on training-state diagnostics \citep{yin2026evopinn}. Whereas these methods derive guidance from code structure, verification, domain requirements, or performance bottlenecks, ConflictGuide diagnoses a persistent mechanism-level conflict between competing behaviors. Quantitative Probes track this conflict after each edit, revealing whether the edit alleviates or aggravates it. Introduced after task gains diminish, this conflict-specific evidence guides subsequent proposals and informs retention. Additional related work appears in the Appendix~\ref{app:related_work}.

\section{Method}
\label{sec:method}
\vspace{-5pt}

As shown in Figure~\ref{fig:overview}, ConflictGuide combines ConflictGuide-Skill with a two-stage evolution pipeline. The skill identifies a model-specific conflict and operationalizes it through qualified probes of the competing behaviors. The pipeline uses scalar task feedback for initial exploration, then probe-derived feedback to guide conflict-aware proposals and provide an auxiliary retention path.

\subsection{Problem Setup}
\label{sec:setup}

Let $x_t$ denote the training code retained at iteration $t$.
A coding agent proposes an edit $e_t$, producing the candidate
$x_t^{e_t}=T_{e_t}(x_t)$, which is evaluated under a fixed experimental budget.
Let $S(x)$ be the scalar task metric, which is minimized in all our experiments,
and define its task gain as
\begin{equation}
G_t = S(x_t)-S(x_t^{e_t}),
\end{equation}
such that $G_t>0$ indicates improved task performance.
Standard AutoResearch conditions each proposal on the current code and a task-feedback history,
\begin{equation}
e_t \sim \pi\!\left(\cdot \mid x_t,\mathcal{H}_t^S\right),
\qquad
\mathcal{H}_t^S=\{(e_i,G_i)\}_{i<t}.
\end{equation}
Although $G_t$ determines whether an edit improves overall performance, it does not reveal how that edit affects the competing behaviors underlying the task metric.
We define a model-specific conflict
$c=(B_A,B_B,M_c)$ by two desirable behaviors, $B_A$ and $B_B$, that interact through a model mechanism $M_c$ such that improving one may impair the other.
ConflictGuide-Skill associates the conflict with a set of probes
$\mathbf{Z}_c(x)=[Z_1(x),\ldots,Z_m(x)]$.
Each probe has a pre-specified desirable direction.
Let
\begin{equation}
\mathbf{G}_{c,t}^{Z}
=
[G_{1,t}^{Z},\ldots,G_{m,t}^{Z}]
\end{equation}
denote the behavior-specific probe improvements of the candidate relative to
the current code. A positive component indicates improvement in the
corresponding behavior.

\subsection{ConflictGuide-Skill}
\label{sec:skill}

ConflictGuide-Skill screens Root--Axis--Mechanism paths from a
literature-grounded taxonomy using evidence from the target model's
architecture, task and data, training objective, task metric, and evaluation
setting. It instantiates $c$ only when the evidence supports two desirable
behaviors coupled by a concrete mechanism $M_c$; otherwise, it abstains.

For an identified conflict, the skill specifies a minimal, non-redundant
probe set
\begin{equation}
Z_j(x;\mathcal{E}_j)
=
g_j\!\left(O_j(x;\mathcal{E}_j)\right),
\qquad j=1,\ldots,m,
\end{equation}
where $O_j$ is an observable from model outputs, internal states, or parameters,
and $g_j$ computes the behavior metric.
A code agent implements the probes as read-only metrics.
The skill also defines a probe-based alleviation function $A_c(e_t)$,
evaluated in the current search context using model-specific reference
and guard conditions.
Before evolution, null calibration and paired evaluation reject unsupported
probes and calibrate the threshold $\tau_c$ for substantial alleviation.
Only qualified outputs $(c,\mathbf{Z}_c,A_c,\tau_c)$ are passed to
ConflictGuide; full taxonomy (Appendix~\ref{app:taxonomy}), screening, abstention, and qualification details
are provided in the Appendix~\ref{app:skill_details}.

\subsection{Two-Stage ConflictGuide Evolution}
\label{sec:evolution}

\paragraph{Stage I: scalar-guided exploration.}
ConflictGuide follows standard AutoResearch for $T_1$ iterations.
The agent observes only $\mathcal{H}_t^S$, supporting broad exploration before
conflict-specific feedback is introduced.
The stage boundary $T_1$ is fixed before evolution; we evaluate its sensitivity
in Section~\ref{sec:experiments}.

\paragraph{Stage II: conflict-aware refinement.}
After Stage I, ConflictGuide augments the evolution history with the qualified
probe improvements:
\begin{equation}
\mathcal{H}_t^{S,Z}
=
\mathcal{H}_{T_1}^{S}
\cup
\{(e_i,G_i,\mathbf{G}_{c,i}^{Z})\}_{T_1\leq i<t},
\qquad
e_t
\sim
\pi\!\left(
\cdot\mid x_t,\mathcal{H}_t^{S,Z},c,\mathbf{Z}_c
\right).
\end{equation}
In Stage II, the agent observes each edit’s task gain and probe-derived behavior changes. The prompt specifies the identified conflict and conflict-relevant components available for modification. This information reveals how prior edits affected the conflict and guides proposals within that scope.

Probe feedback additionally informs retention of marginal task gains and ties.
Let $\tau_S\geq0$ separate substantive from marginal task gains, and let
$\mathcal{B}_c\subseteq[0,\tau_S]$ denote the model-specific task-gain band
eligible for probe-based retention.
ConflictGuide retains an edit when
\begin{equation}
\label{eq:retention}
\operatorname{Keep}(e_t)
=
\mathbb{I}\!\left[
    G_t>\tau_S
    \ \lor\
    \left(
        G_t\in\mathcal{B}_c
        \ \land\
        A_c(e_t)>\tau_c
    \right)
\right].
\end{equation}
The first branch retains substantive task improvements; the second requires
qualified probe-based alleviation within $\mathcal{B}_c$.
Negative task gains are never retained.
Model-specific task-gain bands, reference choices (current parent or frozen
Stage-I anchor), and guard conditions are detailed in
Appendix~\ref{app:experimental_details}.
More generally, Appendix~\ref{app:identifiability} formalizes how probe
feedback can improve the identifiability of conflict-alleviating edits
beyond scalar feedback alone.

\begin{table*}[!t]
\centering
\caption{Five experimental model--conflict settings and primary Probe metrics (Appendix~\ref{app:model_probes}).}
\label{tab:conflict-instances}
\footnotesize
\setlength{\tabcolsep}{3.5pt}
\renewcommand{\arraystretch}{1.05}
\renewcommand\tabularxcolumn[1]{m{#1}}   
\begin{tabularx}{\textwidth}{@{}
  m{0.065\textwidth}                     
  m{0.155\textwidth}                 
  m{0.27\textwidth}                    
  >{\raggedright\arraybackslash}X
@{}}
\toprule
\textbf{Model}
  & \textbf{Domain / Task}
  & \textbf{Conflict mechanism (Axis)}
  & \textbf{Primary Probe metric(s)} \\
\midrule

SpecB--FNO
& Scientific ML / PDE operator learning
& \textbf{R2.M2} (\texttt{space})\newline
  Frequency-structure mismatch:
  dominant-mode fit vs.\ non-dominant-mode utilization
& 
$Z_{\mathrm{ND}}(x)=
\dfrac{
\sum_{n,t}\sum_{k\in\mathcal N}
w_k\lvert R^{(x)}_{n,t,k}\rvert^2
}{
\sum_{n,t}\sum_{k\in\mathcal N}
w_k\lvert Y_{n,t,k}\rvert^2
+\epsilon_{\mathrm{spec}}
}\downarrow$
\\
\addlinespace

SNGP
& Uncertainty-aware classification / OOD detection
& \textbf{R3.M4} (\texttt{objective})\newline
  Compression vs.\ uncertainty:
  predictive fit vs.\ distance-aware feature geometry
& 
$\begin{aligned}
Z_{\mathrm{mar}}
&=Q_{q_{\mathrm{mar}}}\!\left[
\widetilde z_i^{\!\top}\widetilde\mu_{y_i}
-\max_{c\neq y_i}
\widetilde z_i^{\!\top}\widetilde\mu_c
\right]\uparrow\\[-1pt]
Z_{\mathrm{dist}}
&=\max_{g\in\{\mathrm{s},\mathrm{c}\}}
Q_{q_{\mathrm{dist}}}\!\left[
\left|r_{ij}-\widetilde r\right|
\right]_{(i,j)\in\mathcal P_g}\downarrow
\end{aligned}$
\\
\addlinespace

ESN
& Reservoir computing / time-series prediction
& \textbf{R2.M4} (\texttt{time})\newline
  Temporal-component mismatch:
  memory retention vs.\ nonlinear processing
& 
$\begin{aligned}
Z_{\mathrm{mem}}
&=
\mathbb E_{q,t,d,j}\!\left[
\min\!\left\{
\sigma_j\!\left(P_{q,t,d}\right),1
\right\}
\right]\uparrow
\\[-1pt]
Z_{\mathrm{nl}}
&=
\mathbb E_q\!\left[
\frac{
\mathbb E_t
\left\lVert K_{q,t}-\overline K_q\right\rVert_F^2
}{
\mathbb E_t
\left\lVert K_{q,t}\right\rVert_F^2
+\epsilon_{\mathrm{nl}}
}
\right]\uparrow
\end{aligned}$
\\
\addlinespace

TCM--Lite
& Learned image compression / rate--distortion coding
& \textbf{R3.M10} (\texttt{space})\newline
  Compact representation damaging fine structure:
  latent rate vs.\ detail preservation
& 
$\begin{aligned}
Z_{\mathrm{rate}}(x)
&=
-N_{\mathrm{pix}}^{-1}
\sum\nolimits_{u\in\widehat{\mathcal U}_x}
\log_2\widetilde p_x(u)
\;\downarrow
\\[-1pt]
Z_{\mathrm{detail}}(x)
&=
\mathbb E_{s}
\left[
\frac{
\lVert\nabla L_s-\nabla\widehat L_{x,s}\rVert_1
}{
\lVert\nabla L_s\rVert_1+
\lVert\nabla\widehat L_{x,s}\rVert_1+
\epsilon
}
\right]
\;\downarrow
\end{aligned}$
\\
\addlinespace
[-1pt]

GCNII
& Graph learning / node classification
& \textbf{R2.M5} (\texttt{space})\newline
  Relational-structure mismatch:
  useful aggregation vs.\ incompatible messages
& 
$Z_{\mathrm{dis}}(x) = \mathbb{E}_k
\left[
\mathcal L_{\mathrm{NLL}}
\!\left(x;A_{\mathrm{dis}}^{(k)},X,\mathcal V\right)
-
S(x)
\right]\downarrow$
\\

\bottomrule
\end{tabularx}
\end{table*}

\section{Experiments}
\label{sec:experiments}

\begin{table*}[!t]
\centering
\setlength{\belowcaptionskip}{7pt}
\caption{SpecB--FNO on 2D incompressible Navier--Stokes
($\nu\!=\!10^{-5}$; $64\!\times\!64$).
NRMSE is the task metric; ND-NMSE the conflict-specific Probe; Params in
millions. Mean $\pm$ s.d.\ over three paired seeds; \textbf{bold} = lowest
error per round; ``None (Original)'' denotes the unmodified reference.}
\label{tab:specb_fno_results}
\small
\setlength{\tabcolsep}{5pt}
\renewcommand{\arraystretch}{1.0}

\begin{tabular}{@{}cllccc@{}}
\toprule
Round
& Method
& Main Cumulative Edit
& NRMSE $\downarrow$
& ND-NMSE $\downarrow$
& Params (M) $\downarrow$ \\
\midrule

\multirow[c]{3}{*}{0}
& Reference
& None (Original)
& $0.0458 \pm 0.0014$
& $0.0861 \pm 0.0086$
& 328.5 \\

& Autoresearch
& Spatial gate
& $0.0382 \pm 0.0015$
& $0.0617 \pm 0.0050$
& 328.9 \\

& ConflictGuide
& Mode refinement
& $\mathbf{0.0344 \pm 0.0003}$
& $\mathbf{0.0530 \pm 0.0007}$
& 333.5 \\
\midrule

\multirow[c]{3}{*}{1}
& Reference
& None (Original)
& $0.0458 \pm 0.0014$
& $0.0861 \pm 0.0086$
& 328.5 \\

& Autoresearch
& Energy match
& $0.0445 \pm 0.0007$
& $0.0766 \pm 0.0035$
& 403.9 \\

& ConflictGuide
& Multi-scale residual
& $\mathbf{0.0442 \pm 0.0014}$
& $\mathbf{0.0760 \pm 0.0064}$
& 407.2 \\
\midrule

\multirow[c]{3}{*}{2}
& Reference
& None (Original)
& $0.0458 \pm 0.0014$
& $0.0861 \pm 0.0086$
& 328.5 \\

& Autoresearch
& Spectral shaping
& $0.0399 \pm 0.0007$
& $0.0698 \pm 0.0039$
& 350.4 \\

& ConflictGuide
& Cross-bypass
& $\mathbf{0.0398 \pm 0.0007}$
& $\mathbf{0.0695 \pm 0.0037}$
& 350.4 \\
\bottomrule
\end{tabular}
\vspace{-5pt}
\end{table*}

\begin{figure}[!t]
    \centering
    \includegraphics[width=\linewidth]{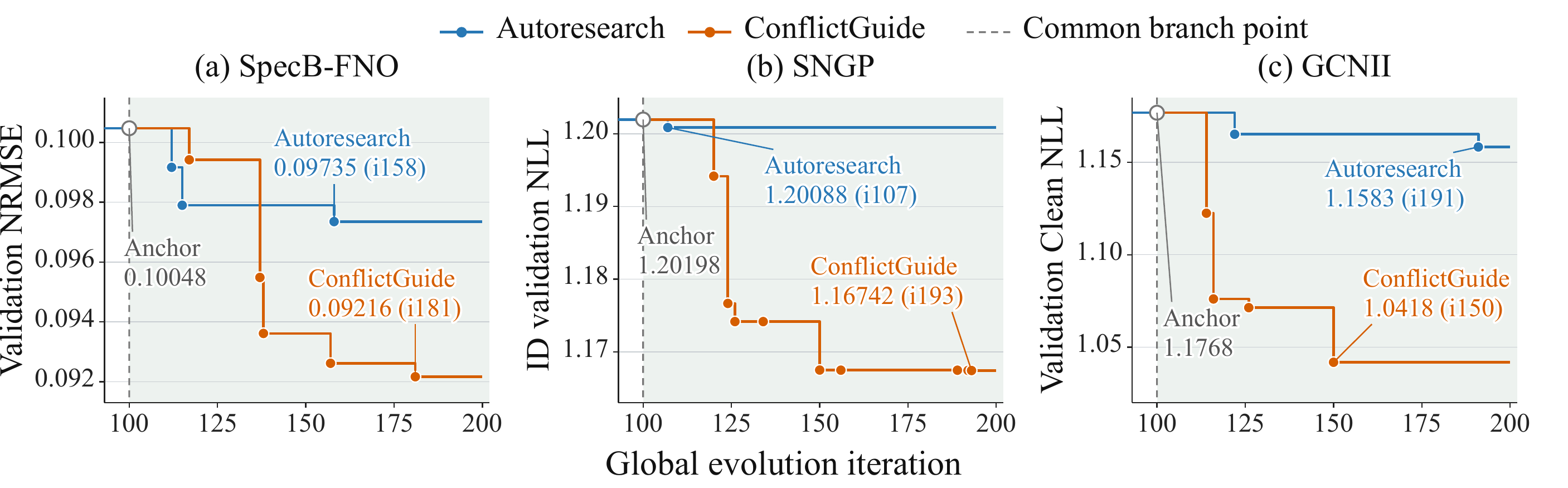}
    \caption{Validation trajectories for one search round on SpecB-FNO, SNGP, and GCNII. Both methods continue from a common branch point with equal budgets; lower is better in all panels.}
    \label{fig:evo_three_models}
\end{figure}

\begin{table}[!t]
\centering
\caption{SNGP on CIFAR-100. Clean ID NLL is the task metric; OOD AUPR (DS) is the corresponding held-out conflict-related outcome.
Mean $\pm$ s.d.\ over three paired seeds; \textbf{bold} = best per round; ``None (Original)'' denotes the
unmodified reference.}
\label{tab:sngp_compact_results}
\small
\setlength{\tabcolsep}{1.3pt}
\renewcommand{\arraystretch}{1.0}
\begin{tabular}{@{}cllccc@{}}
\toprule
\multirow{2}{*}{Round} &
\multirow{2}{*}{Method} &
\multirow{2}{*}{Main Cumulative Edit} &
\multirow{2}{*}{NLL $\downarrow$} &
\multicolumn{2}{c}{OOD AUPR $\uparrow$} \\
\cmidrule(lr){5-6}
& & & & SVHN & CIFAR-10 \\
\midrule

\multirow{3}{*}{0}
& Reference
& None (Original)
& $1.0434 \pm 0.0111$
& $0.7538 \pm 0.0322$
& $0.7304 \pm 0.0029$ \\

& Autoresearch
& Multi-scale RFF
& $1.0578 \pm 0.0004$
& $0.7259 \pm 0.0298$
& $0.7339 \pm 0.0045$ \\

& ConflictGuide
& LN-residual RFF
& $\mathbf{0.8508 \pm 0.0061}$
& $\mathbf{0.7970 \pm 0.0090}$
& $\mathbf{0.7487 \pm 0.0022}$ \\
\midrule

\multirow{3}{*}{1}
& Reference
& None (Original)
& $1.0434 \pm 0.0111$
& $0.7538 \pm 0.0322$
& $0.7304 \pm 0.0029$ \\

& Autoresearch
& Adaptive normalized RFF
& $0.9421 \pm 0.0008$
& $0.7711 \pm 0.0122$
& $0.7455 \pm 0.0024$ \\

& ConflictGuide
& Orthogonal RFF + diff.\ SN
& $\mathbf{0.8794 \pm 0.0092}$
& $\mathbf{0.7729 \pm 0.0167}$
& $\mathbf{0.7497 \pm 0.0056}$ \\
\midrule

\multirow{3}{*}{2}
& Reference
& None (Original)
& $1.0434 \pm 0.0111$
& $0.7538 \pm 0.0322$
& $0.7304 \pm 0.0029$ \\

& Autoresearch
& Fixed orthogonal RFF
& $1.1777 \pm 0.0210$
& $0.7890 \pm 0.0389$
& $0.7181 \pm 0.0049$ \\

& ConflictGuide
& Learnable RFF + mean-field
& $\mathbf{0.8455 \pm 0.0076}$
& $\mathbf{0.8131 \pm 0.0294}$
& $\mathbf{0.7357 \pm 0.0010}$ \\
\bottomrule
\end{tabular}
\end{table}

\begin{table}[!t]
\centering
\caption{GCNII on Wisconsin. Clean accuracy and NLL measure performance; contaminated NLL
measures performance under contamination ($\rho=0.3$). Mean $\pm$ s.d.\ over ten splits, retraining per split,
averaging five frozen contaminations; \textbf{bold} = best per round; ``None (Original)'' = unmodified.}
\label{tab:gcnii_wisconsin_results}
\small
\setlength{\tabcolsep}{1.3pt}
\renewcommand{\arraystretch}{1.0}
\begin{tabular}{@{}cllccc@{}}
\toprule
\multirow{2}{*}{Round} &
\multirow{2}{*}{Method} &
\multirow{2}{*}{Main Cumulative Edit} &
\multirow{2}{*}{Accuracy $\uparrow$} &
\multicolumn{2}{c}{NLL $\downarrow$} \\
\cmidrule(lr){5-6}
& & & & Clean & Contaminated \\
\midrule

\multirow{3}{*}{0}
& Reference
& None (Original)
& $0.7255 \pm 0.0539$
& $0.7838 \pm 0.1681$
& $0.8083 \pm 0.1744$ \\
& Autoresearch
& Disagreement gate
& $0.7255 \pm 0.0555$
& $0.7854 \pm 0.1686$
& $0.8075 \pm 0.1690$ \\
& ConflictGuide
& Residual--neighbor mixer
& $\mathbf{0.7686 \pm 0.0605}$
& $\mathbf{0.7618 \pm 0.2422}$
& $\mathbf{0.7893 \pm 0.2399}$ \\

\midrule
\multirow{3}{*}{1}
& Reference
& None (Original)
& $0.7255 \pm 0.0539$
& $0.7838 \pm 0.1681$
& $0.8083 \pm 0.1744$ \\
& Autoresearch
& Disagreement self-gate
& $0.7922 \pm 0.0648$
& $0.6583 \pm 0.1526$
& $0.6710 \pm 0.1294$ \\
& ConflictGuide
& Depth-aware conflict gate
& $\mathbf{0.7961 \pm 0.0615}$
& $\mathbf{0.6492 \pm 0.1522}$
& $\mathbf{0.6624 \pm 0.1309}$ \\

\midrule
\multirow{3}{*}{2}
& Reference
& None (Original)
& $0.7255 \pm 0.0539$
& $0.7838 \pm 0.1681$
& $0.8083 \pm 0.1744$ \\
& Autoresearch
& Quadratic depth anchor
& $0.8157 \pm 0.0280$
& $0.5669 \pm 0.0835$
& $0.5666 \pm 0.0874$ \\
& ConflictGuide
& Conflict-aware channel scale
& $\mathbf{0.8294 \pm 0.0262}$
& $\mathbf{0.5540 \pm 0.0847}$
& $\mathbf{0.5488 \pm 0.0863}$ \\

\bottomrule
\end{tabular}
\vspace{-7pt}
\end{table}

\begin{table*}[!t]
\centering
\setlength{\belowcaptionskip}{7pt}

\caption{ESN on Mackey--Glass long-horizon forecasting
($\tau,h\!=\!17,84$), testing transfer of probe-guided edits.
NRMSE, MSE, and $R^2$ are task metrics.
Mean $\pm$ s.d.\ over 200 paired runs (20 sequence $\times$ 10 reservoir
seeds); \textbf{bold} = best per round; ``None (Original)'' = unmodified reference.}
\label{tab:esn_mg84_results}

\small
\setlength{\tabcolsep}{2.5pt}
\renewcommand{\arraystretch}{1.0}

\begin{tabular*}{\textwidth}{@{\extracolsep{\fill}}cllccc@{}}
\toprule
Round
& Method
& Main Cumulative Edit
& NRMSE $\downarrow$
& MSE $\downarrow$
& $R^2$ $\uparrow$ \\
\midrule

\multirow[c]{3}{*}{0}
& Reference
& None (Original)
& $0.4261 \pm 0.0417$
& $0.0094 \pm 0.0018$
& $0.8167 \pm 0.0356$ \\

& AutoResearch
& Gated leak
& $0.3173 \pm 0.0262$
& $0.0052 \pm 0.0008$
& $0.8986 \pm 0.0165$ \\

& ConflictGuide
& State--drive coupling
& $\mathbf{0.2285 \pm 0.0308}$
& $\mathbf{0.0027 \pm 0.0007}$
& $\mathbf{0.9468 \pm 0.0138}$ \\

\midrule

\multirow[c]{3}{*}{1}
& Reference
& None (Original)
& $0.4261 \pm 0.0417$
& $0.0094 \pm 0.0018$
& $0.8167 \pm 0.0356$ \\

& AutoResearch
& State--dependent leak
& $0.3246 \pm 0.0264$
& $0.0054 \pm 0.0009$
& $0.8939 \pm 0.0171$ \\

& ConflictGuide
& Dual state coupling
& $\mathbf{0.2561 \pm 0.0424}$
& $\mathbf{0.0034 \pm 0.0011}$
& $\mathbf{0.9326 \pm 0.0207}$ \\

\midrule

\multirow[c]{3}{*}{2}
& Reference
& None (Original)
& $0.4261 \pm 0.0417$
& $0.0094 \pm 0.0018$
& $0.8167 \pm 0.0356$ \\

& AutoResearch
& Adaptive leak gate
& $0.3173 \pm 0.0262$
& $0.0052 \pm 0.0008$
& $0.8986 \pm 0.0165$ \\

& ConflictGuide
& Recurrent--input coupling
& $\mathbf{0.2890 \pm 0.0297}$
& $\mathbf{0.0043 \pm 0.0009}$
& $\mathbf{0.9156 \pm 0.0170}$ \\

\bottomrule
\end{tabular*}
\vspace{-5pt}
\end{table*}

\begin{table*}[!t]
\centering
\setlength{\belowcaptionskip}{7pt}
\caption{TCM--Lite on Kodak at $\lambda\!=\!0.013$. Actual bpp reports
realized rate, MS-SSIM task quality, and $z_{\mathrm{detail}}$ the
conflict-specific detail Probe. Mean $\pm$ s.d.\ over three paired
formal-training seeds; \textbf{bold} = better method per completed round; ``None (Original)'' denotes the
unmodified reference.}
\label{tab:tcm_lite_kodak_results}
\small
\setlength{\tabcolsep}{3pt}
\renewcommand{\arraystretch}{1.0}

\begin{tabular}{@{}cllccc@{}}
\toprule
Round
& Method
& Main Cumulative Edit
& Actual bpp $\downarrow$
& MS-SSIM $\uparrow$
& $z_{\mathrm{detail}}$ $\downarrow$ \\
\midrule

\multirow[c]{3}{*}{0}
& Reference
& None (Original)
& $0.5727 \pm 0.0650$
& $0.8981 \pm 0.0269$
& $0.1918 \pm 0.0295$ \\

& Autoresearch
& Cross-branch gating
& $0.5672 \pm 0.0188$
& $0.9186 \pm 0.0029$
& $0.1744 \pm 0.0076$ \\

& ConflictGuide
& Intra-encoder skip
& $\mathbf{0.5442 \pm 0.0149}$
& $\mathbf{0.9209 \pm 0.0003}$
& $\mathbf{0.1709 \pm 0.0010}$ \\
\midrule

\multirow[c]{3}{*}{1}
& Reference
& None (Original)
& $0.5727 \pm 0.0650$
& $0.8981 \pm 0.0269$
& $0.1918 \pm 0.0295$ \\

& Autoresearch
& Cross-scale gating
& $0.5245 \pm 0.0149$
& $0.9271 \pm 0.0009$
& $0.1550 \pm 0.0003$ \\

& ConflictGuide
& Channel recalibration
& $\mathbf{0.5159 \pm 0.0096}$
& $\mathbf{0.9273 \pm 0.0007}$
& $\mathbf{0.1535 \pm 0.0010}$ \\
\midrule

\multirow[c]{3}{*}{2}
& Reference
& None (Original)
& $0.5727 \pm 0.0650$
& $0.8981 \pm 0.0269$
& $0.1918 \pm 0.0295$ \\

& Autoresearch
& Position-aware context
& $0.5686 \pm 0.0092$
& $0.9199 \pm 0.0010$
& $0.1719 \pm 0.0021$ \\

& ConflictGuide
& Bidirectional gating
& $\mathbf{0.5546 \pm 0.0046}$
& $\mathbf{0.9210 \pm 0.0024}$
& $\mathbf{0.1694 \pm 0.0050}$ \\
\bottomrule
\end{tabular}
\end{table*}

\paragraph{Experimental Setup.}
All models follow a two-stage protocol with model-specific budgets (mostly 100/100; see Appendix~\ref{app:experimental_details}): a task-feedback-only phase followed by two matched continuations from the same branch point. AutoResearch uses task-performance feedback alone, whereas ConflictGuide additionally observes the conflict-specific feedback in Table~\ref{tab:conflict-instances}.
Both methods use Claude Code Opus~4.6~\citep{anthropic2025claudecode} unless otherwise noted. We repeat the full process over three independent search rounds and select one winner from each trajectory for formal evaluation. We consider five heterogeneous settings: \textbf{SpecB-FNO}~\citep{qin2024toward}, two-dimensional Navier--Stokes operator learning at \(64\times64\) resolution~\citep{li2020fourier} using a width-32, four-layer FNO with 32 Fourier modes (validation NRMSE); \textbf{SNGP}~\citep{liu2020simple}, a WRN-28-2-based SNGP on CIFAR-100~\citep{krizhevsky2009learning} (temperature-scaled validation NLL); \textbf{GCNII}~\citep{chen2020simple}, node classification on Chameleon~\citep{pei2020geom} using an eight-layer GCNII with hidden dimension 64 (clean validation NLL); \textbf{ESN}~\citep{inubushi2017reservoir}, a 100-unit echo-state network for NARMA-30~\citep{schrauwen2008improving} sequence prediction (validation NRMSE); and \textbf{TCM-Lite}~\citep{liu2023learned}, learned image compression on DIV2K~\citep{agustsson2017ntire} and Flickr2K~\citep{lim2017enhanced} using a lightweight TCM configuration (estimated rate--distortion loss).

For formal evaluation, we transfer only the evolved source modifications and retrain each winner from scratch under matched settings. SpecB--FNO is evaluated on the held-out Navier--Stokes test set~\citep{li2020fourier}, SNGP on CIFAR-100~\citep{krizhevsky2009learning}, CIFAR-10~\citep{krizhevsky2009learning}, and SVHN OOD dataset~\citep{netzer2011reading}, GCNII across seven node-classification datasets~\citep{pei2020geom,yang2016revisiting}, ESN on Mackey--Glass~\citep{mackey1977oscillation}, and TCM-Lite on Kodak~\citep{liu2023learned}, with additional compression benchmarks~\citep{asuni2013testimages} used for secondary evaluation. Full data splits, model and training configurations, editable components, selection rules, and evaluation protocols are provided in the Appendix~\ref{app:experimental_details}.

\begin{table}[!t]
\centering
\caption{Ablation of ConflictGuide components. Guidance only introduces
conflict-specific guidance; full ConflictGuide adds auxiliary retention. OOD AUPR on SVHN. Mean $\pm$ s.d.; \textbf{bold} = best.}
\label{tab:ablation}
\footnotesize
\setlength{\tabcolsep}{5pt}
\renewcommand{\arraystretch}{1.15}
\begin{tabular}{@{}lcc cccc@{}}
\toprule
\multirow{2}{*}{Configuration}& \multirow{2}{*}{Guidance} & \multirow{2}{*}{Retention} 
& \multicolumn{2}{c}{SpecB-FNO}
& \multicolumn{2}{c}{SNGP} \\
\cmidrule(lr){4-5}
\cmidrule(lr){6-7}
 & \ & \
& NRMSE $\downarrow$ & ND-NMSE $\downarrow$
& NLL $\downarrow$   & OOD AUPR $\uparrow$ \\
\midrule
Baseline
& \xmark & \xmark
& $0.0382_{\pm.0015}$ & $0.0617_{\pm.0050}$
& $1.1777_{\pm.0210}$ & $0.7890_{\pm.0389}$ \\
Guidance only
& \cmark & \xmark
& $0.0348_{\pm.0015}$ & $0.0555_{\pm.0059}$
& $1.0381_{\pm.0406}$ & $0.7914_{\pm.0047}$ \\
ConflictGuide
& \cmark & \cmark
& $\mathbf{0.0344}_{\pm.0003}$ & $\mathbf{0.0530}_{\pm.0007}$
& $\mathbf{0.8455}_{\pm.0076}$ & $\mathbf{0.8131}_{\pm.0294}$ \\
\bottomrule
\end{tabular}
\vspace{-5pt}
\end{table}

\begin{figure}[!t]
    \centering
    \includegraphics[width=\linewidth]{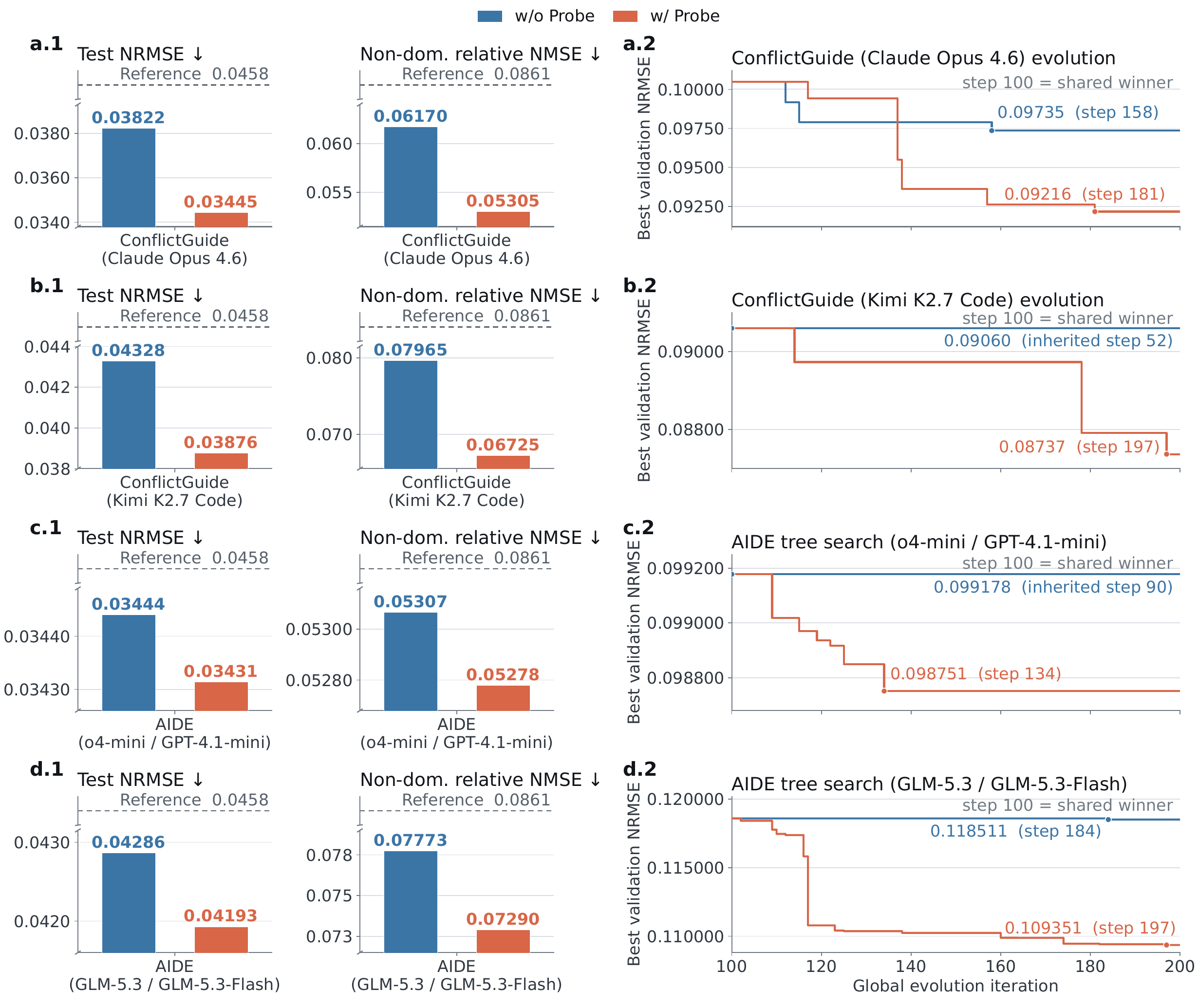}
    \caption{Probe-guided versus unguided evolution across four coding agents (a--d).
Left: test NRMSE and non-dominant relative NMSE (3-seed mean; dashed =
reference; lower is better). Right: validation NRMSE from shared step-100
winner. Blue = unguided; orange = Probe-guided.}
    \label{fig:AIDE_vs_Claude_vs_Kimi}
    \vspace{-8pt}
\end{figure}

\paragraph{Main Results.}
ConflictGuide consistently improves mean task performance over the
unmodified reference and AutoResearch across five model families and
three independent search rounds
(Tables~\ref{tab:specb_fno_results}--\ref{tab:tcm_lite_kodak_results}),
with simultaneous gains in task and conflict-related metrics in most
settings. These results support the effectiveness of conflict-aware
evolution across heterogeneous architectures and datasets. As shown in
Figure~\ref{fig:evo_three_models}, AutoResearch often plateaus after the
common branch point, while ConflictGuide continues to improve under
matched continuation budgets. This divergence supports the usefulness
of probe-derived feedback for continued refinement beyond scalar-only
plateaus.

\begin{figure}[!t]
    \centering
    \includegraphics[width=0.95\linewidth]{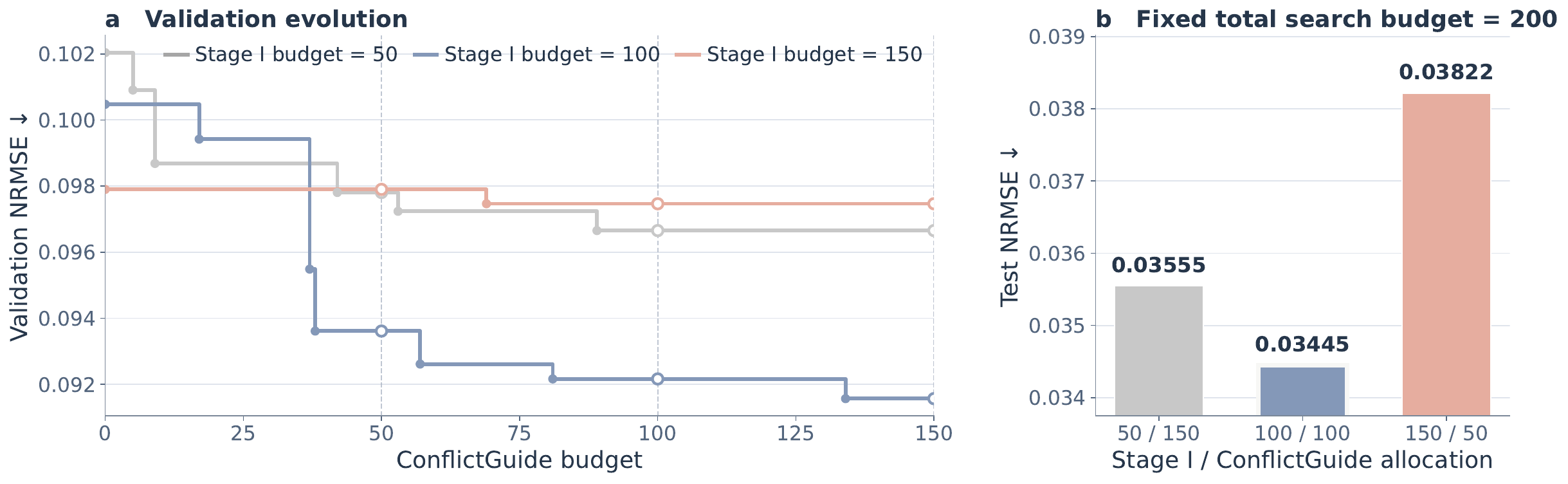}
    \caption{Budget sensitivity. (a) Round-0 validation trajectories for three Stage~I budgets. (b) Test NRMSE under three 200-iteration allocations; 100/100 balances performance and search efficiency.}
    \label{fig:hyperparameter}
    \vspace{-10pt}
\end{figure}

\begin{figure}[!t]
    \centering
    \includegraphics[width=\linewidth]{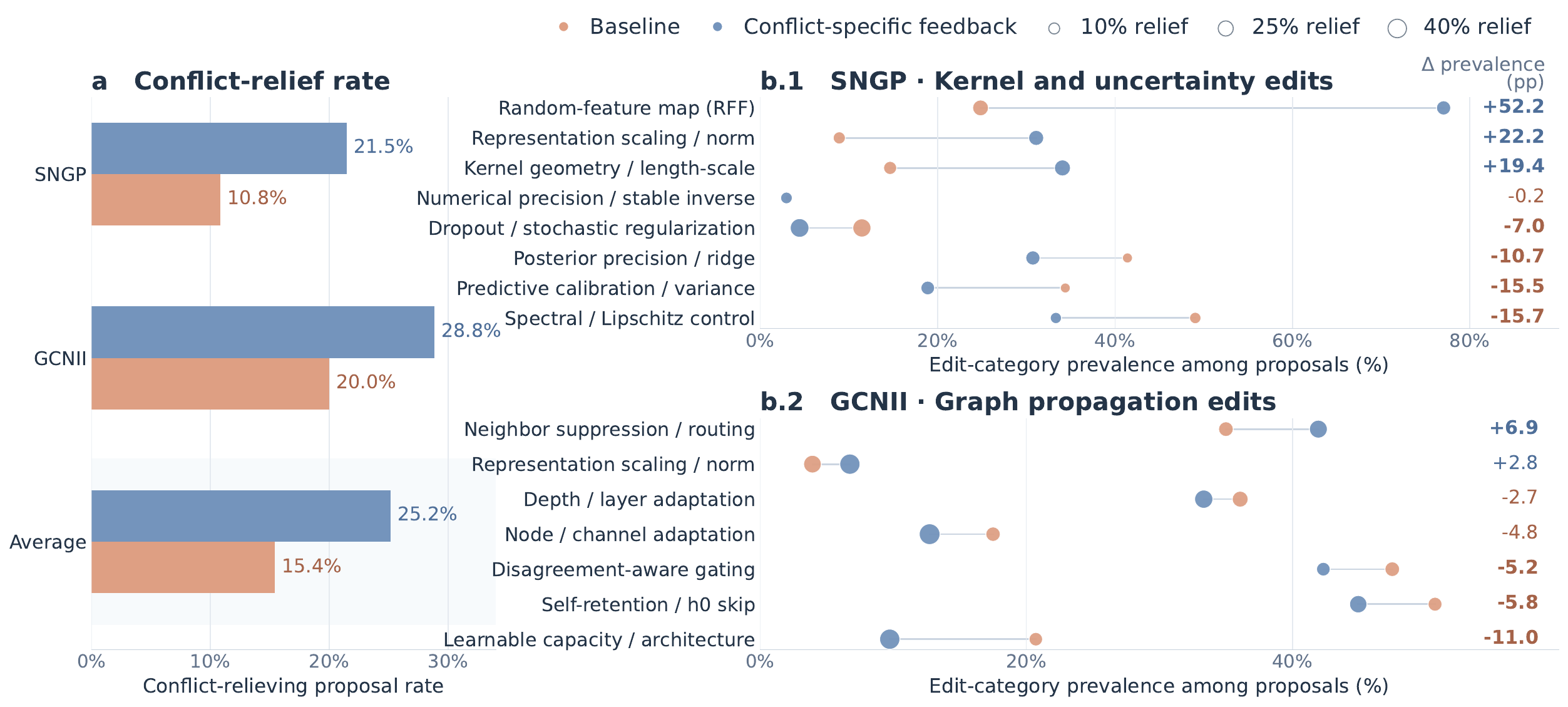}
\caption{Conflict-specific feedback on SNGP and GCNII proposals.
\textbf{(a)} Conflict-relief rates. \textbf{(b.1--b.2)} Edit-category
prevalence; dot area encodes relief rate, and lines/labels show changes
from Baseline ($\Delta$, pp). Overall, feedback increases relief rates
and shifts the distribution of code edits.}
    \label{fig:probe_analysis}
    \vspace{-5pt}
\end{figure}

\paragraph{Ablation Studies.}
As shown in Table~\ref{tab:ablation}, both conflict-specific guidance and
retention contribute to ConflictGuide. Guidance yields larger gains on
Navier--Stokes (NRMSE and ND-NMSE), whereas retention contributes more on
CIFAR-100 (NLL and OOD AUPR), suggesting complementary roles: guidance
redirects proposals via conflict-specific feedback, while retention filters
noisy task gains and preserves conflict-alleviating edits through a
separate path. We further replace the original code agent with Kimi K2.7
Code~\citep{moonshot2026kimi} and integrate Probe guidance into
AIDE~\citep{jiang2025aide} under two LLM configurations
(Figure~\ref{fig:AIDE_vs_Claude_vs_Kimi}). Across all settings,
Probe-derived feedback consistently improves task and conflict-related
metrics, indicating that the benefit is not tied to a particular code agent
or search framework. Integration details are in the Appendix~\ref{app:transfer}.

\paragraph{Hyperparameter Analysis.}
Stage~I and ConflictGuide budgets are varied over \(\{50,100,150\}\).
Figure~\ref{fig:hyperparameter}(a) shows that a 100-iteration
Stage~I provides the most favorable starting point: too few iterations
may not reach the identifiability-limited regime, while too many can
overcommit the trajectory and leave less room for Probe-guided refinement.
Under a fixed total budget of 200 iterations
(Figure~\ref{fig:hyperparameter}(b)), the 100/100 split achieves the
lowest test NRMSE. 
The 100/100 default for models with sufficient evolution capacity
was fixed before inspecting test results; model-specific exceptions
are detailed in Appendix~\ref{app:budget_exceptions}.

\paragraph{Proposals Analysis.}
Figure~\ref{fig:probe_analysis} analyzes proposals across three search
rounds for SNGP and GCNII. Conflict-specific feedback nearly doubles
the relief rate on SNGP (10.8\% to 21.5\%) and raises it from 20.0\%
to 28.8\% on GCNII. The proposal distribution also shifts: SNGP favors
random-feature-map and kernel-geometry edits while reducing generic
regularization, whereas GCNII shifts toward neighbor suppression and
away from broad capacity changes. These results show that Probe feedback
changes not only the relief rate but also the types of edits proposed,
redirecting search toward the targeted conflict mechanisms. Additional analysis is provided in the Appendix~\ref{app:additional_analysis}.

\paragraph{Value of conflict feedback.}
ConflictGuide outperforms the multi-metric and prompt-only alternatives across GCNII metrics and leads on both NLL and OOD AUPR for SNGP (Table~\ref{tab:feedback_baselines}). This suggests that guidance benefits from
measuring the instantiated competing behaviors, rather than providing
additional metrics or describing the trade-off. Appendix~\ref{app:feedback_controls} details both
alternatives.

\begin{table}[t]
\centering
\caption{Comparison of multi-metric, prompt-only, and ConflictGuide feedback on GCNII and SNGP (search round~2). Values are formal-evaluation means $\pm$ s.d.; \textbf{bold} marks the best per metric.}
\label{tab:feedback_baselines}
\small
\setlength{\tabcolsep}{4pt}
\renewcommand{\arraystretch}{1.15}
\begin{tabular}{@{}lcccc@{}}
\toprule
\multirow{2}{*}{Configuration}
& \multicolumn{2}{c}{GCNII}
& \multicolumn{2}{c}{SNGP} \\
\cmidrule(lr){2-3}
\cmidrule(lr){4-5}
& Clean NLL $\downarrow$
& Contaminated NLL $\downarrow$
& NLL $\downarrow$
& OOD AUPR $\uparrow$ \\
\midrule
Multi-metric feedback
& $0.5662 \pm 0.0909$
& $0.5679 \pm 0.0929$
& $0.8918 \pm 0.0087$ & $0.7292 \pm 0.0023$ \\
Trade-off prompt only
& $0.5694 \pm 0.0791$
& $0.5682 \pm 0.0824$
& $1.1959 \pm 0.0234$ & $0.7317 \pm 0.0024$ \\
ConflictGuide
& $\mathbf{0.5540 \pm 0.0847}$
& $\mathbf{0.5488 \pm 0.0863}$
& $\mathbf{0.8455 \pm 0.0076}$ & $\mathbf{0.7357 \pm 0.0010}$ \\
\bottomrule
\end{tabular}
\vspace{-10pt}
\end{table}

\section{Conclusion}

We introduced ConflictGuide, which incorporates competing-behavior feedback into AutoResearch. ConflictGuide-Skill uses a literature-grounded taxonomy to identify model-specific conflicts and design measurable probes. A code agent implements them before qualification and evolution. After scalar-guided exploration, probe-derived signals guide proposals and filter edits with marginal task gains but insufficient conflict alleviation. Experiments across diverse model families and code agents show that this feedback redirects search toward conflict-relevant changes and sustains progress beyond scalar-only plateaus. Limitations and future directions are discussed in the Appendix~\ref{app:limitations}.

\subsection*{AI use statement}

We used generative AI tools to conduct the AutoResearch experiments studied
in this paper and to improve the manuscript's wording. The central research
idea was developed by the authors without generative AI assistance. We
reviewed all AI-assisted work and take responsibility for the final content
of this paper, including AI-assisted text, claims, and artifacts.

\subsection*{Reproducibility statement}

The Method and Experimental Setup sections describe ConflictGuide's
two-stage search, retention rules, and evaluation protocols. Its search
loop can be reproduced by extending the publicly available AutoResearch
codebase with the feedback and retention rules specified in the Method
section. Appendices~\ref{app:skill_details}--\ref{app:transfer} document the datasets and five base models,
including their source references and experimental settings, and provide
detailed definitions of the probe metrics for all five model families.
The ConflictGuide-Skill and experimental code are included in the supplementary material.





\bibliography{iclr2027_conference}
\bibliographystyle{iclr2027_conference}

\clearpage

\appendix
\newcommand{\appplaceholder}{\mbox{}\par}

\section*{Appendix Overview}

\begin{enumerate}[label=\Alph*.]
    \item \hyperref[app:exploratory]{Exploratory Studies and Design Motivation}
    \item \hyperref[app:related_work]{Additional Related Work}
    \item \hyperref[app:skill_details]{ConflictGuide-Skill Details}
    \item \hyperref[app:taxonomy]{Full Root--Axis--Mechanism Taxonomy}
    \item \hyperref[app:identifiability]{Identifiability of Conflict-Alleviating Edits}
    \item \hyperref[app:model_probes]{Model-Specific Conflicts and Probe Metrics}
    \item \hyperref[app:experimental_details]{Experimental Details}
    \item \hyperref[app:feedback_controls]{Feedback Baselines}
    \item \hyperref[app:transfer]{Extension to Other Agents and Frameworks}
    \item \hyperref[app:additional_analysis]{Additional Results and Analyses}
    \item \hyperref[app:limitations]{Limitations and Future Directions}
\end{enumerate}



\section{Exploratory Studies and Design Motivation}
\label{app:exploratory}

This section studies when conflict-specific feedback should be introduced for
SpecB--FNO. The early-feedback study compares Scalar-only and
Probe-from-outset evolution under matched total proposal budgets. The
delayed-feedback study branches from a shared Scalar-only Stage-I checkpoint
and compares ConflictGuide with a Scalar-only continuation under matched
refinement budgets. Because both studies cover only one model family, they
motivate our two-stage design rather than establish a universally optimal
feedback schedule.

\begin{table}[ht]
\centering
\caption{Operational mechanism labels used in the SpecB--FNO proposal audit.}
\label{tab:specb_code_mechanism_categories}
\small
\setlength{\tabcolsep}{4pt}
\renewcommand{\arraystretch}{1.10}
\begin{tabularx}{\columnwidth}{@{}
>{\raggedright\arraybackslash}p{0.32\columnwidth}
>{\raggedright\arraybackslash}X
@{}}
\toprule
\textbf{Mechanism label} & \textbf{Included code changes} \\
\midrule
Frequency-specific spectral
& Fourier-mode weighting, selection, filtering, or spectral operators. \\
Spatial/local pathway
& Spatial or pointwise feature extraction and local helper paths alongside
the spectral path. \\
Normalization/activation/channel
& Normalization, nonlinear activation, channel width, or channel projection. \\
Stage fusion/residual scale
& Fusion between stages or branches and scaling of residual corrections. \\
Optimization/regularization
& Optimizer, schedule, loss regularization, or training-time stabilization. \\
\bottomrule
\end{tabularx}
\end{table}

\begin{figure*}[ht]
\centering
\includegraphics[width=\textwidth]{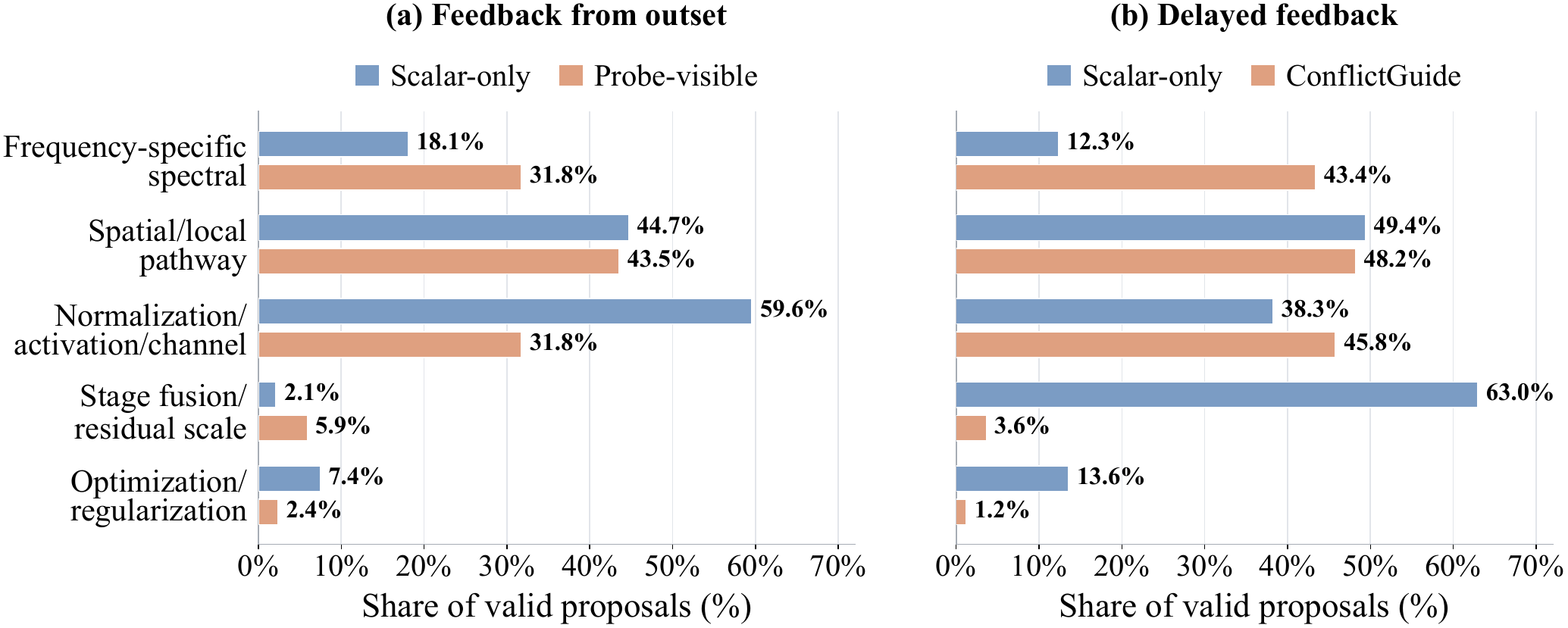}
\caption{Distribution of code-edit mechanisms in the SpecB--FNO
exploratory searches. (a) Feedback from the outset, with 94 valid
Scalar-only proposals and 85 valid Probe-visible proposals. (b) Delayed
feedback from a common Stage-I branch point, with 81 valid Scalar-only
proposals and 83 valid ConflictGuide proposals. Percentages are computed
within each arm. Mechanism labels are non-exclusive; shares therefore need not sum to 100\%.}
\label{fig:specb_code_proposal_distribution}
\end{figure*}

\begin{table}[ht]
\centering
\caption{Full-scale evaluation with feedback provided from the outset.
Mean $\pm$ standard deviation over three training seeds; lower is better.}
\label{tab:specb_early_feedback}
\small
\setlength{\tabcolsep}{5pt}
\begin{tabular}{@{}lcc@{}}
\toprule
\textbf{Configuration}
& \textbf{Test NRMSE $\downarrow$}
& \textbf{ND-NMSE $\downarrow$} \\
\midrule
Reference
& $0.045822 \pm 0.001432$
& $0.086148 \pm 0.008565$ \\
Scalar-only
& $\mathbf{0.037916 \pm 0.001519}$
& $\mathbf{0.061121 \pm 0.005326}$ \\
Probe from outset
& $0.046471 \pm 0.002050$
& $0.086498 \pm 0.009130$ \\
\bottomrule
\end{tabular}
\end{table}

\paragraph{Shared setup and proposal audit.}
Both studies use the SpecB--FNO task, proxy-search configuration, training
protocol, and full-scale evaluation described in
Appendix~\ref{app:experimental_details}. The task metric, conflict Probes, safety
metrics, and qualification thresholds are defined in
Appendix~\ref{app:model_probes}. All non-feedback settings are fixed within
each comparison. To characterize proposal distributions, we annotate code
edits with five non-exclusive mechanism labels spanning frequency-specific
spectral, spatial/local, architectural, fusion, and optimization changes
(Table~\ref{tab:specb_code_mechanism_categories}). Figure~\ref{fig:specb_representative_code_edits}
shows representative implementations, including per-mode reweighting and a
gated bypass as two distinct frequency-specific interventions.

\paragraph{Early Competing-Behavior Feedback}
\label{app:specb_early_feedback}

Providing Probe feedback from the outset shifted proposals toward frequency-specific spectral edits: their share increased from 18.1\% to 31.8\%, while normalization, activation, or channel-structure edits fell from 59.6\% to 31.8\%. Yet this narrower focus did not translate into stronger performance: Scalar-only evolution improved both Test NRMSE and ND-NMSE over the reference, whereas Probe-from-outset improved neither. Together, these results suggest that early conflict feedback steered the search toward spectral compensation before sufficiently broad task-oriented exploration.

\paragraph{Delayed Competing-Behavior Feedback}
\label{app:specb_delayed_feedback}

\begin{table}[t]
\centering
\caption{Full-scale evaluation after delayed feedback. Mean $\pm$ standard
deviation over three paired training seeds; lower is better.}
\label{tab:specb_delayed_feedback}
\small
\setlength{\tabcolsep}{5pt}
\begin{tabular}{@{}lcc@{}}
\toprule
\textbf{Configuration}
& \textbf{Test NRMSE $\downarrow$}
& \textbf{ND-NMSE $\downarrow$} \\
\midrule
Reference
& $0.045822 \pm 0.001432$
& $0.086148 \pm 0.008565$ \\
Scalar-only
& $0.038224 \pm 0.001466$
& $0.061698 \pm 0.004965$ \\
ConflictGuide
& $\mathbf{0.034447 \pm 0.000342}$
& $\mathbf{0.053048 \pm 0.000724}$ \\
\bottomrule
\end{tabular}
\end{table}

After the shared Scalar-only Stage I, ConflictGuide redirected the proposal budget toward the identified conflict mechanism: frequency-specific spectral edits rose from 12.3\% to 43.4\%, while stage-fusion and optimization-related edits sharply declined. From the same branch point, ConflictGuide reduced Test NRMSE by 9.9\% and ND-NMSE by 14.0\% relative to the matched Scalar-only continuation. These aligned shifts in proposals and performance support the two-stage rationale: broad task exploration first identifies a strong solution, after which conflict feedback focuses refinement on the under-addressed spectral behavior. This experiment motivates the delayed schedule without implying that it is optimal for every model or conflict.


\newcommand{\SpecBCodeLine}[2][0pt]{%
  {\scriptsize\ttfamily\hspace*{#1}\detokenize{#2}\par}}
\newcommand{\SpecBCodeNote}[1]{%
  {\scriptsize #1\par}}
\newcommand{\SpecBCodeTitle}[1]{%
  {\bfseries\small #1\par}\vspace{2pt}\hrule\vspace{4pt}}

\begin{figure*}[ht]
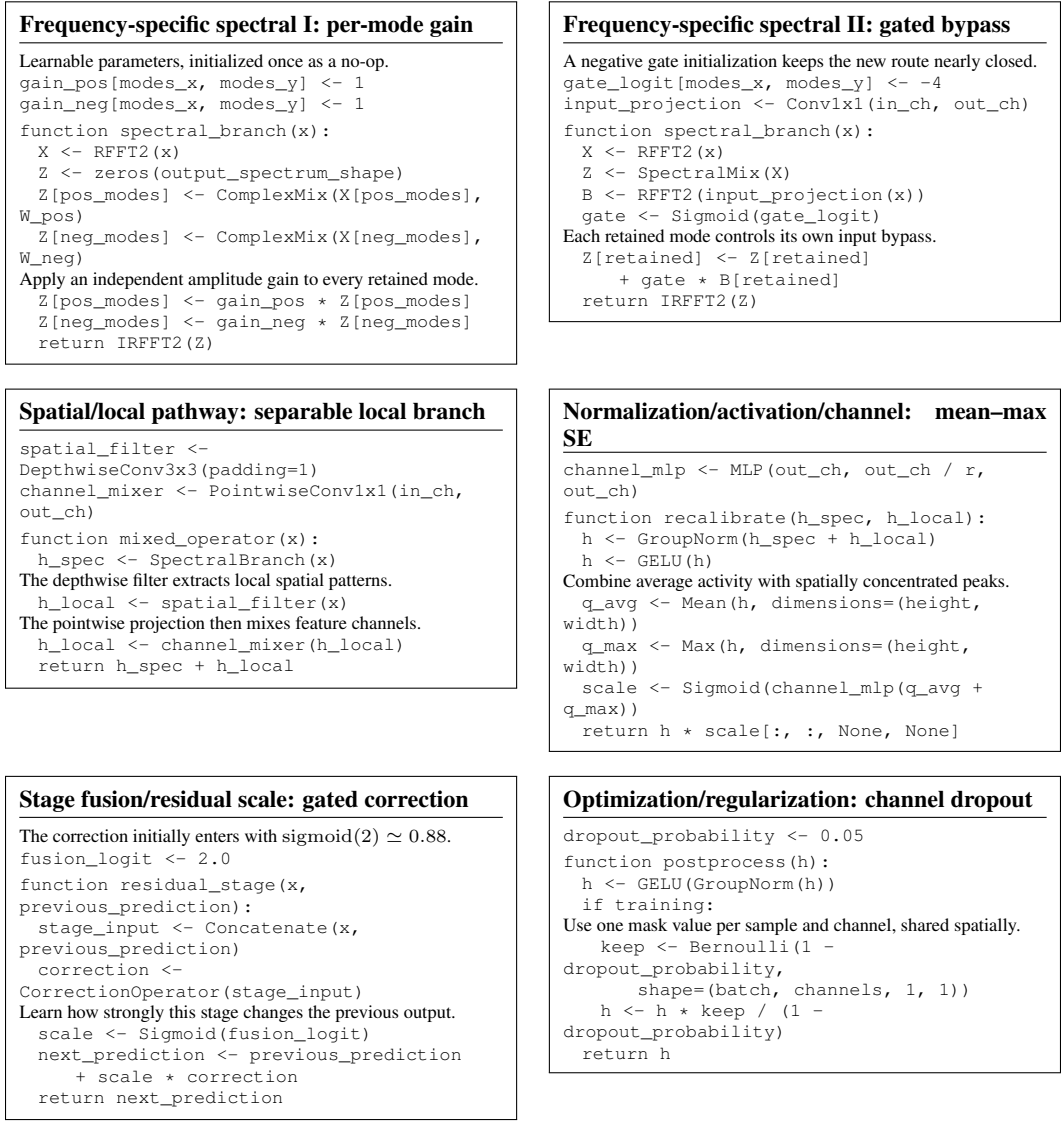

\centering
\ifdefined\nolinenumbers\nolinenumbers\fi

\begin{minipage}[t]{0.485\textwidth}
\vspace{0pt}
\setlength{\fboxsep}{5pt}
\fbox{%
\begin{minipage}[t]{\dimexpr\linewidth-2\fboxsep-2\fboxrule\relax}
\SpecBCodeTitle{Frequency-specific spectral I: per-mode gain}
\SpecBCodeNote{Learnable parameters, initialized once as a no-op.}
\SpecBCodeLine{gain_pos[modes_x, modes_y] <- 1}
\SpecBCodeLine{gain_neg[modes_x, modes_y] <- 1}
\vspace{2pt}
\SpecBCodeLine{function spectral_branch(x):}
\SpecBCodeLine[1em]{X <- RFFT2(x)}
\SpecBCodeLine[1em]{Z <- zeros(output_spectrum_shape)}
\SpecBCodeLine[1em]{Z[pos_modes] <- ComplexMix(X[pos_modes], W_pos)}
\SpecBCodeLine[1em]{Z[neg_modes] <- ComplexMix(X[neg_modes], W_neg)}
\SpecBCodeNote{Apply an independent amplitude gain to every retained mode.}
\SpecBCodeLine[1em]{Z[pos_modes] <- gain_pos * Z[pos_modes]}
\SpecBCodeLine[1em]{Z[neg_modes] <- gain_neg * Z[neg_modes]}
\SpecBCodeLine[1em]{return IRFFT2(Z)}
\end{minipage}}
\end{minipage}
\hfill
\begin{minipage}[t]{0.485\textwidth}
\vspace{0pt}
\setlength{\fboxsep}{5pt}
\fbox{%
\begin{minipage}[t]{\dimexpr\linewidth-2\fboxsep-2\fboxrule\relax}
\SpecBCodeTitle{Frequency-specific spectral II: gated bypass}
\SpecBCodeNote{A negative gate initialization keeps the new route nearly closed.}
\SpecBCodeLine{gate_logit[modes_x, modes_y] <- -4}
\SpecBCodeLine{input_projection <- Conv1x1(in_ch, out_ch)}
\vspace{2pt}
\SpecBCodeLine{function spectral_branch(x):}
\SpecBCodeLine[1em]{X <- RFFT2(x)}
\SpecBCodeLine[1em]{Z <- SpectralMix(X)}
\SpecBCodeLine[1em]{B <- RFFT2(input_projection(x))}
\SpecBCodeLine[1em]{gate <- Sigmoid(gate_logit)}
\SpecBCodeNote{Each retained mode controls its own input bypass.}
\SpecBCodeLine[1em]{Z[retained] <- Z[retained]}
\SpecBCodeLine[3em]{+ gate * B[retained]}
\SpecBCodeLine[1em]{return IRFFT2(Z)}
\end{minipage}}
\end{minipage}

\vspace{0.8em}

\begin{minipage}[t]{0.485\textwidth}
\vspace{0pt}
\setlength{\fboxsep}{5pt}
\fbox{%
\begin{minipage}[t]{\dimexpr\linewidth-2\fboxsep-2\fboxrule\relax}
\SpecBCodeTitle{Spatial/local pathway: separable local branch}
\SpecBCodeLine{spatial_filter <- DepthwiseConv3x3(padding=1)}
\SpecBCodeLine{channel_mixer <- PointwiseConv1x1(in_ch, out_ch)}
\vspace{2pt}
\SpecBCodeLine{function mixed_operator(x):}
\SpecBCodeLine[1em]{h_spec <- SpectralBranch(x)}
\SpecBCodeNote{The depthwise filter extracts local spatial patterns.}
\SpecBCodeLine[1em]{h_local <- spatial_filter(x)}
\SpecBCodeNote{The pointwise projection then mixes feature channels.}
\SpecBCodeLine[1em]{h_local <- channel_mixer(h_local)}
\SpecBCodeLine[1em]{return h_spec + h_local}
\end{minipage}}
\end{minipage}
\hfill
\begin{minipage}[t]{0.485\textwidth}
\vspace{0pt}
\setlength{\fboxsep}{5pt}
\fbox{%
\begin{minipage}[t]{\dimexpr\linewidth-2\fboxsep-2\fboxrule\relax}
\SpecBCodeTitle{Normalization/activation/channel: mean--max SE}
\SpecBCodeLine{channel_mlp <- MLP(out_ch, out_ch / r, out_ch)}
\vspace{2pt}
\SpecBCodeLine{function recalibrate(h_spec, h_local):}
\SpecBCodeLine[1em]{h <- GroupNorm(h_spec + h_local)}
\SpecBCodeLine[1em]{h <- GELU(h)}
\SpecBCodeNote{Combine average activity with spatially concentrated peaks.}
\SpecBCodeLine[1em]{q_avg <- Mean(h, dimensions=(height, width))}
\SpecBCodeLine[1em]{q_max <- Max(h, dimensions=(height, width))}
\SpecBCodeLine[1em]{scale <- Sigmoid(channel_mlp(q_avg + q_max))}
\SpecBCodeLine[1em]{return h * scale[:, :, None, None]}
\end{minipage}}
\end{minipage}

\vspace{0.8em}

\begin{minipage}[t]{0.485\textwidth}
\vspace{0pt}
\setlength{\fboxsep}{5pt}
\fbox{%
\begin{minipage}[t]{\dimexpr\linewidth-2\fboxsep-2\fboxrule\relax}
\SpecBCodeTitle{Stage fusion/residual scale: gated correction}
\SpecBCodeNote{The correction initially enters with $\mathrm{sigmoid}(2)\simeq0.88$.}
\SpecBCodeLine{fusion_logit <- 2.0}
\vspace{2pt}
\SpecBCodeLine{function residual_stage(x, previous_prediction):}
\SpecBCodeLine[1em]{stage_input <- Concatenate(x, previous_prediction)}
\SpecBCodeLine[1em]{correction <- CorrectionOperator(stage_input)}
\SpecBCodeNote{Learn how strongly this stage changes the previous output.}
\SpecBCodeLine[1em]{scale <- Sigmoid(fusion_logit)}
\SpecBCodeLine[1em]{next_prediction <- previous_prediction}
\SpecBCodeLine[3em]{+ scale * correction}
\SpecBCodeLine[1em]{return next_prediction}
\end{minipage}}
\end{minipage}
\hfill
\begin{minipage}[t]{0.485\textwidth}
\vspace{0pt}
\setlength{\fboxsep}{5pt}
\fbox{%
\begin{minipage}[t]{\dimexpr\linewidth-2\fboxsep-2\fboxrule\relax}
\SpecBCodeTitle{Optimization/regularization: channel dropout}
\SpecBCodeLine{dropout_probability <- 0.05}
\vspace{2pt}
\SpecBCodeLine{function postprocess(h):}
\SpecBCodeLine[1em]{h <- GELU(GroupNorm(h))}
\SpecBCodeLine[1em]{if training:}
\SpecBCodeNote{Use one mask value per sample and channel, shared spatially.}
\SpecBCodeLine[2em]{keep <- Bernoulli(1 - dropout_probability,}
\SpecBCodeLine[4em]{shape=(batch, channels, 1, 1))}
\SpecBCodeLine[2em]{h <- h * keep / (1 - dropout_probability)}
\SpecBCodeLine[1em]{return h}
\end{minipage}}
\end{minipage}

\vspace{0.5em}

\caption{Representative code edits for the five SpecB--FNO mechanism labels. The two frequency-specific examples contrast direct per-mode reweighting with a mode-gated bypass. The pseudocode shows the added parameters and forward computations, omitting tensor-shape handling and other implementation details.}
\label{fig:specb_representative_code_edits}
\vspace{-10pt}
\end{figure*}

\section{Additional Related Work}
\label{app:related_work}

\subsection{Automated Scientific Agents and Model Evolution}

LLM-based agents increasingly automate scientific research at different
levels. General scientific agents support stages of the research
workflow, including literature analysis, hypothesis generation, experiment
design, implementation, and iterative refinement
~\citep{lu2026towards,schmidgall2025agent,baek2025researchagent}. A second
line of work focuses on automated algorithm and model discovery. LLaMEA
evolves executable algorithms through iterative LLM-based generation and
evaluation~\citep{van2024llamea}, while AlphaEvolve extends evaluator-driven
code evolution to broader algorithmic and scientific problems
~\citep{novikov2025alphaevolve}. ASI-Evolve further explores model
architectures, training data, and learning algorithms~\citep{xu2026asi},
with related systems targeting autonomous architecture discovery and
domain-specific model design~\citep{xu2026neural,liu2026agentfold}.
More closely related to our setting, systems for automated ML experimentation
place coding agents in iterative experimental loops over existing ML
implementations. AutoResearch repeatedly modifies and evaluates training code
under a fixed experimental budget~\citep{karpathy2026autoresearch}, while
AutoSOTA automates the reproduction and subsequent improvement of existing
models~\citep{li2026autosota}. ConflictGuide falls within this paradigm,
building on an AutoResearch-style coding-agent loop that iteratively modifies
and evaluates model implementations.

\subsection{Feedback Signals in Automated Model Evolution}

Automated model evolution need not rely solely on a scalar task metric.
LLaMEA-SAGE extracts structural and complexity features from generated code,
models their relationship with performance, and translates the resulting
explanations into mutation guidance \citep{van2026llamea}. NOVA combines
modification history, verification diagnostics, metric changes, and trajectory
memory into an architecture gradient for recommender architecture
evolution \citep{liu2026nova}. Beyond performance- and structure-derived
feedback, ReVeal uses reliable self-verification and tool-based evaluation to
support iterative code improvement \citep{jin2026reveal}. Other systems
introduce criteria beyond the standard task objective. Physics-Audited
Agentic SciML evaluates candidates against machine-checkable physics
requirements and advisory numerical probes \citep{abueidda2026physics},
while automated alignment researchers use benchmark suites for specific
alignment failures to guide their mitigation \citep{yueh2026automated}.
Related work also treats the evaluation metric itself as an object of
evolution, co-evolving evaluation metrics with agent skills when reliable
task metrics are unavailable \citep{zhang2026grades}.

Collectively, these studies highlight both the limitations of scalar-only
feedback and the value of feedback tailored to what must be resolved during
evolution. Existing approaches derive such signals from code structure,
verification, external requirements, or well-characterized failure modes.
ConflictGuide focuses on a complementary yet underexplored form of feedback
in ML model evolution: how candidate edits affect competing behaviors. Such
feedback becomes especially important when further progress depends on
balancing these behaviors, while their behavior-specific effects remain
obscured by the scalar task metric. ConflictGuide exposes these effects
through quantitative Probes, providing conflict-specific feedback to guide
subsequent evolution.

\subsection{Diagnosis-Guided Automated Model Evolution}

Beyond providing richer feedback, a more targeted line of work uses diagnosis
to determine what should be modified next. Self-EvolveRec augments
recommendation metrics with a Model Diagnosis Tool and qualitative user
feedback, translating diagnosed weaknesses into directional guidance for
model evolution \citep{kim2026self}. GoalEvolve identifies dominant
bottlenecks from multi-objective target gaps, localizes them to the relevant
optimization stages, and uses these diagnoses to focus source-code
modifications \citep{liu2026goalevolve}. EvoPINN similarly summarizes
training traces into diagnostics of convergence, stability, and physics-loss
behavior, conditioning representation or training-program proposals on the
diagnosed state \citep{yin2026evopinn}. Unlike general feedback enrichment,
these methods explicitly diagnose a limiting factor and use that diagnosis to
direct the next modification. Related agentic systems extend this diagnosis-to-intervention pattern beyond
automated model evolution. Mechanist proposes and tests mechanistic
hypotheses about model behavior and translates validated mechanisms into
targeted interventions \citep{wang2026mechanist}. AutoSaddler diagnoses
failure traces to generate targeted patches for agent harnesses
\citep{park2026autosaddler}, while AgentDebugX organizes agent debugging
around detection, root-cause attribution, recovery, and rerun
\citep{zhu2026agentdebugx}.

ConflictGuide brings this diagnosis-guided paradigm to competing behaviors in
ML model evolution. For a given model--task setting, a model-design conflict
represents a persistent mechanism-level tension whose state evolves with the
model. ConflictGuide therefore maintains a consistent diagnostic target while
recomputing quantitative Probes after each edit, providing an up-to-date view
of how the competing behaviors have shifted. This consistent diagnostic
coordinate enables edits to be compared across the evolution trajectory,
revealing whether each modification alleviates or aggravates the conflict and
providing targeted evidence to guide subsequent proposals.

\section{ConflictGuide-Skill}
\label{app:skill_details}

ConflictGuide-Skill packages the conflict taxonomy and its operational
procedures as a reusable agent skill. Given a target model, the skill first
maps model- and task-specific evidence to the fixed taxonomy, then instantiates
the selected mechanism as a pair of competing behaviors, and finally designs
and qualifies Probes that expose behavior-specific changes not captured by the
scalar task metric $S$. Throughout this process, the skill distinguishes
three conceptual levels: a \emph{taxonomy mechanism} is reusable conflict
vocabulary; a \emph{behavioral conflict} is its model-specific realization;
and a \emph{Probe} is a measurement that operationalizes one of the competing
behaviors.

The skill supports three modes. \textsc{Analyze} performs conflict
identification and Probe design without modifying model code.
\textsc{Implement} additionally integrates the minimum read-only diagnostic
code required to compute the fixed Probes. \textsc{Qualify} evaluates an
existing Probe Card through null calibration and paired replay. None of these
modes starts model evolution automatically.

\begin{table}[!th]
\centering
\small
\caption{ConflictGuide-Skill workflow. Each stage must be completed before its
output is used by the next stage.}
\label{tab:skill_workflow}
\begin{tabularx}{\linewidth}{
    @{}
    >{\raggedright\arraybackslash}p{0.08\linewidth}
    >{\raggedright\arraybackslash}p{0.19\linewidth}
    >{\raggedright\arraybackslash}X
    >{\raggedright\arraybackslash}p{0.25\linewidth}
    @{}
}
\toprule
Stage & Operation & Main evidence & Output \\
\midrule

0 &
Model context &
Architecture, task, data, objective, $S$, editable scope, and runtime
budget &
Normalized Model Context Card \\

1 &
Taxonomy screening &
Target-model code and configuration, interpreted through the fixed
Root--Axis--Mechanism taxonomy &
Taxonomy selection or explicit abstention \\

2 &
Conflict instantiation &
Selected mechanism, task relevance, shared component or resource, and
bidirectional interference &
Model-specific competing behaviors and an observability analysis of
$S$ \\

3 &
Probe design &
Existing outputs, deterministic diagnostics, and lightweight read-only
observables &
Fixed Conflict Probe Card and implementation specification \\

4 &
Probe qualification &
Null replicates and pre-specified parent--candidate replay pairs &
Qualified, rejected, or inconclusive Probes \\

\bottomrule
\end{tabularx}
\end{table}

\subsection{Interface and workflow}

\paragraph{Inputs.}
The skill receives a target-model context comprising the architecture and
forward path, task and dataset, training objective, scalar task metric,
evolution boundary, available evaluation outputs, and runtime budget. When
source code and configuration files are available, the agent extracts this
information directly and records the corresponding evidence locations.
Unavailable fields are marked as unknown rather than inferred from the model
name.

\paragraph{Workflow.}
ConflictGuide-Skill follows the five-stage workflow summarized in
Table~\ref{tab:skill_workflow}. Each stage has an explicit evidence boundary.
In particular, conflict selection precedes Probe construction and does not use
evolution trajectories, final-test results, or successful evolved
architectures. This prevents a conflict from being selected retrospectively
to explain an observed evolution outcome.

\paragraph{Outputs.}
A successful run returns the selected taxonomy entry, the instantiated
behavior pair, their shared coupling, an analysis of what the scalar task
metric captures or obscures, the minimal Probe set, its qualification state,
and unresolved assumptions. These elements are also serialized as a
machine-checkable Conflict Probe Card. If the available evidence is
insufficient, the skill returns an abstention or rejection record instead of
forcing a conflict.

\subsection{Taxonomy screening and conflict instantiation}

\paragraph{Taxonomy basis.}
ConflictGuide-Skill uses the fixed Root--Axis--Mechanism taxonomy introduced
in Section~\ref{app:taxonomy}. The taxonomy was constructed through
agent-assisted literature collection followed by researcher consolidation and
verification. It provides mechanism-level vocabulary for identifying
behavioral conflicts across model families, rather than prescribing
model-specific behavior pairs.

\paragraph{Taxonomy screening.}
For each target model, the skill first identifies candidate Roots, restricts
them to their permitted axes, and then selects the most relevant mechanisms
from the fixed taxonomy. The selection is grounded in evidence from the model
architecture, task and data, training objective, evaluation setting, and
component interfaces. Causal links between mechanisms and optimization
signatures are recorded when supported by the available evidence.

Screening uses only information available before evolution. It does not rely
on evolution trajectories, ablations, held-out final-test results, or
successful evolved solutions. The skill must use the Root, axis, and mechanism
identifiers defined in the taxonomy and cannot invent or rename taxonomy
entries. A compute test further distinguishes resource limitations from
structural conflicts: if substantially greater compute would remove the
tension, a resource-related Root is considered; otherwise, the skill examines
structural, informational, distributional, normative, data-related, or
evaluation-related Roots.

\begin{table}[!th]
\centering
\small
\caption{Main abstention and rejection conditions used by
ConflictGuide-Skill.}
\label{tab:skill_abstention}
\begin{tabularx}{\linewidth}{
    @{}
    >{\raggedright\arraybackslash}p{0.55\linewidth}
    >{\raggedright\arraybackslash}X
    @{}
}
\toprule
Condition & Skill response \\
\midrule

Insufficient model, task, data, objective, or scalar-metric context &
\textbf{Abstain.} The taxonomy selection cannot be grounded in
target-specific evidence. \\

No two independently desirable behaviors &
\textbf{Reject.} The proposed pair does not constitute a task-relevant
behavioral conflict. \\

No identifiable shared component, representation, parameter path, or budget &
\textbf{Reject.} The proposed behaviors lack a defensible coupling mechanism. \\

Only one interference direction is plausible &
\textbf{Reject.} The case describes a one-sided failure rather than competing
behaviors. \\

Behavior-specific observables are unavailable or incomparable across
candidates &
\textbf{Abstain or reject.} The conflict cannot be operationalized reliably
under the current evaluation setting. \\

A Probe duplicates the scalar task metric, leaks unavailable information, or
depends on post-evolution tuning &
\textbf{Reject.} The proposed measurement does not provide valid fixed
feedback. \\

\bottomrule
\end{tabularx}
\end{table}

\paragraph{Model-specific instantiation.}
After selecting a taxonomy mechanism, the skill instantiates it as two
model-specific competing behaviors. Both behaviors must be independently
desirable for the stated task or deployment setting. The skill identifies the
shared representation, component, parameter path, or resource that couples
them and specifies both interference directions: why improving behavior A may
harm behavior B, and vice versa. Each direction must be supported by available
evidence or explicitly marked as a falsifiable hypothesis.

The skill then assesses whether the scalar task metric can distinguish changes
in the two behaviors. It records what the metric captures, what it aggregates
away, and whether its conflict visibility is \texttt{yes},
\texttt{partially}, or \texttt{no}. Evaluation mechanisms such as metric
insensitivity or aggregate masking are included only when supported by the
evidence. Limited metric observability alone is insufficient to establish a
behavioral conflict.

\paragraph{Abstention and rejection.}
The skill does not infer conflicts from model names or transfer them directly
from superficially similar examples. It abstains when the available model,
task, data, objective, or evaluation context is insufficient to support a
taxonomy selection. It rejects an instantiation when the two behaviors are
not independently desirable, lack a defensible shared coupling, or do not
exhibit plausible interference in both directions. It also rejects Probes
that duplicate the scalar task metric, depend on post-evolution tuning, leak
unavailable information, or are not comparable across candidates. The main
conditions are summarized in Table~\ref{tab:skill_abstention}.

An abstention record identifies the evidence needed to resume the analysis,
whereas a rejection record states which conflict or Probe requirement failed.
In either case, the skill stops before Probe qualification.

\subsection{Probe design, implementation, and qualification}

\paragraph{Probe design.}
Given an instantiated conflict, the skill designs the smallest fixed Probe set
that reveals behavior-specific changes not already visible in the scalar task
metric $S$. Candidate observables are prioritized in the following
order: direct ground-truth measurements, existing model outputs, internal
states, representation geometry, gradient quantities, known data structure,
deterministic counterfactual inference, and fixed heuristics. Separately
trained diagnostic models or synthetic auxiliary tasks are used only when no
direct observable is available and require explicit justification.

A separate Probe is not required for each behavior. If $S$ already
measures one behavior at the relevant granularity, a single additional Probe
for the other behavior is sufficient. Before evolution, the skill fixes the
evaluation data, preprocessing, partitions, reference values, perturbations,
randomness, aggregation, normalization, and desirable metric directions.
Probe definitions cannot depend on errors made by future candidates,
final-test information unavailable during evolution, or thresholds chosen
after observing evolution outcomes.

Each Probe records the fields summarized in
Table~\ref{tab:probe_specification}. These fields make the intended behavior,
additional information, computational cost, and possible failure modes
inspectable before the Probe is exposed to the evolution agent.

\begin{table}[h]
\centering
\small
\caption{Core fields of a Conflict Probe specification.}
\label{tab:probe_specification}
\begin{tabularx}{\linewidth}{
    @{}
    >{\raggedright\arraybackslash}p{0.35\linewidth}
    >{\raggedright\arraybackslash}X
    @{}
}
\toprule
Field & Description \\
\midrule

Target behavior &
The declared competing behavior measured by the Probe. \\

Observable and computation &
Source value, formula or algorithm, sampling, and aggregation. \\

Direction and normalization &
Whether an increase or decrease is desirable and how values are made
comparable across candidates. \\

Evaluation setting &
Frozen dataset, split, references, perturbations, and randomness policy. \\

Additional requirements &
Whether extra training, labels, hooks, checkpoints, or model outputs are
required. \\

Relation to scalar task
metric &
What behavior-specific information the Probe adds beyond the scalar task
metric. \\

Cost and validity &
Runtime and memory cost, leakage controls, interpretation, and known failure
modes. \\

\bottomrule
\end{tabularx}
\end{table}

\paragraph{Read-only implementation.}
In \textsc{Implement} mode, a code agent adds only the minimum evaluation
function or diagnostic hook required to compute the fixed Probes. The
implementation may load a checkpoint, access an existing output or internal
state, and emit metric values. It may not modify the training objective,
gradients, training data, model parameters, optimization procedure, or
candidate-retention rule. Probe implementation therefore remains a
measurement operation rather than an intervention. The modified files and
Probe entrypoints are recorded in the Probe Card.

\paragraph{Calibration and model-specific checks.}
Probe calibration assesses measurement variability under no intended
model change, using model-specific sampling units and threshold
estimators. Additional checks include structural-variant comparisons,
candidate replay, and perturbation controls. Their scope and decision
criteria depend on the model-specific protocol. Initial checks may
precede Stage I, while continuation calibration is performed at the
Stage-I boundary before probe-guided search.
A worked example for SpecB--FNO is provided in
Section~\ref{app:skill_worked_example}.

\subsection{Outputs and reproducibility}

ConflictGuide-Skill produces a structured Conflict Probe Card whose main
fields are listed in Table~\ref{tab:probe_card_outputs}. The same information
is first presented as a concise human-readable summary, allowing researchers
to inspect the selected mechanism, evidence boundary, competing behaviors,
coupling, observability gap, and qualification state before using the
machine-readable artifact.

The complete taxonomy is provided in
Section~\ref{app:taxonomy}. The machine-readable taxonomy index, output
schemas, validation scripts, and model-specific example Probe Cards will be
released in a public repository upon acceptance. The schemas specify the
required fields for model context, behavior instantiation, and Conflict Probe
Cards, while the validation scripts check taxonomy selections and generated
cards for structural consistency. The examples illustrate the expected
reasoning process and output format rather than serving as model-name lookup
rules: the skill independently screens the taxonomy for every new target
model.

The skill also separates measurement specification from evolution policy.
The Probe Card defines the scalar task metric, behavior-specific Probes, and
their calibrated variability thresholds, but does not encode the
Stage-I/Stage-II schedule, candidate-retention routes, or
conflict-alleviation rule. These choices belong to the evolution protocol
described in the main paper. This separation allows the conflict definition
and its measurements to be inspected and qualified independently of the
subsequent evolution algorithm.

\begin{table}[t]
\centering
\small
\caption{Inspectable outputs in a Conflict Probe Card.}
\label{tab:probe_card_outputs}
\begin{tabularx}{\linewidth}{
    @{}
    >{\raggedright\arraybackslash}p{0.25\linewidth}
    >{\raggedright\arraybackslash}X
    @{}
}
\toprule
Card component & Recorded information \\
\midrule

Model context &
Architecture, task, data, objective, scalar task metric, evolution boundary,
budget, and evidence sources. \\

Taxonomy selection &
Selection status, Root, axis, mechanism, causal links, optimization
signatures, and abstention reason when applicable. \\

Behavioral conflict &
Competing behaviors, desirable directions, shared coupling, interference
directions, confidence, and rejection reason when applicable. \\

Feedback observability &
Definition and direction of the scalar task metric, conflict visibility, and
behavior-specific information it aggregates away. \\

Probes &
Fixed Probe specifications, information added beyond the scalar task metric,
computational costs, and failure modes. \\

Qualification &
Null calibration, scalar-tied informativeness test, direction test, decision
state, and rejection conditions. \\

Implementation and warnings &
Modified files, diagnostic entrypoints, unresolved assumptions, and validity
warnings. \\

\bottomrule
\end{tabularx}
\end{table}

\subsection{Worked Example: SpecB--FNO}
\label{app:skill_worked_example}

\paragraph{Model evidence and conflict instantiation.}
SpecB--FNO predicts fluid dynamics using Fourier-mode mixing and a
residual-correction pathway. The conflict is mapped to R2.M2
(frequency-structure mismatch) on the space axis. The two desirable
behaviors are accurate prediction of dominant spectral components and
accurate prediction of non-dominant components. Both depend on shared
model components. The working hypothesis is that emphasizing dominant
components can leave low-energy modes underfit, while stronger
non-dominant correction can interfere with dominant-mode accuracy.
This coupling is a model-specific hypothesis motivated by the
architecture.

\paragraph{Probe construction.}
The scalar task metric is validation NRMSE, which aggregates prediction
errors across the field. A target-derived cumulative-energy mask
separates dominant and non-dominant Fourier components independently
of candidate predictions. The primary Probe is non-dominant relative
NMSE, $Z_{\mathrm{ND}}$, with lower values indicating better prediction
of non-dominant components. Dominant-mode relative NMSE and late-rollout
NRMSE serve as guard metrics. The complete definitions are provided in
Appendix~\ref{app:probe_specb_fno}.

\paragraph{Initial qualification.}
Before Stage I, an initial panel evaluated three structural variants
under three seeds, yielding nine evaluation records. The associated
qualification report records that all model-specific gates passed and
sets \texttt{qualified\_for\_evolution=true}. These records concern the
specified structural-variant checks; they are not nine scalar-tied
parent--candidate comparisons.

\paragraph{Calibration before Stage II.}
After Stage I, the continuation protocol was calibrated using four
paired no-edit comparisons and a four-seed transition replay. This
calibration was completed and frozen before probe-guided continuation.
The associated record also documents implementation repairs involving
parameter materialization, optimizer coverage, and stage-fusion
training.

For the documented historical continuation, the direct task route
required $G_t\geq5\times10^{-4}$. The auxiliary route required
$G_t\geq2.5\times10^{-4}$ and at least a $1\%$ relative improvement in
$Z_{\mathrm{ND}}$, together with the registered dominant-mode and
late-rollout anchor guards. These are the recorded execution rules
for this continuation.

\section{Full Root--Axis--Mechanism Taxonomy}
\label{app:taxonomy}

This appendix documents the fixed taxonomy used by ConflictGuide-Skill.  The
taxonomy contains nine top-level Roots, six Axes, 110 Root-specific mechanisms,
and three standalone optimization artifacts.  R7 is further divided into
R7a--R7c because impossibility, normative--utility trade-offs, and boundary
calibration require different diagnostic logic.  The inventory below preserves
all active mechanism identifiers, names, and core diagnostic definitions from
the operational taxonomy.  To keep the paper appendix inspectable, we omit
repeated axis-specific examples, the machine-readable output schema, and
implementation instructions; these remain in the released Skill specification.

\begin{figure}[h]
    \centering
    \includegraphics[width=\linewidth]{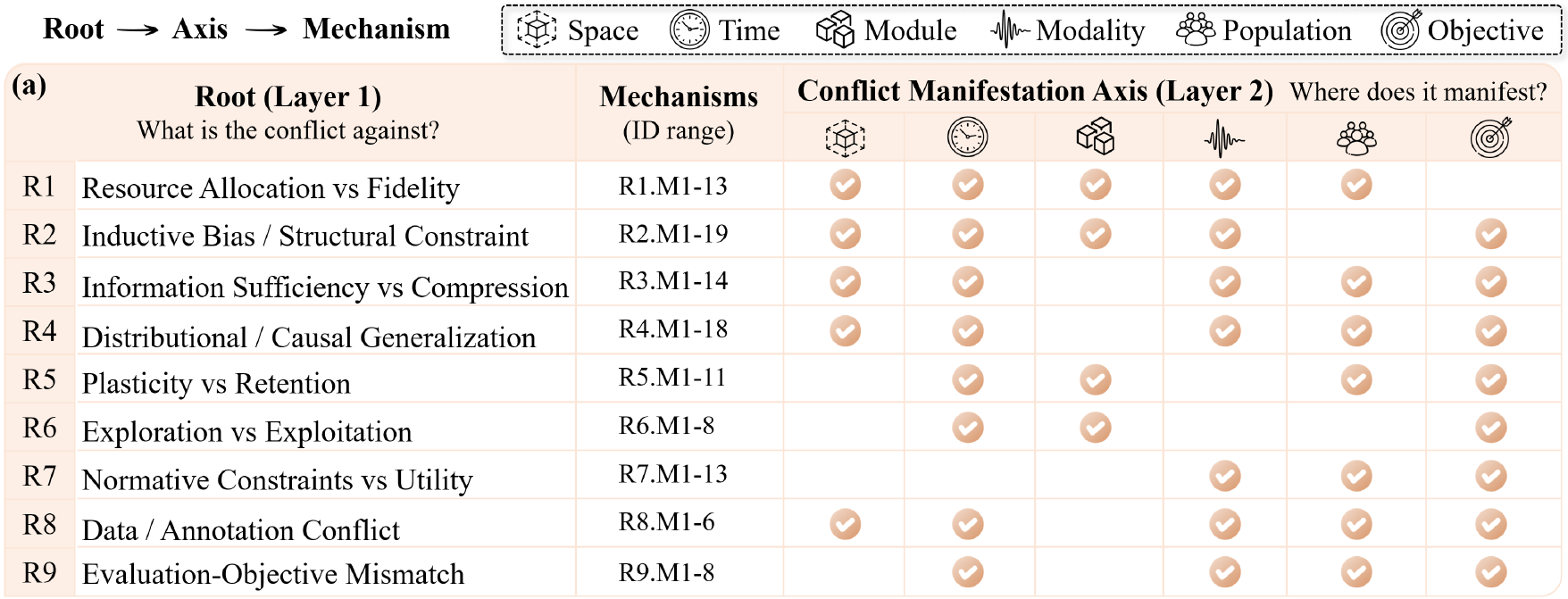}
    \caption{Root--Axis--Mechanism taxonomy.}
    \label{fig:app_taxonomy}
\end{figure}

\begin{table*}[h]
\centering
\small
\setlength{\tabcolsep}{4pt}
\renewcommand{\arraystretch}{1.10}
\caption{Root-level semantics and selection boundaries.
Figure~\ref{fig:app_taxonomy} shows the admissible axes and mechanism ranges;
this table provides the complementary criteria used to distinguish and
causally relate the Roots.}
\label{tab:taxonomy_roots}
\begin{tabularx}{\textwidth}{
    @{}
    >{\raggedright\arraybackslash}p{0.045\textwidth}
    >{\raggedright\arraybackslash}p{0.205\textwidth}
    >{\raggedright\arraybackslash}p{0.12\textwidth}
    >{\raggedright\arraybackslash}X
    >{\raggedright\arraybackslash}p{0.18\textwidth}
    @{}
}
\toprule
Root &
Conflict is against &
Compute test &
Primary discriminator &
Typical causal role \\
\midrule

R1 &
Compute, memory, latency, parameter count, or bandwidth &
Partially resolvable &
The conflict substantially eases with greater capacity, resolution, or
evaluation budget. &
May induce downstream information loss or structural compromise. \\

R2 &
Mismatch between an inductive bias or structural constraint and the signal &
Not resolvable &
The imposed prior is too weak, too strong, or incompatible with the
signal structure; architectural change is required. &
Often upstream of shortcut learning or failure under distribution shift. \\

R3 &
Information loss imposed by compression or abstraction &
Not resolvable &
Required information has been discarded or incompatible information
demands must share a limited representation. &
Often downstream of resource limits and upstream of uncertainty or
generalization failures. \\

R4 &
The gap between the training distribution and deployment conditions &
Not resolvable &
Performance relies on non-causal or distribution-specific evidence that
fails under domain, group, temporal, or intervention shifts. &
Usually a downstream symptom; upstream causes should be identified. \\

R5 &
New adaptation competing with retained knowledge or behavior &
Partially resolvable &
The conflict arises sequentially across training stages rather than
simultaneously among objectives. &
May follow distribution, label-process, or normative changes. \\

R6 &
A fixed search budget divided between coverage and solution quality &
Partially resolvable &
The conflict concerns exploration of solutions, actions, or hypotheses,
not coverage of demographic populations. &
May expose resource limits or aggravate structural and normative violations. \\

R7 &
Externally specified human values, rights, or policy constraints &
Not resolvable; exogenous &
The opposing requirement originates outside the model and may involve
impossibility, a normative--utility frontier, or boundary calibration. &
Can motivate later constraint training and its associated capability cost. \\

R8 &
Instability, ambiguity, sparsity, or inconsistency in the supervision process &
Not resolvable &
The target label or annotation process is itself uncertain or changing;
model-side changes alone cannot remove the conflict. &
Typically upstream of overfitting, shortcut learning, and apparent OOD failure. \\

R9 &
Mismatch between what is measured and what is actually desired &
Not resolvable &
The metric, benchmark, or evaluator cannot detect the relevant failure
or diverges from the intended objective. &
Upstream of other diagnoses; affected evidence should be treated as provisional. \\

\bottomrule
\end{tabularx}
\end{table*}

\subsection{Taxonomy conventions and selection rules}
\label{app:taxonomy-conventions}

\paragraph{Root--Axis--Mechanism representation.}
A \emph{Root} specifies what the conflict fundamentally pulls against and
therefore whether it can be removed by additional compute or instead requires a
change in architecture, information, data, objectives, or evaluation.  An
\emph{Axis} specifies where the conflict manifests: \texttt{space},
\texttt{time}, \texttt{module}, \texttt{modality}, \texttt{population}, or
\texttt{objective}.  A \emph{Mechanism} specifies the concrete failure pattern.
Each Root has an axis whitelist.  A mechanism manifesting on multiple axes is
recorded once per axis with separate evidence; this repetition does not create
a new mechanism.

\paragraph{Selection order and evidence boundary.}
ConflictGuide-Skill first selects a Root, then an admissible axis, and finally
the most relevant mechanism.  Every selection must be justified using the
target model's architecture, task and data, training objective, and evaluation
setting.  Relevance screening does not inspect evolution outcomes, ablations,
clean-run results, or held-out test results.  It cannot invent or rename Root,
axis, or mechanism identifiers.

\paragraph{Root discriminator.}
The primary test asks whether $100\times$ more compute would dissolve the
conflict.  Conflicts that disappear or substantially ease are generally routed
to R1, R5, or R6; persistent conflicts are routed to R2, R3, R4, R7, R8, or R9.
When the answer is partial, both Roots are retained and linked rather than
forced into a single category.  For example, a memory wall may induce an R1
context bottleneck that in turn causes R3 information loss.

\paragraph{Causal links and optimization signatures.}
Mechanism-level \texttt{downstream\_of} and \texttt{upstream\_of} links encode
causal structure.  R4 failures are often symptoms, whereas R8 and R9 frequently
act upstream; R9 additionally makes evidence based on the affected metric
provisional.  Optimization signatures---including gradient-direction conflict,
gradient-scale imbalance, loss domination, convergence-rhythm conflict,
representation-demand conflict, Pareto incompatibility, and weighted-sum
masking---annotate mechanisms rather than form a separate Root.

\paragraph{Independent-parameters test.}
If a conflict would remain with completely independent parameters, its
substantive Root is retained and any gradient pattern is only an optimization
signature.  If the conflict would disappear, it is a pure parameter-sharing or
update-merging artifact and is represented by \texttt{OPT.M1--OPT.M3}.

\subsection{Root and axis overview}
\label{app:taxonomy-overview}

The whitelists implement routing decisions.  In particular, R2 and R6 exclude
\texttt{population}: population shifts belong to R4, protected-group trade-offs
to R7, and population-dependent annotation failures to R8.  R6 concerns
solution-space coverage, not demographic coverage.

\subsection{Complete mechanism inventory}
\label{app:taxonomy-inventory}

Table~\ref{tab:taxonomy-mechanisms} lists all 110 active mechanisms.  The
definitions retain the diagnostic question from the fixed library; detailed
axis-specific manifestations and cross-references are omitted only when they
repeat the general rules above.

\begingroup
\footnotesize
\setlength{\tabcolsep}{4pt}
\renewcommand{\arraystretch}{0.92}
\begin{longtable}{@{}>{\RaggedRight\arraybackslash}p{0.10\linewidth} >{\RaggedRight\arraybackslash}p{0.25\linewidth} >{\RaggedRight\arraybackslash}p{0.59\linewidth}@{}}
\caption{Complete Root-specific mechanism inventory.}\label{tab:taxonomy-mechanisms}\\
\toprule
ID & Mechanism & Core diagnostic definition \\
\midrule
\endfirsthead
\multicolumn{3}{l}{\small\itshape Table~\ref{tab:taxonomy-mechanisms} continued}\\
\toprule
ID & Mechanism & Core diagnostic definition \\
\midrule
\endhead
\midrule
\multicolumn{3}{r}{\small\itshape Continued on next page}\\
\endfoot
\bottomrule
\endlastfoot

\multicolumn{3}{@{}l}{\rule{0pt}{2.5ex}\textbf{R1: Resource allocation vs fidelity}}\\[-1pt]

\texttt{R1.M1} & Insufficient fidelity & Does the model systematically underfit the target because the allocated resolution, step count, or capacity is too low? \\[1pt]

\texttt{R1.M2} & Uniform allocation mismatch & Does a uniform allocation waste resources on easy cases while remaining insufficient for hard cases, long-tail cases, or complex samples? \\[1pt]

\texttt{R1.M3} & Insufficient expressive bandwidth & Is the current positional encoding, basis, feature map, latent grid, channel rank, codebook, recurrent width, or local representation insufficient to express the required rate of variation? - Primarily \texttt{space} and \texttt{time}. May be \texttt{module} when a specific branch's rank is the limit, or \texttt{modality} when a specific modality's encoder is the limit. - If the loss is \emph{informational} rather than \emph{budgetary} (i.e.~more capacity would not help because the representation deliberately discards the factor), use R3 instead. \\[1pt]

\texttt{R1.M4} & Added freedom causing overfitting & After adding local parameters, high-frequency encoding, extra steps, extra depth, or extra experts, does the model become more likely to fit noise, label errors, spurious patterns, or training-set-specific details? \\[1pt]

\texttt{R1.M5} & Cross-partition consistency issue & Do outputs become inconsistent across the partitions created by the chosen axis? \\[1pt]

\texttt{R1.M6} & Local fidelity vs global structure competition & After strengthening local or short-range fidelity, does the model sacrifice global layout, long-range dependency, overall semantics, global constraints, or cross-region coordination? \\[1pt]

\texttt{R1.M7} & Cost superlinearity & Does covering a small number of hard cases require a disproportionate global increase in cost? \\[1pt]

\texttt{R1.M8} & Hardware / throughput wall & Does the model theoretically need more fidelity but is practically limited by the deployment envelope? \\[1pt]

\texttt{R1.M9} & Allocation imbalance & Are compute loads, sample counts, gradient amounts, or assignments severely imbalanced across the partitions of the chosen axis? \\[1pt]

\texttt{R1.M10} & Insufficient sharing & Does excessive partitioning prevent common patterns from being reused across similar regions, steps, modules, modalities, or clients? \\[1pt]

\texttt{R1.M11} & Redundant partitioning & Do multiple experts, adapters, heads, branches, local modules, steps, or scales repeatedly learn similar patterns instead of forming an effective division of labor? \\[1pt]

\texttt{R1.M12} & Allocation / iteration mechanism pathology & Is the mechanism that decides the allocation, or the operator that drives the iteration, itself unstable or degenerate? - Routing collapse (\texttt{module}): routing persistently favors a small number of experts, paths, adapters, token groups, or modules, leaving others underused. - Dynamic allocation instability (\texttt{module}, \texttt{space}, \texttt{time}): routing, adaptive depth, token selection, region refinement, or expert assignment is highly sensitive to small input perturbations, training stage, or batch composition. - Iterative dynamics pathology (\texttt{time}): the iteration operator itself causes error accumulation across steps, Jacobian contraction or expansion, representation collapse over depth, or vanishing/exploding propagation - the failure grows with step count rather than being relieved by it. - This is a \textbf{control-loop or operator failure, not a budget shortfall}. Do not select it merely because capacity is insufficient - that is R1.M1. Do not select it for state leaving the valid region without operator pathology - that is R1.M5 \texttt{time}. - Modality routing collapse is \texttt{modality}. \\[1pt]

\texttt{R1.M13} & Partitioning vs interpretability & As the number of modules, steps, scales, or routing paths increases, does model behavior become harder to interpret, attribute, or debug? \\[1pt]

\multicolumn{3}{@{}l}{\rule{0pt}{2.5ex}\textbf{R2: Inductive bias vs signal structure}}\\[-1pt]

\texttt{R2.M1} & Single-bias residual \emph{\texttt{under\_constrained}} & Does the dominant architecture, basis, operator, prior, or attention pattern fit only part of the signal while leaving systematic residuals on other structures? \\[1pt]

\texttt{R2.M2} & Frequency-structure mismatch \emph{\texttt{under\_constrained}} & Is the model biased toward low-frequency, smooth, or global patterns while ignoring high-frequency signals, abrupt changes, periodicity, spikes, or local oscillations? \\[1pt]

\texttt{R2.M3} & Scale-structure mismatch \emph{\texttt{under\_constrained}} & Is the model good at one scale but unable to handle local details, mid-scale compositions, and global layout simultaneously? - Primarily \texttt{space}; multi-timescale problems are \texttt{time}; cross-modal scale mismatch is \texttt{modality}. \\[1pt]

\texttt{R2.M4} & Temporal-component mismatch \emph{\texttt{under\_constrained}} & Does the model struggle to handle trend, seasonality, shocks, regime switches, long memory, and short-term fluctuations at the same time? \\[1pt]

\texttt{R2.M5} & Relational-structure mismatch \emph{\texttt{under\_constrained}} & Is the model good at local texture or token patterns but weak at object relations, graph relations, entity interactions, compositional rules, or causal relations? \\[1pt]

\texttt{R2.M6} & Structural property violation \emph{\texttt{under\_constrained}} & Does the model violate a property the task requires it to preserve? - Invariance (\texttt{space}): output changes when input order, entity permutation, node indexing, or patch order changes. - Equivariance (\texttt{space}, \texttt{time}): output fails to transform correspondingly under rotation, translation, scaling, coordinate transform, time shift, or graph isomorphism. - Monotonicity (\texttt{objective}, \texttt{space}): output moves non-monotonically, unstably, or in the wrong direction with respect to key variables. - Conservation (\texttt{objective}, \texttt{time}): output violates required conservation of mass, energy, probability, flow, budget, or inventory. - Geometric / topological (\texttt{space}): invalid coordinates, broken structures, boundary misalignment, incorrect spatial relations. - Valid state space (\texttt{time}): predicted state, latent state, intermediate state, or action falls into a physically, semantically, rule-wise, or task-definition-wise invalid region. - Cross-modal legality (\texttt{modality}): outputs valid per modality but jointly invalid. \\[1pt]

\texttt{R2.M7} & Weak-constraint shortcut \emph{\texttt{under\_constrained}} & Does the model exploit input order, absolute coordinates, data-collection artifacts, format patterns, or other structurally invalid regularities to obtain superficial performance? \\[1pt]

\texttt{R2.M8} & Global vs local bias mismatch \emph{\texttt{under\_constrained}} & Does global attention, global latent representation, or a low-rank operator ignore local anomalies; or does a local operator or convolution ignore long-range dependencies and global structure? \\[1pt]

\texttt{R2.M9} & Bias fragility under regime shift \emph{\texttt{under\_constrained}} & Is the current inductive bias suited only to the dominant regime in the training distribution, so that it fails when the proportions of structural components change? \\[1pt]

\texttt{R2.M10} & Local freedom breaking global legality \emph{\texttt{under\_constrained}} & Are local modules, adapters, experts, branches, or refinements individually reasonable, but collectively violating global structural constraints? \\[1pt]

\texttt{R2.M11} & Hybrid-branch redundancy \emph{\texttt{under\_constrained}} & After combining multiple branches, bases, experts, operators, or biases, do they fail to form clear roles and instead repeatedly fit the same type of structure? \\[1pt]

\texttt{R2.M12} & Branch interference \emph{\texttt{under\_constrained}} & Do different inductive-bias branches interfere with each other, causing structural components to be misassigned, redundantly explained, or canceled out? \\[1pt]

\texttt{R2.M13} & Fusion imbalance \emph{\texttt{under\_constrained}} & Does fusion, gating, attention, residual mixing, or aggregation persistently favor one branch, preventing other structural biases from contributing? \\[1pt]

\texttt{R2.M14} & Constraint-induced underfitting \emph{\texttt{over\_constrained}} & After enforcing invariance, equivariance, monotonicity, smoothness, or conservation, does the model fail to express true local heterogeneity, symmetry breaking, or non-ideal dynamics? \\[1pt]

\texttt{R2.M15} & Approximate symmetry over-enforcement \emph{\texttt{over\_constrained}} & Is the task symmetry only approximate, so that strict constraints incorrectly erase real deviations? \\[1pt]

\texttt{R2.M16} & Constraint vs noise conflict \emph{\texttt{over\_constrained}} & Do observation noise, measurement error, label error, or simulation error make hard structural constraints too rigid? - Primarily \texttt{objective}; \texttt{space} / \texttt{time} / \texttt{modality} for where the noise lives. - Record \texttt{downstream\_of} an R8 mechanism when the noise is an annotation or data-process problem. \\[1pt]

\texttt{R2.M17} & Prior vs data conflict \emph{\texttt{over\_constrained}} & Does a physics prior, symbolic rule, domain constraint, or hand-specified dynamics pull against what the data actually shows? - Primarily \texttt{objective}. - When trained with shared parameters, record \texttt{optimization\_signature} with \texttt{gradient\_direction\_conflict}. \\[1pt]

\texttt{R2.M18} & Excessive bias complexity \emph{\texttt{over\_constrained}} & After combining too many inductive biases, does the model become difficult to tune, unstable to train, hard to interpret, or dependent on complex module interactions? \\[1pt]

\texttt{R2.M19} & Cross-modal bias incompatibility \emph{\texttt{under\_constrained} or \texttt{over\_constrained}} & Do different modalities require different scales, abstraction levels, temporal alignments, or inductive biases, but get forced into the same shared space or the same operator? \\[1pt]

\multicolumn{3}{@{}l}{\rule{0pt}{2.5ex}\textbf{R3: Information sufficiency vs compression}}\\[-1pt]

\texttt{R3.M1} & Bottleneck information loss & Does a latent bottleneck, pooling, downsampling, low-rank projection, quantization, or summary discard task-relevant detail? \\[1pt]

\texttt{R3.M2} & Excessive abstraction & Does a high-level abstract representation erase local differences, rare patterns, fine-grained classes, boundary information, or individual variation? \\[1pt]

\texttt{R3.M3} & Invariance deleting useful information & Does a factor removed for invariance actually contain useful information for certain tasks, slices, or conditions? \\[1pt]

\texttt{R3.M4} & Compression vs uncertainty & Does representation compression remove the evidence needed to estimate uncertainty, making confidence unreliable? - Primarily \texttt{objective}; \texttt{space} / \texttt{time} / \texttt{modality} for where the evidence was lost; \texttt{population} when uncertainty is lost specifically for some groups. \\[1pt]

\texttt{R3.M5} & Calibration degradation & Is accuracy normal while confidence, probability, interval, risk score, or uncertainty estimate is distorted? - Primarily \texttt{objective}; \texttt{modality} or \texttt{population} when calibration degrades on specific modalities or groups. \\[1pt]

\texttt{R3.M6} & Detail retention vs overfitting & Does retaining too much input detail cause the model to learn noise, individual-sample artifacts, spurious features, or privacy-sensitive information? \\[1pt]

\texttt{R3.M7} & Reconstruction vs prediction conflict & If the representation serves reconstruction, does it retain too much prediction-irrelevant information; if it serves prediction, does it lose information needed for reconstruction or fine-grained reasoning? \\[1pt]

\texttt{R3.M8} & Context summarization loss & After compressing long context, long sequences, historical trajectories, or multi-document input, are key facts, boundary conditions, minority evidence, or remote dependencies lost? - Primarily \texttt{time}; \texttt{space} for non-temporal token positions; \texttt{modality} for multi-document or multimodal input. \\[1pt]

\texttt{R3.M9} & Quantization / discretization damage & Does quantization, discrete tokenization, codebooks, hashing, or compact encoding damage continuous values, subtle differences, or rare-state expression? \\[1pt]

\texttt{R3.M10} & Compact representation damaging fine structure & Does low-rank factorization, compressed encoding, shared embedding, or other compact representation damage fine-grained structural expression? - Primarily \texttt{space}; \texttt{time} for fine temporal structure; \texttt{modality} for shared representations; \texttt{objective} for multi-task shared representations. \\[1pt]

\texttt{R3.M11} & Insufficient rank & Are representation dimension, latent rank, attention rank, embedding size, or memory slots insufficient to preserve all necessary factors simultaneously? \\[1pt]

\texttt{R3.M12} & Nuisance retention & Does the model fail to compress nuisance factors, leaving style, domain, speaker, background, format, or collection conditions in the representation? \\[1pt]

\texttt{R3.M13} & Insufficient sufficiency & Is an intermediate representation no longer sufficient for the downstream task, so that even a fully capable head, decoder, or policy cannot recover performance? - Primarily \texttt{objective}; \texttt{time} for intermediate sequence states; \texttt{modality} for per-modality representations; \texttt{space} for spatial encodings. \\[1pt]

\texttt{R3.M14} & Multi-objective representation demand conflict & Is the shared representation simultaneously required to retain detail, compress invariances, align modalities, reconstruct inputs, predict rewards, and satisfy constraints, creating incompatible demands? \\[1pt]

\multicolumn{3}{@{}l}{\rule{0pt}{2.5ex}\textbf{R4: Empirical fit vs robust/OOD/causal generalization}}\\[-1pt]

\texttt{R4.M1} & Shortcut learning & Does the model rely on background, format, position, templates, co-occurring words, texture, style, metadata, or collection artifacts as shortcuts? \\[1pt]

\texttt{R4.M2} & Spurious correlation & Does the model treat features that are correlated with the label in training but non-causal as primary prediction evidence? - Any of \texttt{space}, \texttt{time}, \texttt{modality}, \texttt{population}. \\[1pt]

\texttt{R4.M3} & ID accuracy masking OOD failure & Is average validation performance normal while performance under domain shift, time shift, site shift, sensor shift, style shift, population shift, or train/inference-procedure shift degrades significantly? - Primarily \texttt{time}, \texttt{modality}, \texttt{population}; \texttt{space} when the spatial distribution changes. - Includes train/inference mismatch (teacher forcing vs free running, quantization gap, prompt-distribution shift): the deployment procedure is a different distribution from the training procedure. \\[1pt]

\texttt{R4.M4} & Hidden group-performance failure & Do aggregate metrics hide systematic failures on specific groups, slices, minority classes, rare regimes, or long-tail samples? \\[1pt]

\texttt{R4.M5} & Non-robust feature reliance & Does the model use features that are imperceptible to humans, sensitive to tiny perturbations, or semantically unstable to improve standard accuracy? - Primarily \texttt{space}, \texttt{modality}; \texttt{time} for sequence perturbations. \\[1pt]

\texttt{R4.M6} & Causal feature underlearning & Are truly causal, stable, or cross-environment-valid features underlearned because their statistical signal is weaker? - Any R4 axis, depending on where the causal feature lives. \\[1pt]

\texttt{R4.M7} & Causality violation & Does the model use future information, leaked variables, posterior features, label-related shortcuts, or invalid temporal dependency? \\[1pt]

\texttt{R4.M8} & Counterfactual instability & When only non-causal factors change while core semantics remain the same, does the model output change when it should not? \\[1pt]

\texttt{R4.M9} & Augmentation mismatch & Does data augmentation, domain randomization, or robust regularization improve robustness to some perturbations while harming the original distribution or other perturbation types? \\[1pt]

\texttt{R4.M10} & Adversarial robustness vs clean accuracy & Does improving adversarial, worst-case, or perturbation robustness reduce clean accuracy or average performance? \\[1pt]

\texttt{R4.M11} & OOD calibration issue & Is the model still overconfident on OOD inputs, shifted slices, rare groups, or adversarial inputs? \\[1pt]

\texttt{R4.M12} & Worst-case vs average-case conflict & Does improving worst-group, worst-domain, or worst-case performance reduce average-case performance? \\[1pt]

\texttt{R4.M13} & Client / domain conflict & In federated or multi-domain training, do updates from different clients, domains, sites, or populations harm each other? \\[1pt]

\multicolumn{3}{@{}l}{\rule{0pt}{2.5ex}\textbf{R5: Plasticity vs retention}}\\[-1pt]

\texttt{R5.M1} & Catastrophic forgetting & After sequentially learning new tasks, domains, or data, does the model significantly forget old tasks, domains, or skills? \\[1pt]

\texttt{R5.M2} & Insufficient plasticity & After freezing parameters, limiting updates, or applying strong regularization to protect old knowledge, does the model fail to learn new tasks or distributions effectively? \\[1pt]

\texttt{R5.M3} & Shared parameter overwrite & Do updates from new data overwrite parts of the shared backbone, embeddings, memory, normalization statistics, or latent space that are important for old tasks? \\[1pt]

\texttt{R5.M4} & Representation drift & After a new training stage, do intermediate representations drift so that old heads, adapters, classifiers, decoders, or policies no longer fit? \\[1pt]

\texttt{R5.M5} & Insufficient replay / memory & Is memory, replay buffer, exemplar set, summary, or rehearsal data insufficient to preserve coverage of old distributions? \\[1pt]

\texttt{R5.M6} & New-old distribution conflict & Are new and old data distributions incompatible in label rules, input style, objective functions, user preferences, or environment dynamics? \\[1pt]

\texttt{R5.M7} & Personalization vs global generality & Does adaptation to a specific user, client, domain, or site damage the general capability of the global model? \\[1pt]

\texttt{R5.M8} & Module proliferation & To avoid forgetting, does the system keep adding task-specific adapters, experts, memory, or modules, causing parameter and management cost to grow continuously? \\[1pt]

\texttt{R5.M9} & Old constraint forgetting & After new-task updates, does the model forget old safety constraints, physical constraints, format constraints, style constraints, or behavior boundaries? \\[1pt]

\texttt{R5.M10} & Calibration drift & After continual updates, does confidence calibration drift on old domains, new domains, or mixed domains? \\[1pt]

\texttt{R5.M11} & Recency bias & Does the model over-adapt to recent data, causing long-term stable patterns or low-frequency old patterns to be overwritten? \\[1pt]

\texttt{R5.M12} & Negative backward transfer & Does learning a new task fail to help old tasks and instead reduce old-task performance? \\[1pt]

\texttt{R5.M13} & Capability tax from constraint training & After alignment, safety, fairness, privacy, or policy training, does the model degrade on certain core capabilities, open-ended tasks, creative tasks, or professional tasks? \\[1pt]

\multicolumn{3}{@{}l}{\rule{0pt}{2.5ex}\textbf{R6: Exploration vs exploitation}}\\[-1pt]

\texttt{R6.M1} & Insufficient exploration & Does the system converge too early to a small number of modes, actions, paths, candidates, experts, or solution-space regions? \\[1pt]

\texttt{R6.M2} & Insufficient diversity & Are outputs, generated samples, candidate solutions, trajectories, retrieved items, or action sequences highly repetitive, homogeneous, or mode-collapsed? - Primarily \texttt{objective}; \texttt{time} for the decoding process; \texttt{module} for expert-level collapse. \\[1pt]

\texttt{R6.M3} & Insufficient coverage & Does the model ignore long-tail modes, rare solutions, low-frequency classes, minority strategies, rare states, or non-mainstream candidates? \\[1pt]

\texttt{R6.M4} & Excessive exploration & After increasing randomness, diversity, candidate breadth, or novelty, do outputs become less stable, lower quality, or less aligned with the task goal? \\[1pt]

\texttt{R6.M5} & Fidelity vs diversity conflict & Does increasing diversity, temperature, entropy, search breadth, or novelty reduce factuality, coherence, precision, reward, or task success? \\[1pt]

\texttt{R6.M6} & Excessive exploitation & Does the model over-select current high-score, high-probability, high-reward, or high-confidence candidates, missing potentially better candidates with lower initial scores? \\[1pt]

\texttt{R6.M7} & Search breadth vs depth & Under a fixed budget, does increasing candidate count reduce evaluation depth, rollout length, verification strength, or refinement quality per candidate? \\[1pt]

\texttt{R6.M8} & Beam / top-k degeneration & Do beam search, top-k, top-p, reranking, or greedy selection produce repetition, templating, local optima, or lack of novelty? \\[1pt]

\texttt{R6.M9} & Novelty vs constraint conflict & When pursuing novelty, coverage, or diversity, does the model become more likely to violate structural constraints, task constraints, safety constraints, or real-world feasibility? - Primarily \texttt{objective}; \texttt{module} when driven by expert routing; \texttt{time} for the planning process. - Record \texttt{upstream\_of} R2.M6 (structural) or R7 (normative) as appropriate. \\[1pt]

\texttt{R6.M10} & Rare mode vs common quality & Does improving rare-mode coverage damage common-case quality, average performance, or mainstream user experience? \\[1pt]

\texttt{R6.M11} & Exploration target misidentification & Does the system mistake noisy samples, outliers, unlearnable samples, or annotation-unstable samples for high-value exploration targets? \\[1pt]

\multicolumn{3}{@{}l}{\rule{0pt}{2.5ex}\textbf{R7: Normative constraints vs utility}}\\[-1pt]

\texttt{R7a.M1} & Fairness criteria incompatibility & Are different fairness criteria - calibration, equalized odds, demographic parity, individual fairness, subgroup fairness - provably impossible to satisfy simultaneously under the given base rates? \\[1pt]

\texttt{R7a.M2} & Preference aggregation impossibility & Do different populations, users, annotators, policy sources, or preference models disagree in a way that admits no consistent aggregate ordering? \\[1pt]

\texttt{R7a.M3} & Privacy-utility lower bound & Is the required privacy guarantee (e.g.~a given epsilon) mathematically incompatible with the required accuracy at the available sample size? \\[1pt]

\texttt{R7b.M1} & Privacy vs accuracy & Does differential privacy, noise injection, gradient clipping, data minimization, redaction, or anonymization reduce accuracy or fine-grained performance? \\[1pt]

\texttt{R7b.M2} & Privacy vs personalization & After reducing user-data retention or limiting personalized memory, does the model struggle to provide long-term consistent, personalized, or context-relevant service? \\[1pt]

\texttt{R7b.M3} & Fairness vs average performance & Does improving performance for certain groups, minority slices, or worst groups reduce overall average performance? \\[1pt]

\texttt{R7b.M4} & Safety vs helpfulness & Do safety constraints cause overly conservative answers, insufficient information, or failure to complete otherwise reasonable tasks? \\[1pt]

\texttt{R7b.M5} & Harmlessness vs truthfulness & To avoid potential risk, does the model weaken, blur, evade, or distort information that should be provided accurately? \\[1pt]

\texttt{R7b.M6} & Helpfulness vs policy compliance & Does the user's requested task goal conflict with policy, legal, ethical, safety, or deployment constraints? \\[1pt]

\texttt{R7b.M7} & Constraint stacking & Do privacy, fairness, safety, truthfulness, helpfulness, legal compliance, and business rules conflict with each other or jointly compress the usable output space below what the task requires? \\[1pt]

\texttt{R7c.M1} & Over-refusal & Does the model extend safety boundaries to safe requests, causing benign requests to be incorrectly refused? \\[1pt]

\texttt{R7c.M2} & Under-refusal & Does the model still provide assistance on high-risk, violating, misleading, or unauthorized requests where it should not? \\[1pt]

\texttt{R7c.M3} & Uneven boundary generalization & Does the policy behave inconsistently across languages, domains, user groups, expression styles, context lengths, or modalities? \\[1pt]

\multicolumn{3}{@{}l}{\rule{0pt}{2.5ex}\textbf{R8: Data/annotation conflict}}\\[-1pt]

\texttt{R8.M1} & Annotator disagreement & Do different annotators, raters, preference sources, or labeling passes assign different labels to the same or near-identical inputs, so that no single target is recoverable? \\[1pt]

\texttt{R8.M2} & Label definition ambiguity & Is the class boundary, rating scale, or target definition itself underspecified, so that the ceiling is set by the definition rather than by the model? \\[1pt]

\texttt{R8.M3} & Noise vs true long tail indistinguishability & Can the training signal not separate mislabeled samples from rare-but-real patterns, so that noise-robust training also suppresses genuine long-tail structure? \\[1pt]

\texttt{R8.M4} & Sparse / uneven supervision & Are local labels sparse, supervision uneven, or long-tail regions under-sampled, so that added local freedom yields unstable predictions? \\[1pt]

\texttt{R8.M5} & Label process drift & Have labeling conventions, guidelines, annotator pools, or rating rubrics shifted over time, so that ``old'' and ``new'' labels encode different functions? \\[1pt]

\texttt{R8.M6} & Label leakage & Does the annotation process itself introduce features correlated with the label that will not exist at deployment (annotation order, tooling artifacts, reviewer identity)? \\[1pt]

\texttt{R8.M7} & Aggregation loss & Does collapsing multiple annotations into a single hard label (majority vote, mean rating) discard the disagreement signal that carries real information about ambiguity or subpopulation difference? \\[1pt]

\multicolumn{3}{@{}l}{\rule{0pt}{2.5ex}\textbf{R9: Evaluation--objective mismatch}}\\[-1pt]

\texttt{R9.M1} & Proxy-goal divergence & Does optimization pressure on a visible reward, score, ranker, or proxy metric drive the system away from the true objective while the proxy continues to improve? (reward hacking) \\[1pt]

\texttt{R9.M2} & Metric insensitivity & Is the metric structurally incapable of registering the failure mode that matters - averaging it away, not sampling it, or scoring it as correct? \\[1pt]

\texttt{R9.M3} & Benchmark artifact overfitting & Does the model overfit benchmark format, annotation style, sampling bias, data-cleaning procedure, or evaluation protocol rather than the underlying capability? \\[1pt]

\texttt{R9.M4} & Benchmark saturation & Has the metric ceased to discriminate - near-ceiling scores that no longer track real capability differences? \\[1pt]

\texttt{R9.M5} & Validation-deployment gap & Does the evaluation distribution, interaction pattern, or scoring procedure differ from deployment in ways that make validation results non-transferable? \\[1pt]

\texttt{R9.M6} & Judge / reward-model unreliability & Is the evaluator itself (an LLM judge, learned reward model, or heuristic scorer) biased, gameable, or inconsistent, so that its scores do not support the conclusions drawn from them? \\[1pt]

\texttt{R9.M7} & Aggregate masking & Does a single headline number hide the trade-off structure underneath, so that a metric-neutral change is reported as no change when it is in fact a redistribution? \\[1pt]

\end{longtable}
\endgroup

\subsection{Optimization artifacts and boundary decisions}
\label{app:taxonomy-boundaries}

\begin{table}[t]
\centering
\small
\caption{Standalone optimization artifacts, used only when the conflict
disappears with independent parameters.}
\label{tab:taxonomy-opt}
\begin{tabular}{@{}p{0.11\linewidth}p{0.25\linewidth}p{0.56\linewidth}@{}}
\toprule
ID & Artifact & Diagnostic definition \\
\midrule
\texttt{OPT.M1} & Gradient scale imbalance & Gradients from some objectives are persistently larger, so shared training primarily serves them although the substantive demands are compatible. \\
\texttt{OPT.M2} & Convergence rhythm conflict & Objectives converge at different speeds, so later updates damage early-converged objectives or slow objectives drag down shared training. \\
\texttt{OPT.M3} & Weighted-sum masking & Aggregation in the training objective hides degradation of individual objectives while the weighted sum appears normal. \\
\bottomrule
\end{tabular}
\end{table}

The taxonomy applies the following consolidation boundaries.  First, R1
captures a budgetary capacity limit, whereas R2 captures prior--signal mismatch
and R3 captures information loss that additional compute alone cannot repair.
Second, R4 describes model failure under shift; R8 captures upstream label or
annotation failure; and R9 captures failure of the measurement apparatus.
Third, sequential overwrite belongs to R5, while simultaneous interference is
assigned to the relevant substantive Root and annotated with an optimization
signature.  Fourth, R6 covers exploration of solutions or actions, whereas
population coverage is routed to R4 or R7.  Fifth, R7a represents an
incompatibility that has no jointly satisfying solution, R7b represents a real
normative--utility frontier, and R7c concerns calibration of a decision
boundary.  Finally, \texttt{OPT.M3} concerns aggregation by the training
objective, while R9.M7 concerns aggregation by the evaluation metric; both may
hold simultaneously.

The selection stage outputs taxonomy assignments and their evidence only.  It
does not design probes, controlled faults, experiments, repairs, training
strategies, or code modifications.  These operations are handled by subsequent
ConflictGuide-Skill stages.  The active inventory contains 110
\texttt{R*.M*} mechanisms and three \texttt{OPT.M*} artifacts; no historical ID
migration table is part of the fixed library.

\section{Identifiability of Conflict-Alleviating Edits}
\label{app:identifiability}

We formalize the information provided by probes at a fixed code state and
candidate evaluation. Here, \emph{identifiability} refers to the ability to
infer whether an edit alleviates a conflict from the feedback available to
the agent. The analysis concerns feedback information; it does not assume
that the coding agent uses that information optimally.

\paragraph{Behavioral target and observations.}
For analysis, let $b_A(x)$ and $b_B(x)$ denote population-level scores for
the desirable behaviors $B_A$ and $B_B$, respectively, with larger values
indicating better behavior. Define the candidate's true behavior changes
by
\begin{equation}
\boldsymbol{\Delta}_{c,t}^{B}
=
\begin{bmatrix}
b_A(x_t^{e_t})-b_A(x_t) \\
b_B(x_t^{e_t})-b_B(x_t)
\end{bmatrix}.
\end{equation}
Let $\mathcal R_c\subseteq\mathbb R^2$ be a fixed set of behavior changes
judged to alleviate conflict $c$, and define
\begin{equation}
Y_t
=
\mathbb I\!\left[
\boldsymbol{\Delta}_{c,t}^{B}\in\mathcal R_c
\right].
\label{eq:true_alleviation_event}
\end{equation}
For example, $\mathcal R_c$ could require both behaviors to be
non-decreasing and at least one to improve. This population-level event
is the target of inference; it is distinct from the operational criterion
$A_c(e_t)>\tau_c$, which is computed from probes using the model-specific
reference and guard conditions.

Let $\mathcal W_t$ contain the information common to the two feedback
settings before the candidate's evaluation, including $x_t$, $e_t$, and
the shared search context. The registered retention rule and its fixed
settings $\tau_S$, $\tau_c$, and $\mathcal B_c$ are also included in
$\mathcal W_t$. The scalar and probe-augmented information sets for the
\emph{same} candidate are
\begin{equation}
\mathcal F_t^{S}
=
\mathcal W_t\vee\sigma(G_t),
\qquad
\mathcal F_t^{S,Z}
=
\mathcal F_t^{S}
\vee\sigma\!\left(\mathbf G_{c,t}^{Z},A_c(e_t)\right),
\label{eq:feedback_information_sets}
\end{equation}
where $\vee$ joins information sets and $\sigma(\cdot)$ denotes the
information generated by an observation. The score $A_c(e_t)$ is
computable from the probe feedback, including the required reference
values and guard measurements. Probabilities below represent uncertainty
about behavioral effects conditional on the information available to
the agent.

\paragraph{Why scalar gain can be ambiguous.}
Consider an illustrative setting in which the task gain locally equals
the sum of the two behavior changes. For any $u>0$, the changes
\begin{equation}
(2u,-u)
\qquad\text{and}\qquad
(u/2,u/2)
\label{eq:scalar_collision_example}
\end{equation}
both yield $G_t=u$. The first harms $B_B$, whereas the second improves
both behaviors. Thus, the scalar gain alone does not determine whether
the edit jointly improves the behaviors. Probes that distinguish the
two changes can resolve this ambiguity.
Eq.~(\ref{eq:scalar_collision_example}) illustrates a possibility
under scalar aggregation; it is not an assumption about the task
metrics in our experiments.

\paragraph{Proposition 1: conditional information gain.}
Let
\begin{equation}
p_t^{S}
=
\mathbb P(Y_t=1\mid\mathcal F_t^{S}),
\qquad
p_t^{S,Z}
=
\mathbb P(Y_t=1\mid\mathcal F_t^{S,Z}).
\end{equation}
Suppose that, on a set of positive probability,
$0<p_t^{S}<1$ and the conditional distribution of
$\mathbf G_{c,t}^{Z}$ differs between $Y_t=1$ and $Y_t=0$
after conditioning on $\mathcal F_t^{S}$. Then observing probes strictly
reduces the optimal squared-error risk for predicting $Y_t$:
\begin{equation}
\begin{aligned}
R_S-R_{S,Z}
&=
\mathbb E\!\left[
    (p_t^{S,Z}-p_t^{S})^2
\right]
>0,\\
R_S
&=
\inf_{\widehat p\text{ measurable w.r.t. }\mathcal F_t^{S}}
\mathbb E[(Y_t-\widehat p)^2],\\
R_{S,Z}
&=
\inf_{\widehat p\text{ measurable w.r.t. }\mathcal F_t^{S,Z}}
\mathbb E[(Y_t-\widehat p)^2].
\end{aligned}
\label{eq:probe_bayes_risk}
\end{equation}
The reduction in optimal logarithmic-loss risk is likewise
\begin{equation}
H(Y_t\mid\mathcal F_t^{S})
-
H(Y_t\mid\mathcal F_t^{S,Z})
=
\mathbb E\!\left[
D_{\mathrm{KL}}\!\left(
\operatorname{Bern}(p_t^{S,Z})
\,\Vert\,
\operatorname{Bern}(p_t^{S})
\right)
\right]
>0.
\label{eq:probe_conditional_information}
\end{equation}

\emph{Proof.}
The optimal squared-error predictor under an information set is its
conditional expectation. Because
$\mathcal F_t^{S}\subseteq\mathcal F_t^{S,Z}$, the tower property gives
\begin{equation}
\mathbb E[p_t^{S,Z}\mid\mathcal F_t^{S}]
=
p_t^{S}.
\end{equation}
Expanding
$Y_t-p_t^{S}
=(Y_t-p_t^{S,Z})+(p_t^{S,Z}-p_t^{S})$ and taking expectations,
the cross term vanishes by conditional expectation. Hence
\begin{equation}
\mathbb E[(Y_t-p_t^{S})^2]
=
\mathbb E[(Y_t-p_t^{S,Z})^2]
+
\mathbb E[(p_t^{S,Z}-p_t^{S})^2],
\end{equation}
which proves the equality in
Eq.~(\ref{eq:probe_bayes_risk}).

To establish strictness, suppose instead that
$p_t^{S,Z}=p_t^{S}$ almost surely. For any measurable set $D$ of probe
values, conditional expectation would then give
\begin{equation}
\begin{aligned}
&\mathbb P\!\left(
Y_t=1,\,
\mathbf G_{c,t}^{Z}\in D
\mid\mathcal F_t^{S}
\right)\\
&\qquad=
p_t^{S}\,
\mathbb P\!\left(
\mathbf G_{c,t}^{Z}\in D
\mid\mathcal F_t^{S}
\right).
\end{aligned}
\end{equation}
Where $0<p_t^{S}<1$, this implies equal conditional probe
distributions for $Y_t=1$ and $Y_t=0$, contradicting the assumption.
Therefore $p_t^{S,Z}\neq p_t^{S}$ with positive probability, making the
squared-error reduction strictly positive. Finally, the standard
conditional-entropy identity yields
Eq.~(\ref{eq:probe_conditional_information}); its KL divergence is
strictly positive wherever the two posteriors differ.
\hfill$\square$

The proposition requires probes to contain information about true
conflict alleviation beyond the scalar gain and shared context. Probe
qualification provides empirical grounds for using the measurements,
but does not by itself prove this assumption for every candidate. A
strict reduction in squared or logarithmic loss also need not produce
a strict reduction in binary classification error: the posterior may
change without crossing a decision threshold.

\paragraph{Identifiability of the operational retention decision.}
The retention rule in Eq.~(\ref{eq:retention}) makes a separate,
directly observable use of probes. Define
\begin{equation}
M_t=\mathbb I[G_t\in\mathcal B_c],
\qquad
Q_t=\mathbb I[A_c(e_t)>\tau_c],
\qquad
q_t=\mathbb P(Q_t=1\mid\mathcal F_t^{S}).
\end{equation}
Since $\mathcal B_c\subseteq[0,\tau_S]$, the two branches of
Eq.~(\ref{eq:retention}) are disjoint:
\begin{equation}
\operatorname{Keep}(e_t)
=
\mathbb I[G_t>\tau_S]+M_tQ_t.
\label{eq:retention_decomposition}
\end{equation}
For binary retention decisions based only on $\mathcal F_t^{S}$,
the smallest possible probability of disagreeing with the operational
rule is
\begin{equation}
\inf_{\delta\text{ measurable w.r.t. }\mathcal F_t^{S}}
\mathbb P\!\left[
\delta\neq\operatorname{Keep}(e_t)
\right]
=
\mathbb E\!\left[
M_t\min\{q_t,1-q_t\}
\right].
\label{eq:scalar_retention_ambiguity}
\end{equation}
When $M_t=0$, the rule is determined by $G_t$.
When $M_t=1$, a scalar-only decision must predict the unobserved
binary value $Q_t$, whose optimal conditional error is
$\min\{q_t,1-q_t\}$. In contrast, $Q_t$ and
$\operatorname{Keep}(e_t)$ are measurable with respect to
$\mathcal F_t^{S,Z}$. Thus, if
$\mathbb P(M_t=1,\;0<q_t<1)>0$, scalar feedback cannot exactly
reproduce the probe-based retention decision, whereas probe feedback
can. This statement concerns the specified operational rule; its
agreement with the population-level event $Y_t$ depends on probe
validity.

For a common sequence of evaluated edits, the Stage-II history
$\mathcal H_t^{S,Z}$ refines the corresponding scalar history
$\mathcal H_t^{S}$ by recording the probe changes of earlier edits.
Proposition 1 therefore gives a conditional information rationale for
using that history to guide subsequent proposals. It does not imply
that every probe is informative, that a coding agent necessarily uses
the additional information well, or that information gain alone
guarantees improved task performance.

\section{Model-Specific Conflicts and Probe Metrics}
\label{app:model_probes}

This appendix instantiates the model-specific conflicts summarized in
Table~\ref{tab:conflict-instances} and gives the complete definitions of their
Probe metrics.  Each conflict is represented by two desirable behaviors,
$B_A$ and $B_B$, coupled through a shared model mechanism $M_c$.  The Probes
are designed to expose changes in these behaviors that may be hidden by the
scalar task metric $S$.

This section focuses on the conflict definitions and measurement procedures.
Dataset splits, training schedules, evolution budgets, calibration thresholds,
and retention rules are reported separately with the experimental
configuration.  Unless stated otherwise, all Probe construction choices,
including masks, sample pairs, graph perturbations, filters, scales, anchors,
and random seeds, are frozen before conflict-aware evolution.  Test data are
not used to construct or calibrate the Probes.

\subsection{Common Notation and Reporting Protocol}
\label{app:probe_common}

Let $f_\theta$ denote a trained candidate model with parameters $\theta$.
We use $\mathcal T$, $\mathcal C$, and $\mathcal V$ for the frozen training,
Probe-construction or calibration, and keep-validation sets, respectively.
The official test set is excluded from all Probe construction and online
evaluation.

A model-specific conflict is written as
\begin{equation}
c=(B_A,B_B,M_c),
\end{equation}
where $B_A$ and $B_B$ are desirable behaviors and $M_c$ is the mechanism
through which they interact.  Its Probe vector is
\begin{equation}
\mathbf Z_c(\theta)
=
\bigl[Z_1(\theta),\ldots,Z_m(\theta)\bigr].
\end{equation}
The desirable direction of every component is stated explicitly using
$\uparrow$ or $\downarrow$.  Probe components are exposed separately rather
than combined with $S$ through a model-dependent weighted sum.

Each model-specific subsection identifies the taxonomy path, competing
behaviors, and shared mechanism. It defines the scalar task metric $S$,
the conflict-specific Probe metrics, and any model-specific guard metrics.
For each metric, the formulas and accompanying text specify the measured
quantities, desirable direction, and interpretation. The subsection also
identifies construction choices held fixed across candidates and explains
how the search-time measurements relate to formal evaluation.

Probe qualification is specified in the experimental protocol. It assesses
determinism, whether observed Probe changes exceed no-edit evaluation noise,
and whether the Probes distinguish behavioral effects among candidates with
similar task performance. Numerical qualification thresholds and retention
rules are given in Appendix~\ref{app:experimental_details}.

\subsection{SpecB--FNO}
\label{app:probe_specb_fno}

\paragraph{Conflict instantiation.}

SpecB--FNO performs autoregressive operator learning for two-dimensional
incompressible Navier--Stokes dynamics. Its conflict instantiates
\textbf{R2.M2}, frequency-structure mismatch, on the \texttt{space} axis.
The competing behaviors are
\begin{align}
B_A &: \text{accurate prediction of dominant, energy-carrying modes},
\\
B_B &: \text{accurate prediction of non-dominant spectral structure}.
\end{align}
Both behaviors depend on the shared spectral transformations,
residual-correction path, and fusion of the spectral and spatial branches.
Edits to these components therefore need not benefit the two behaviors
equally.

Although full-field error depends on all frequencies, it tends to be driven
by modes carrying most of the target energy. A candidate may consequently
improve the scalar task metric by reducing dominant-mode error while leaving
lower-energy spectral structure underfit. We use \(Z_{\mathrm{ND}}\) to
measure this non-dominant behavior separately.

\paragraph{Scalar task metric.}

Let \(x\) denote a candidate implementation. For sample \(n\) and rollout
step \(t\), let \(\mathbf y_{n,t}\in\mathbb R^{H\times W}\) and
\(\widehat{\mathbf y}^{(x)}_{n,t}\in\mathbb R^{H\times W}\) denote the
target and predicted vorticity fields in physical scale. The scalar task
metric is the mean per-sample, per-step full-field NRMSE:
\begin{equation}
\label{eq:specb-task}
S(x)
=
S_{\mathrm{NRMSE}}(x)
=
\frac{1}{|\mathcal V_{\mathrm{keep}}|T}
\sum_{n\in\mathcal V_{\mathrm{keep}}}
\sum_{t=1}^{T}
\frac{
\left\|
\widehat{\mathbf y}^{(x)}_{n,t}
-
\mathbf y_{n,t}
\right\|_2
}{
\left\|\mathbf y_{n,t}\right\|_2
+
\epsilon_{\mathrm{field}}
},
\qquad
S(x)\downarrow ,
\end{equation}
where \(\mathcal V_{\mathrm{keep}}\) is the fixed keep-validation set,
\(T\) is the rollout length, and
\(\epsilon_{\mathrm{field}}>0\) is a numerical-stability constant. The norm
is taken over the spatial grid. This metric measures overall predictive
fidelity but does not isolate error in non-dominant spectral modes.

\paragraph{Fourier representation and spectral partition.}

Let \(\mathcal F_{\mathrm r}\) denote the orthonormal real-valued
two-dimensional Fourier transform. The target coefficients and prediction
residual coefficients are
\begin{align}
Y_{n,t,k}
&=
\mathcal F_{\mathrm r}
\left(
\mathbf y_{n,t}
\right)_k,
\\
R^{(x)}_{n,t,k}
&=
\mathcal F_{\mathrm r}
\left(
\widehat{\mathbf y}^{(x)}_{n,t}
-
\mathbf y_{n,t}
\right)_k,
\end{align}
where \(k\) indexes a Fourier coefficient. Let \(\mathcal K\) denote the
fixed set of Fourier modes represented by the spectral branch, and let
\(w_k\) denote the corresponding real-FFT Parseval weight.

The dominant/non-dominant partition is constructed once using only targets
from a calibration set \(\mathcal C\), which is disjoint from
\(\mathcal V_{\mathrm{keep}}\). For each \(k\in\mathcal K\), define its
aggregate calibration energy as
\begin{equation}
\label{eq:specb-cal-energy}
\mathcal E^{\mathrm{cal}}_k
=
\sum_{n\in\mathcal C}
\sum_{t=1}^{T}
w_k
\left|Y_{n,t,k}\right|^2.
\end{equation}
Let \(\pi\) order the retained modes by decreasing calibration energy:
\[
\mathcal E^{\mathrm{cal}}_{\pi_1}
\ge
\mathcal E^{\mathrm{cal}}_{\pi_2}
\ge
\cdots .
\]
Using a fixed cumulative-energy threshold \(\rho_E\), define
\begin{equation}
\label{eq:specb-cutoff}
r^\star
=
\min\left\{
r:
\frac{
\sum_{j=1}^{r}
\mathcal E^{\mathrm{cal}}_{\pi_j}
}{
\sum_{k\in\mathcal K}
\mathcal E^{\mathrm{cal}}_k
}
\ge
\rho_E
\right\}.
\end{equation}
The dominant and non-dominant spectral sets are
\begin{equation}
\label{eq:specb-mode-sets}
\mathcal D
=
\{\pi_1,\ldots,\pi_{r^\star}\},
\qquad
\mathcal N
=
\mathcal K\setminus\mathcal D.
\end{equation}
The resulting partition is fixed across all candidates. Candidate
predictions are never used to construct it.

\paragraph{Primary probe.}

The primary probe is the normalized residual energy over the non-dominant
spectral set:
\begin{equation}
\label{eq:specb-znd}
Z_{\mathrm{ND}}(x)
=
\frac{
\displaystyle
\sum_{n\in\mathcal V_{\mathrm{keep}}}
\sum_{t=1}^{T}
\sum_{k\in\mathcal N}
w_k
\left|R^{(x)}_{n,t,k}\right|^2
}{
\displaystyle
\sum_{n\in\mathcal V_{\mathrm{keep}}}
\sum_{t=1}^{T}
\sum_{k\in\mathcal N}
w_k
\left|Y_{n,t,k}\right|^2
+
\epsilon_{\mathrm{spec}}
},
\qquad
Z_{\mathrm{ND}}(x)\downarrow ,
\end{equation}
where \(\epsilon_{\mathrm{spec}}>0\) is a numerical-stability constant.
The numerator and denominator pool spectral energy over the same
keep-validation samples, rollout steps, and frozen non-dominant modes.
Thus, \(Z_{\mathrm{ND}}\) measures residual energy relative to target energy
within the representable non-dominant spectral structure.

\paragraph{Guard metrics.}

We use two auxiliary metrics to detect whether an improvement in
\(Z_{\mathrm{ND}}\) is accompanied by worse dominant-mode fidelity or
late-rollout accuracy. The dominant-mode normalized spectral error is
defined analogously to \(Z_{\mathrm{ND}}\), with \(\mathcal N\) replaced by
\(\mathcal D\):
\begin{equation}
\label{eq:specb-zd}
Z_{\mathrm D}(x)
=
\frac{
\displaystyle
\sum_{n\in\mathcal V_{\mathrm{keep}}}
\sum_{t=1}^{T}
\sum_{k\in\mathcal D}
w_k
\left|R^{(x)}_{n,t,k}\right|^2
}{
\displaystyle
\sum_{n\in\mathcal V_{\mathrm{keep}}}
\sum_{t=1}^{T}
\sum_{k\in\mathcal D}
w_k
\left|Y_{n,t,k}\right|^2
+
\epsilon_{\mathrm{spec}}
},
\qquad
Z_{\mathrm D}(x)\downarrow .
\end{equation}

Let \(\mathcal T_{\mathrm{late}}\subseteq\{1,\ldots,T\}\) denote the fixed
set of late rollout steps. The late-rollout relative error is
\begin{equation}
\label{eq:specb-zlate}
Z_{\mathrm{late}}(x)
=
\frac{1}{
|\mathcal V_{\mathrm{keep}}|
|\mathcal T_{\mathrm{late}}|
}
\sum_{n\in\mathcal V_{\mathrm{keep}}}
\sum_{t\in\mathcal T_{\mathrm{late}}}
\frac{
\left\|
\widehat{\mathbf y}^{(x)}_{n,t}
-
\mathbf y_{n,t}
\right\|_2
}{
\left\|\mathbf y_{n,t}\right\|_2
+
\epsilon_{\mathrm{field}}
},
\qquad
Z_{\mathrm{late}}(x)\downarrow .
\end{equation}

Here, \(Z_{\mathrm{ND}}\) is the primary conflict-specific probe, whereas
\(Z_{\mathrm D}\) and \(Z_{\mathrm{late}}\) guard against degraded
dominant-mode fidelity and late-rollout accuracy, respectively. These
quantities are not combined with \(S(x)\) through a weighted scalar
objective. Their roles in candidate retention are specified by the common
evolution protocol.

\paragraph{Relation to formal evaluation.}

Search-time \(Z_{\mathrm{ND}}\) and formal ND-NMSE use the same functional
definition and measure the same non-dominant spectral behavior. Their
spectral partitions are constructed and frozen separately within the
search-time and held-out evaluation protocols. Full-field NRMSE remains the
primary task metric, while \(Z_{\mathrm D}\) and \(Z_{\mathrm{late}}\) are
used only as search-time guards. The retained Fourier set, Parseval
convention, cumulative-energy threshold, rollout configuration, and
evaluation splits are specified in
Appendix~\ref{app:experimental_details}.

\subsection{SNGP}
\label{app:probe_sngp}

\paragraph{Conflict instantiation.}

SNGP combines a discriminative feature extractor with random Fourier
features (RFFs), a Gaussian-process output layer, and mean-field predictive
correction. Its conflict instantiates \textbf{R3.M4}, compression versus
uncertainty, on the \texttt{objective} axis. The competing behaviors are
\begin{align}
B_A &: \text{discriminative class separation},
\\
B_B &: \text{preservation of input-relative geometry in the pre-GP
representation}.
\end{align}
They interact through the shared representation consumed by the RFF--GP
head. Strong discriminative compression can improve class separation and
predictive fit while non-uniformly distorting the local geometry used by the
GP layer to express distance-aware uncertainty. Conversely, preserving this
geometry can constrain aggressive discriminative compression.

We operationalize the two behaviors with
\begin{equation}
\label{eq:sngp-probe-vector}
\mathbf Z_{\mathrm{SNGP}}(x)
=
\left[
Z_{\mathrm{mar}}(x),
Z_{\mathrm{dist}}(x)
\right],
\qquad
Z_{\mathrm{mar}}\uparrow,
\quad
Z_{\mathrm{dist}}\downarrow,
\end{equation}
where $x$ denotes a candidate implementation.
$Z_{\mathrm{mar}}$ measures vulnerable-tail class separation, whereas
$Z_{\mathrm{dist}}$ measures non-uniform distortion of input-relative
geometry in the pre-GP representation. The latter is a mechanism-aligned
proxy for the geometry supporting distance-aware uncertainty, rather than a
direct estimate of predictive uncertainty.

\paragraph{Scalar task metric.}

Let $\boldsymbol\ell_i^{(x)}\in\mathbb R^C$ denote the
mean-field-corrected logits produced by candidate $x$ for example $i$.
A candidate-specific temperature is estimated on a calibration split:
\begin{equation}
\label{eq:sngp-temperature}
\widehat T_x
=
\underset{T>0}{\arg\min}
\left[
-\frac{1}{|\mathcal C_{\mathrm{temp}}|}
\sum_{i\in\mathcal C_{\mathrm{temp}}}
\log
\operatorname{softmax}
\left(
\frac{\boldsymbol\ell_i^{(x)}}{T}
\right)_{y_i}
\right].
\end{equation}
The scalar task metric is the temperature-scaled negative log-likelihood
evaluated on the keep-validation split:
\begin{equation}
\label{eq:sngp-task}
S(x)
=
-\frac{1}{|\mathcal V_{\mathrm{keep}}|}
\sum_{i\in\mathcal V_{\mathrm{keep}}}
\log
\operatorname{softmax}
\left(
\frac{\boldsymbol\ell_i^{(x)}}{\widehat T_x}
\right)_{y_i},
\qquad
S(x)\downarrow.
\end{equation}
The calibration and keep-validation splits are disjoint. This scalar measures
aggregate predictive fit, but does not reveal whether an improvement results
from stronger class separation, better-preserved geometry, or a trade-off
between the two.

\paragraph{Probe 1: vulnerable-tail centroid margin.}

Let $\mathbf z_i^{(x)}$ denote the deterministic representation immediately
before the GP classifier. We first normalize each representation:
\begin{equation}
\label{eq:sngp-normalized-feature}
\widetilde{\mathbf z}_i^{(x)}
=
\frac{\mathbf z_i^{(x)}}
{\max\{\|\mathbf z_i^{(x)}\|_2,\epsilon_z\}},
\end{equation}
where $\epsilon_z>0$ is a fixed numerical-stability constant. For each class
$c$, a candidate-specific normalized centroid is computed from the frozen
training indices $\mathcal T_c$:
\begin{align}
\overline{\boldsymbol\mu}_c^{(x)}
&=
\frac{1}{|\mathcal T_c|}
\sum_{j\in\mathcal T_c}
\widetilde{\mathbf z}_j^{(x)},
\\
\widetilde{\boldsymbol\mu}_c^{(x)}
&=
\frac{\overline{\boldsymbol\mu}_c^{(x)}}
{\max\{\|\overline{\boldsymbol\mu}_c^{(x)}\|_2,\epsilon_z\}}.
\label{eq:sngp-centroid}
\end{align}
For each keep-validation example, its correct-class versus
nearest-wrong-class cosine margin is
\begin{equation}
\label{eq:sngp-example-margin}
m_i^{(x)}
=
\widetilde{\mathbf z}_i^{(x)\top}
\widetilde{\boldsymbol\mu}_{y_i}^{(x)}
-
\max_{c\neq y_i}
\widetilde{\mathbf z}_i^{(x)\top}
\widetilde{\boldsymbol\mu}_c^{(x)}.
\end{equation}
The probe summarizes the lower tail of these margins:
\begin{equation}
\label{eq:sngp-margin}
Z_{\mathrm{mar}}(x)
=
Q_{q_{\mathrm{mar}}}
\left(
\{m_i^{(x)}:i\in\mathcal V_{\mathrm{keep}}\}
\right),
\qquad
Z_{\mathrm{mar}}(x)\uparrow,
\end{equation}
where $q_{\mathrm{mar}}$ is a fixed lower-quantile level. This probe emphasizes
examples with weak class separation rather than average separation.

\paragraph{Probe 2: input--feature distance distortion.}

Let $\mathbf u_i\in\mathbb R^{D_u}$ denote the fixed preprocessed input, and
define the normalized input-space distance
\begin{equation}
d_u(i,j)
=
\frac{\|\mathbf u_i-\mathbf u_j\|_2}{\sqrt{D_u}}.
\end{equation}
Before evolution, the evaluator constructs fixed same-class and cross-class
pair panels,
\begin{equation}
\mathcal P_{\mathrm{same}}
\quad\text{and}\quad
\mathcal P_{\mathrm{cross}},
\end{equation}
from neighborhoods in the input space. These panels depend only on the
frozen data and are reused for every candidate.

For each frozen pair $(i,j)$, its distance in the candidate-dependent
pre-GP representation is
\begin{equation}
\label{eq:sngp-feature-distance}
d_z^{(x)}(i,j)
=
\frac{\|\mathbf z_i^{(x)}-\mathbf z_j^{(x)}\|_2}{\sqrt{D_z}},
\end{equation}
and its log distance ratio is
\begin{equation}
\label{eq:sngp-log-ratio}
r_{ij}^{(x)}
=
\log
\frac{\max\{d_z^{(x)}(i,j),\epsilon_d\}}
{\max\{d_u(i,j),\epsilon_d\}},
\end{equation}
where $\epsilon_d>0$ is fixed. A candidate-specific shared median,
\begin{equation}
\label{eq:sngp-shared-median}
\widetilde r^{(x)}
=
\operatorname{median}
\left(
\left\{
r_{ij}^{(x)}
:
(i,j)\in
\mathcal P_{\mathrm{same}}
\cup
\mathcal P_{\mathrm{cross}}
\right\}
\right),
\end{equation}
removes a common log-distance shift and therefore makes the probe insensitive
to uniform global rescaling of the representation.

For each pair group $g\in\{\mathrm{same},\mathrm{cross}\}$, define
\begin{equation}
D_g^{(x)}
=
Q_{q_{\mathrm{dist}}}
\left(
\left\{
\left|r_{ij}^{(x)}-\widetilde r^{(x)}\right|
:
(i,j)\in\mathcal P_g
\right\}
\right),
\end{equation}
where $q_{\mathrm{dist}}$ is a fixed upper-quantile level. The distortion
probe is
\begin{equation}
\label{eq:sngp-distortion}
Z_{\mathrm{dist}}(x)
=
\max_{g\in\{\mathrm{same},\mathrm{cross}\}}
D_g^{(x)},
\qquad
Z_{\mathrm{dist}}(x)\downarrow.
\end{equation}
It measures pair-dependent distortion relative to the input space, rather
than absolute feature scale, GP posterior variance, or corrected predictive
uncertainty.

\paragraph{Guard metric.}

Classification accuracy on the keep-validation split is used only as a
safety guard:
\begin{equation}
\label{eq:sngp-accuracy-guard}
Z_{\mathrm{acc}}(x)
=
\frac{1}{|\mathcal V_{\mathrm{keep}}|}
\sum_{i\in\mathcal V_{\mathrm{keep}}}
\mathbf 1
\left[
\arg\max_c \ell_{i,c}^{(x)}=y_i
\right],
\qquad
Z_{\mathrm{acc}}(x)\uparrow.
\end{equation}
Because positive temperature scaling does not change the predicted class,
this metric is computed directly from the mean-field-corrected logits.
For a Stage-II candidate \(x'\) and the fixed Stage-I anchor \(a\), the guard is
\begin{equation}
\label{eq:sngp-guard}
\mathcal A_{\mathrm{SNGP}}(x')
=
\mathbf 1
\left[
Z_{\mathrm{acc}}(x')
\ge
Z_{\mathrm{acc}}(a)-\delta_{\mathrm{acc}}
\right].
\end{equation}
The guard prevents probe improvements from being retained at the cost of a
material loss in classification accuracy. It is not included in the probe
vector or combined with NLL through a weighted scalar objective.

\paragraph{Relation to formal evaluation.}

The search-time quantities serve different roles from the held-out evaluation
metrics. The scalar objective \(S\) measures validation NLL during evolution,
whereas predictive fit is formally assessed by clean NLL on held-out test
data. These quantities measure the same aggregate behavior but are not
numerically identical because they use different data and evaluation
protocols. The probes \(Z_{\mathrm{mar}}\) and \(Z_{\mathrm{dist}}\) expose
in-distribution representation mechanisms associated with class separation
and distance-aware uncertainty, respectively; they do not directly optimize
an OOD dataset or predictive uncertainty score. Formal uncertainty
performance is instead evaluated on held-out OOD data using a ranking-based
Dempster--Shafer metric. Finally, \(Z_{\mathrm{acc}}\) is used only as a
search-time safety guard, rather than as a conflict probe or primary formal
evaluation metric. The quantile levels, pair-panel construction, numerical
constants, and guard tolerance are specified in
Appendix~\ref{app:experimental_details}.

\subsection{ESN}
\label{app:probe_esn}

\paragraph{Conflict instantiation.}

An echo-state network (ESN) uses a recurrent reservoir to transform an input
sequence before fitting a linear readout. Its conflict instantiates
\textbf{R2.M4}, temporal-component mismatch, on the \texttt{time} axis. The
competing behaviors are
\begin{align}
B_A &: \text{retention of sensitivity to past reservoir states},
\\
B_B &: \text{state-dependent nonlinear temporal processing}.
\end{align}
Both behaviors are controlled by the reservoir state transition. Dynamics
that remain close to linear may preserve past-state sensitivity but provide
limited nonlinear transformation, whereas strong state-dependent contraction
or saturation may erase that sensitivity over long delays.

Let $x$ denote a candidate implementation of the reservoir update. For
trajectory realization $q$, define
\begin{equation}
\label{eq:esn-transition}
\mathbf h_{q,t}^{(x)}
=
F_x\!\left(
\mathbf h_{q,t-1}^{(x)},\mathbf u_{q,t}
\right),
\end{equation}
where $\mathbf h_{q,t}^{(x)}\in\mathbb R^{n_h}$ is the reservoir state,
$\mathbf u_{q,t}$ is the current input, and $q$ identifies the frozen
sequence--reservoir realization used by the evaluator.

The conflict-specific Probe vector is
\begin{equation}
\label{eq:esn-probe-vector}
\mathbf Z_{\mathrm{ESN}}(x)
=
\left[
Z_{\mathrm{mem}}(x),
Z_{\mathrm{nl}}(x)
\right],
\qquad
Z_{\mathrm{mem}}\uparrow,
\quad
Z_{\mathrm{nl}}\uparrow .
\end{equation}
The first Probe measures the transmission of local state sensitivity across
time, whereas the second measures how the local state--input transition varies
over the realized trajectory.

\paragraph{Scalar task metric.}

For candidate $x$ and trajectory realization $q$, a ridge readout is fitted on
the registered readout-training segment. Let
$\widehat y_{q,t}^{(x)}$ and $y_{q,t}$ denote the prediction and target on the
disjoint evolution-validation segment $\mathcal V_q$, and let
\begin{equation}
\overline y_q
=
\frac{1}{|\mathcal V_q|}
\sum_{t\in\mathcal V_q}y_{q,t}.
\end{equation}
The realization-level normalized error is
\begin{equation}
\label{eq:esn-realization-nrmse}
S_q(x)
=
\sqrt{
\frac{
\sum_{t\in\mathcal V_q}
\left(\widehat y_{q,t}^{(x)}-y_{q,t}\right)^2
}{
\sum_{t\in\mathcal V_q}
\left(y_{q,t}-\overline y_q\right)^2
}
}.
\end{equation}
Thus, the normalization uses centered target energy; it is not based on target
range or on an already standardized RMSE. For the frozen set
$\mathcal Q_{\mathrm{evo}}$ of evolution realizations, the unique scalar metric
used for search-time task feedback is
\begin{equation}
\label{eq:esn-task}
S(x)
=
\frac{1}{|\mathcal Q_{\mathrm{evo}}|}
\sum_{q\in\mathcal Q_{\mathrm{evo}}}S_q(x),
\qquad
S(x)\downarrow .
\end{equation}
MSE and $R^2$ may be recorded as task diagnostics, but they do not participate
in search-time retention; formal evaluation reports them as additional task
metrics.

\paragraph{Probe 1: Jacobian-product memory retention.}

For trajectory $q$, the recurrent-state Jacobian at time $t$ is
\begin{equation}
\label{eq:esn-state-jacobian}
J_{q,t}^{(x)}
=
\frac{
\partial F_x\!\left(
\mathbf h_{q,t-1}^{(x)},\mathbf u_{q,t}
\right)
}{
\partial \mathbf h_{q,t-1}^{(x)}
}.
\end{equation}
For delay $d$, define the ordered Jacobian product
\begin{equation}
\label{eq:esn-jacobian-product}
P_{q,t,d}^{(x)}
=
J_{q,t}^{(x)}
J_{q,t-1}^{(x)}
\cdots
J_{q,t-d+1}^{(x)}
=
\frac{
\partial \mathbf h_{q,t}^{(x)}
}{
\partial \mathbf h_{q,t-d}^{(x)}
}.
\end{equation}
It measures how local perturbations to a past reservoir state are transmitted
to the state at the anchor time. It does not directly measure recovery of a
past input, because the input Jacobian is not included.

Let $\mathcal T_{\mathrm{anc}}$ be the frozen set of anchor times and
$\mathcal D_{\mathrm{lag}}=\{1,\ldots,d_{\max}\}$ the evaluated delays. An
anchor is therefore the pair $(q,t)$ with
$q\in\mathcal Q_{\mathrm{evo}}$ and
$t\in\mathcal T_{\mathrm{anc}}$. For one realization,
\begin{equation}
\label{eq:esn-memory-per-realization}
Z_{\mathrm{mem}}^{(q)}(x)
=
\frac{1}{
|\mathcal T_{\mathrm{anc}}|
|\mathcal D_{\mathrm{lag}}|
n_h
}
\sum_{t\in\mathcal T_{\mathrm{anc}}}
\sum_{d\in\mathcal D_{\mathrm{lag}}}
\sum_{j=1}^{n_h}
\min\!\left\{
\sigma_j\!\left(P_{q,t,d}^{(x)}\right),1
\right\}.
\end{equation}
The reported Probe averages over the frozen realizations,
\begin{equation}
\label{eq:esn-memory}
Z_{\mathrm{mem}}(x)
=
\frac{1}{|\mathcal Q_{\mathrm{evo}}|}
\sum_{q\in\mathcal Q_{\mathrm{evo}}}
Z_{\mathrm{mem}}^{(q)}(x),
\qquad
Z_{\mathrm{mem}}(x)\uparrow .
\end{equation}
Clipping each singular value at one prevents expansive directions from
receiving more credit than unit transmission. Higher values therefore mean
that local sensitivity to more past-state directions is retained over the
evaluated delays. The clipping does not by itself distinguish unit
transmission from expansion and should not be interpreted as a complete
stability measure.

\paragraph{Probe 2: local nonlinear variation.}

To characterize state-dependent changes in the transition, define the joint
state--input Jacobian
\begin{equation}
\label{eq:esn-joint-jacobian}
K_{q,t}^{(x)}
=
\frac{
\partial F_x\!\left(
\mathbf h_{q,t-1}^{(x)},\mathbf u_{q,t}
\right)
}{
\partial
[\mathbf h_{q,t-1}^{(x)};\mathbf u_{q,t}]
},
\end{equation}
where $[\cdot;\cdot]$ denotes concatenation. Its realization-specific
anchor mean is
\begin{equation}
\overline K_q^{(x)}
=
\frac{1}{|\mathcal T_{\mathrm{anc}}|}
\sum_{t\in\mathcal T_{\mathrm{anc}}}
K_{q,t}^{(x)}.
\end{equation}
The normalized variation for realization $q$ is
\begin{equation}
\label{eq:esn-nonlinearity-per-realization}
Z_{\mathrm{nl}}^{(q)}(x)
=
\frac{
\displaystyle
\frac{1}{|\mathcal T_{\mathrm{anc}}|}
\sum_{t\in\mathcal T_{\mathrm{anc}}}
\left\lVert
K_{q,t}^{(x)}-\overline K_q^{(x)}
\right\rVert_F^2
}{
\displaystyle
\frac{1}{|\mathcal T_{\mathrm{anc}}|}
\sum_{t\in\mathcal T_{\mathrm{anc}}}
\left\lVert K_{q,t}^{(x)}\right\rVert_F^2
+\epsilon_{\mathrm{nl}}
}.
\end{equation}
The second Probe is
\begin{equation}
\label{eq:esn-nonlinearity}
Z_{\mathrm{nl}}(x)
=
\frac{1}{|\mathcal Q_{\mathrm{evo}}|}
\sum_{q\in\mathcal Q_{\mathrm{evo}}}
Z_{\mathrm{nl}}^{(q)}(x),
\qquad
Z_{\mathrm{nl}}(x)\uparrow .
\end{equation}
Here $\lVert\cdot\rVert_F$ is the Frobenius norm. An affine transition has a
location-independent Jacobian and hence zero numerator. The denominator
removes overall Jacobian scale, so the Probe responds to state--input-dependent
variation rather than to a uniformly large linear response. It is a
mechanism-level proxy for nonlinear processing, not a direct measure of
task-useful nonlinearity; large local variation is interpreted together with
the scalar task metric and the companion memory Probe.

\paragraph{Guard metric.}

No additional model-specific stability guard is registered for ESN. In
particular, the evaluator does not compute or select on a separate
Jacobian-expansion statistic. Candidate validation still enforces finite and
deterministic state transitions, the registered operation bound, and
dependence on both recurrent and input projections. The conflict-aware
retention route additionally requires non-negative task gain and a calibrated
improvement in one Probe without a beyond-threshold degradation of the other.
Thus, no edit with degraded search-time NRMSE can be retained through the
Probe route. Exact thresholds and retention logic are given in
Appendix~\ref{app:experimental_details}.

\paragraph{Relation to formal evaluation.}

Sequence realizations, reservoir matrices, anchor times, and delay indices are
fixed across candidates, while candidate-dependent states and Jacobians are
recomputed under this common workload. \(Z_{\mathrm{mem}}\) and
\(Z_{\mathrm{nl}}\) are readout-free search-time Probes of reservoir
dynamics, whereas the scalar task metric is computed after fitting the ridge
readout. This separates the measured reservoir mechanisms from downstream
predictive performance.

Formal NRMSE uses the same centered-target-energy normalization as
Eq.~(\ref{eq:esn-realization-nrmse}), but is evaluated on independent
held-out sequence--reservoir pairs; formal MSE and \(R^2\) are additional
task-reporting metrics. Mackey--Glass long-horizon transfer tests whether the
selected reservoir dynamics translate into improved downstream forecasting,
but its prediction error is not identical to either Probe. The exact anchor
count, delay range, sequence lengths, reservoir seeds, numerical constant,
singular-value computation, and transfer horizons are specified in
Appendix~\ref{app:experimental_details}.

\subsection{GCNII}
\label{app:probe_gcnii}

\paragraph{Conflict instantiation.}

GCNII propagates node representations through a normalized graph adjacency
while retaining an initial residual and an identity mapping.  Its conflict
instantiates \textbf{R2.M5}, relational-structure mismatch, on the
\texttt{space} axis.  The competing behaviors are
\begin{align}
B_A &: \text{effective use of informative neighborhood messages},
\\
B_B &: \text{resistance to incompatible-message contamination}.
\end{align}
They interact through the same propagation path: suppressing neighborhood
propagation can discard useful relational evidence, whereas stronger or less
selective aggregation can amplify incompatible messages.  Clean validation
NLL provides task-level evidence of useful aggregation, but it does not
isolate neighborhood utility from node features and classifier quality.  The
conflict-specific Probe therefore measures the additional loss induced by
frozen high-disagreement message substitutions.

The conflict-specific Probe vector contains one scalar:
\begin{equation}
\label{eq:gcnii-probe-vector}
\mathbf Z_{\mathrm{GCNII}}(x)
=
\left[Z_{\mathrm{dis}}(x)\right],
\qquad
Z_{\mathrm{dis}}(x)\downarrow,
\end{equation}
where $x$ denotes a candidate implementation evaluated by the frozen training
and validation procedure.

\paragraph{Scalar task metric.}

Let $A$ denote the clean graph representation, $X$ the node-feature matrix,
and $\mathcal V_{\mathrm{keep}}$ the keep-validation nodes.  Let
$\mathcal N(A)$ denote the fixed GCNII preprocessing operator that inserts
self-loops and applies the evaluator's adjacency normalization.  For candidate
$x$, $p_x(\cdot\mid\mathcal N(A),X)_i$ denotes the predictive distribution of
the trained candidate at node $i$.  Define
\begin{equation}
\label{eq:gcnii-nll}
\mathcal L_{\mathrm{NLL}}
\left(x;A,X,\mathcal V_{\mathrm{keep}}\right)
=
-\frac{1}{|\mathcal V_{\mathrm{keep}}|}
\sum_{i\in\mathcal V_{\mathrm{keep}}}
\log
p_x
\left(
y_i\mid\mathcal N(A),X
\right).
\end{equation}
The search-time scalar task metric is
\begin{equation}
\label{eq:gcnii-task}
S(x)
=
\mathcal L_{\mathrm{NLL}}
\left(x;A,X,\mathcal V_{\mathrm{keep}}\right),
\qquad
S(x)\downarrow.
\end{equation}
Clean accuracy is recorded as a task diagnostic, but $S(x)$ is the scalar
metric used for online task comparison.

\paragraph{Frozen disagreement intervention.}

Let $a$ be the frozen Stage-I anchor.  To obtain a label-independent signal
for intervention construction, evaluate its trained predictor using only
self-loops:
\begin{equation}
\label{eq:gcnii-anchor-self-prediction}
\mathbf q_i
=
p_a
\left(
\,\cdot\mid\mathcal N(A_{\mathrm{self}}),X
\right)_i,
\end{equation}
where $\mathbf q_i$ is a class-probability vector and
$A_{\mathrm{self}}$ contains no inter-node messages.  For receiver $i$ and
candidate source $j$, define
\begin{equation}
\overline{\mathbf q}_{ij}
=
\frac{\mathbf q_i+\mathbf q_j}{2}
\end{equation}
and
\begin{equation}
\label{eq:gcnii-js}
\operatorname{JS}(\mathbf q_i,\mathbf q_j)
=
\frac{1}{2}
\operatorname{KL}
\left(
\mathbf q_i\Vert\overline{\mathbf q}_{ij}
\right)
+
\frac{1}{2}
\operatorname{KL}
\left(
\mathbf q_j\Vert\overline{\mathbf q}_{ij}
\right),
\end{equation}
where $\operatorname{KL}$ is the Kullback--Leibler divergence.

For a fixed fraction $\rho$ of eligible incoming non-self messages to
keep-validation receivers, the original source is replaced by a frozen
train-or-validation non-neighbor selected from a degree-matched pool to
maximize Eq.~(\ref{eq:gcnii-js}).  Repeating this construction produces $K$
frozen raw graph realizations
\begin{equation}
A_{\mathrm{dis}}^{(1)},\ldots,A_{\mathrm{dis}}^{(K)}.
\end{equation}
The construction reads neither ground-truth labels nor test nodes.  Thus,
\emph{the intervention construction} is label-free; the subsequent NLL
evaluation is not, because it necessarily uses keep-validation labels.

Replacements operate on the evaluator's stored receiver--source edge
representation.  They do not impose additional reciprocal edges.  Every
clean or perturbed raw graph is subsequently processed by the same self-loop
and normalization operator $\mathcal N$, so the intervention changes message
sources without changing the GCNII graph-preprocessing convention.  The
anchor predictions, selected replacement pairs, and all intervention
realizations are frozen before conflict-aware evolution and cannot be changed
by a candidate.  Exact values of $\rho$ and $K$, the degree-matched pool,
construction seeds, tie handling, and graph hashes are specified in
Appendix~\ref{app:experimental_details}.

\paragraph{Primary Probe.}

For a trained candidate $x$, first evaluate Eq.~(\ref{eq:gcnii-task}) on the
clean graph.  Without retraining or fine-tuning $x$, perform an additional
forward evaluation on each frozen perturbed graph.  The disagreement
sensitivity is
\begin{equation}
\label{eq:gcnii-disagreement}
Z_{\mathrm{dis}}(x)
=
\frac{1}{K}
\sum_{k=1}^{K}
\left[
\mathcal L_{\mathrm{NLL}}
\left(
x;
A_{\mathrm{dis}}^{(k)},
X,
\mathcal V_{\mathrm{keep}}
\right)
-
S(x)
\right],
\qquad
Z_{\mathrm{dis}}(x)\downarrow.
\end{equation}
Subtracting the candidate's own clean NLL removes its candidate-specific
clean-loss level and isolates the incremental cost of disagreement
contamination.  A smaller value therefore indicates greater resistance to
incompatible incoming messages, rather than merely better clean
classification.  The quantity is not clipped at zero: a negative realization
is valid and means that the particular frozen intervention did not increase
NLL.  Online parent--candidate comparisons use the same intervention set and
paired training seeds; their aggregation and retention rule are given in
Appendix~\ref{app:experimental_details}.

\paragraph{Guard metric.}

GCNII introduces no additional model-specific guard metric.  A small
$Z_{\mathrm{dis}}$ alone is insufficient, because a low-quality predictor
that is uniformly insensitive on both clean and perturbed graphs could also
produce a small contaminated-minus-clean difference.  The common task
condition therefore requires non-degraded clean validation NLL before a Probe
improvement can support retention.  Clean NLL is the scalar task metric, not
an additional conflict Probe; exact tie handling and numerical tolerances are
specified in Appendix~\ref{app:experimental_details}.

\paragraph{Relation to formal evaluation.}

$Z_{\mathrm{dis}}$ is a search-time measure of the incremental validation
loss caused by frozen high-disagreement message substitutions.  Formal
evaluation applies the same contaminated-minus-clean functional to held-out
evaluation nodes:
\begin{equation}
\label{eq:gcnii-formal-contamination-gap}
\Delta\mathrm{NLL}_{\mathrm{contam}}
=
\mathrm{NLL}_{\mathrm{contaminated}}
-
\mathrm{NLL}_{\mathrm{clean}}.
\end{equation}
This difference, rather than contaminated NLL alone, is the formal quantity
most directly corresponding to the Probe.  The formal stress test and the
search-time Probe use the same sensitivity functional but need not use the
same intervention-construction information: formal contamination may use
held-out ground-truth incompatibility because it is computed only after
source selection and cannot affect evolution.  Clean accuracy and clean NLL
remain task-performance metrics. On Wisconsin, ConflictGuide reduces mean contaminated NLL in all three
rounds, whereas the mean contamination-induced increase is reduced only
in round~2 relative to AutoResearch. Formal replacement rules, perturbation
realizations, and aggregation across evaluation splits are specified in
Appendix~\ref{app:experimental_details}.

\subsection{TCM--Lite}
\label{app:probe_tcm}

\paragraph{Conflict instantiation.}

TCM--Lite is a learned image codec comprising an analysis transform,
quantized main and hyper latents, an entropy model, and a synthesis transform.
Its conflict instantiates \textbf{R3.M10}, compact representation damaging
fine structure, on the \texttt{space} axis.  The competing behaviors are
\begin{align}
B_A &: \text{entropy-efficient latent coding},
\\
B_B &: \text{multiscale fine-detail preservation}.
\end{align}
The two behaviors interact through the shared quantized representation and
synthesis path.  Reducing the entropy of the transmitted latents can suppress
edges and textures, whereas preserving additional spatial detail can require
a less compact representation.

The conflict-specific Probe vector is
\begin{equation}
\label{eq:tcm-probe-vector}
\mathbf Z_{\mathrm{TCM}}(x)
=
\left[
Z_{\mathrm{rate}}(x),
Z_{\mathrm{detail}}(x)
\right],
\qquad
Z_{\mathrm{rate}}(x)\downarrow,
\quad
Z_{\mathrm{detail}}(x)\downarrow,
\end{equation}
where $x$ denotes a candidate implementation evaluated under the frozen
training and validation procedure.  The first Probe exposes the estimated
coding rate, while the second measures multiscale luminance-gradient
distortion.  Thus, $Z_{\mathrm{detail}}$ operationalizes fine-detail
preservation through luminance-gradient structure; it does not directly
measure chromatic detail.

Because the scalar rate--distortion objective aggregates rate and pixel
distortion, similar scalar values can mask different changes in coding rate
and spatial detail.  The two behaviors are therefore exposed separately.
This observability limitation is not treated as a second physical conflict.

\paragraph{Scalar task metric.}

Let $\mathcal V_{\mathrm{keep}}$ denote the fixed evolution-validation image
set.  For image $n$, let $\mathbf Y_n\in[0,1]^{C\times H_n\times W_n}$ be the
target and let $\widehat{\mathbf Y}_n^{(x)}$ be the reconstruction produced by
candidate $x$, where $C=3$.  Define the total number of evaluated spatial
pixels as
\begin{equation}
N_{\mathrm{pix}}
=
\sum_{n\in\mathcal V_{\mathrm{keep}}}H_nW_n
\end{equation}
and the mean RGB reconstruction error as
\begin{equation}
\label{eq:tcm-mse}
D_{\mathrm{mse}}(x)
=
\frac{
\displaystyle
\sum_{n\in\mathcal V_{\mathrm{keep}}}
\left\|
\mathbf Y_n-\widehat{\mathbf Y}_n^{(x)}
\right\|_F^2
}{
C N_{\mathrm{pix}}
}.
\end{equation}

At the fixed online operating point, the scalar task metric is the
rate--distortion objective
\begin{equation}
\label{eq:tcm-task}
S(x)
=
Z_{\mathrm{rate}}(x)
+
\lambda_{\mathrm{RD}}\,
\kappa_{\mathrm{pix}}^2
D_{\mathrm{mse}}(x),
\qquad
S(x)\downarrow,
\end{equation}
where $\lambda_{\mathrm{RD}}>0$ is the frozen rate--distortion multiplier and
$\kappa_{\mathrm{pix}}$ converts normalized image differences to the pixel
scale used by the evaluator.  The exact operating point and pixel convention
are specified in Appendix~\ref{app:experimental_details}.  Although
$Z_{\mathrm{rate}}$ contributes to $S(x)$, the scalar value does not reveal
whether a change arose from rate, pixel distortion, or an offsetting movement
between the two.

\paragraph{Probe 1: estimated latent rate.}

Let $\widehat{\mathbf y}^{(x)}$ and
$\widehat{\mathbf z}^{(x)}$ denote the quantized main and hyper latents
produced by candidate $x$.  Let
$p_x(\widehat y_i^{(x)})$ and
$p_x(\widehat z_j^{(x)})$ denote the corresponding entropy-model
likelihoods.  This notation leaves implicit all model-specific conditioning,
including hyperprior information and the entropy context available for the
main latent.

The estimated numbers of coded bits for the two latent streams are
\begin{align}
\operatorname{bits}_{y}(x)
&=
-\sum_i
\log_2
\max\left\{
p_x\!\left(\widehat y_i^{(x)}\right),
\epsilon_p
\right\},
\\
\operatorname{bits}_{z}(x)
&=
-\sum_j
\log_2
\max\left\{
p_x\!\left(\widehat z_j^{(x)}\right),
\epsilon_p
\right\},
\end{align}
where the indices range over every latent symbol generated for
$\mathcal V_{\mathrm{keep}}$ and $\epsilon_p>0$ prevents taking the logarithm
of zero.

The estimated-rate Probe is
\begin{equation}
\label{eq:tcm-rate}
Z_{\mathrm{rate}}(x)
=
\frac{
\operatorname{bits}_{y}(x)
+
\operatorname{bits}_{z}(x)
}{
N_{\mathrm{pix}}
},
\qquad
Z_{\mathrm{rate}}(x)\downarrow.
\end{equation}
It is measured in estimated bits per input pixel.  A lower value indicates
that the candidate assigns higher likelihood to its quantized latent
representation and therefore predicts a more compact coded representation.
It is an entropy-model estimate and need not exactly equal the number of bits
produced by arithmetic coding.

\paragraph{Probe 2: multiscale detail distortion.}

Let $\mathcal L(\cdot)$ denote the fixed RGB-to-luminance transform,
$\mathcal D_s(\cdot)$ the fixed downsampling operator at scale
$s\in\mathcal S$, and $\nabla$ the fixed two-direction Sobel operator.  For
target image $n$ and candidate $x$, define
\begin{align}
L_{n,s}
&=
\mathcal D_s\!\left(\mathcal L(\mathbf Y_n)\right),
\\
\widehat L_{n,s}^{(x)}
&=
\mathcal D_s\!\left(
\mathcal L(\widehat{\mathbf Y}_n^{(x)})
\right).
\end{align}
The normalized gradient discrepancy at scale $s$ is
\begin{equation}
\label{eq:tcm-detail-scale}
D_s(x)
=
\frac{
\displaystyle
\sum_{n\in\mathcal V_{\mathrm{keep}}}
\left\|
\nabla L_{n,s}
-
\nabla\widehat L_{n,s}^{(x)}
\right\|_1
}{
\displaystyle
\sum_{n\in\mathcal V_{\mathrm{keep}}}
\left(
\left\|\nabla L_{n,s}\right\|_1
+
\left\|\nabla\widehat L_{n,s}^{(x)}\right\|_1
\right)
+
\epsilon_{\mathrm{detail}}
},
\end{equation}
where the $\ell_1$ norm sums over spatial positions and both Sobel-gradient
directions, and $\epsilon_{\mathrm{detail}}>0$ stabilizes the normalization.

The multiscale detail Probe is
\begin{equation}
\label{eq:tcm-detail}
Z_{\mathrm{detail}}(x)
=
\frac{1}{|\mathcal S|}
\sum_{s\in\mathcal S}D_s(x),
\qquad
Z_{\mathrm{detail}}(x)\downarrow.
\end{equation}
This symmetric normalization reduces sensitivity to the absolute amount of
gradient energy in the image.  A lower value indicates that the reconstruction
more closely preserves the target's multiscale luminance-edge and texture
structure.  It should not be interpreted as a complete perceptual-quality
metric or as a measurement of color-detail preservation.  The scale set,
luminance coefficients, downsampling rule, Sobel kernels and padding
convention, and numerical stabilizers are specified in
Appendix~\ref{app:experimental_details}.

\paragraph{Guard metric.}

TCM--Lite uses no additional model-specific guard metric. The scalar
rate--distortion metric \(S(x)\) provides the aggregate task measure alongside
the rate and detail Probes. A change in either Probe alone does not establish
an improvement in overall rate--distortion performance.

\paragraph{Relation to formal evaluation.}

The evolution-validation images and Probe transformations are fixed across
candidates. Online \(Z_{\mathrm{rate}}\) is computed from entropy-model
likelihoods, whereas formal rate evaluation uses realized arithmetic-coded
bitstreams on held-out images.

Let \(b_n^{(x)}\) be the total number of coded payload bytes in the main- and
hyper-latent streams for held-out image \(n\). Its actual rate is
\begin{equation}
\label{eq:tcm-actual-bpp-image}
\operatorname{bpp}_{\mathrm{actual},n}(x)
=
\frac{8b_n^{(x)}}{H_nW_n}.
\end{equation}
The reported dataset-level rate is the unweighted mean over images:
\begin{equation}
\label{eq:tcm-actual-bpp}
\operatorname{bpp}_{\mathrm{actual}}(x)
=
\frac{1}{N_{\mathrm{img}}}
\sum_{n=1}^{N_{\mathrm{img}}}
\operatorname{bpp}_{\mathrm{actual},n}(x).
\end{equation}
Actual bpp measures the same coding-rate behavior as \(Z_{\mathrm{rate}}\),
but need not equal the likelihood-based estimate.

Formal detail evaluation applies the same luminance transform, multiscale
downsampling, Sobel operator, and normalized gradient discrepancy to held-out
images, then averages the per-image results. It measures the same
detail-preservation behavior as \(Z_{\mathrm{detail}}\), although the data
and aggregation differ from the online patch-level Probe. MS-SSIM and PSNR
provide complementary reconstruction-quality measures. Held-out metrics
are unavailable during evolution. The exact operating points, image and
coding conventions, and Probe settings are specified in
Appendix~\ref{app:experimental_details}.

\section{Experimental Details}
\label{app:experimental_details}

This section specifies the data isolation, evolution substrates, editable
boundaries, proposal accounting, and held-out evaluation used across the five
model families. We follow the notation and retention rules defined in
Sections~\ref{sec:setup} and~\ref{sec:evolution}; model-specific Probe
definitions are provided in Appendix~\ref{app:model_probes}. Held-out metrics
are computed only after the terminal source has been frozen and are never used
during evolution.

\subsection{Common two-stage evolution protocol}
\label{app:common_evolution_protocol}

\paragraph{Candidate evaluation and state isolation.}
Each iteration uses a fresh coding-agent session. A candidate is evaluated
under the same frozen data split, training seeds, perturbations, and resource
limits that produced the registered metric record for the current incumbent.
The candidate is initialized according to the model-specific training
procedure rather than from the incumbent's trained weights.

The incumbent need not be retrained or reevaluated at every iteration. Its
metric record may be reused while its source hash and all evaluator, split,
seed, and frozen-asset hashes remain unchanged. Evaluators requiring explicit
paired reevaluation instead recompute the incumbent and candidate within the
same invocation. When an edit is retained, its source and registered metric
record become the new incumbent. No candidate is warm-started from inherited
weights, optimizer state, fitted temperature, or covariance state. The
controller history, source hashes, decisions, and eligible incumbent-metric
cache persist as evolution metadata.

\paragraph{Matched two-stage construction.}
For each search round, identified by its search seed, Stage~I is run
independently from the registered reference under task-only feedback. Its
terminal incumbent, denoted by \(a_r=x_{T_1}^{(r)}\), is the common branch
point for that round. Thus, the three reported search rounds contain three
independently generated Stage-I trajectories; a Stage-I trajectory is shared
only by the matched Stage-II continuations within the same round.

Within each round, the scalar-only and ConflictGuide continuations start from
the same source hash and use the same evaluator, data seeds, editable source
boundary, training recipe, resource limits, and proposal budget, as detailed
in the Appendix~\ref{app:budget_exceptions}. The
coding-agent model and base prompt contract are also fixed. The current source
and accumulated history naturally diverge after branching because the two
arms may retain different edits.

The scalar-only continuation follows the standard AutoResearch retention
rule, whereas ConflictGuide follows Eq.~(\ref{eq:retention}) and additionally
receives the qualified conflict and Probe feedback, together with registered
route or rejection information where available. Where the evaluator computes
hidden Probes in the scalar-only arm to match evaluation cost, these values
are neither exposed to the agent nor used for retention. Rejected and invalid
proposals leave the incumbent unchanged. In trajectory plots, the incumbent
is carried forward until the next accepted edit and through the budget
endpoint. The reported result is the terminal incumbent, not a candidate
selected using held-out metrics.

\paragraph{Proposal and failure accounting.}
A proposal consumes one iteration once the controller has materialized a
candidate-source artifact, including candidates that subsequently fail static
validation, crash, produce non-finite values, exceed the memory limit, or time
out. A service failure before a candidate artifact is produced is retried
without consuming the proposal budget. Each attempted source, source hash,
parent hash, evaluation status, metric record, retention decision, and
incumbent update is recorded in an append-only ledger. After every rejection,
the exact parent source is restored.


\subsection{Model-specific evolution and evaluation protocols}
\label{app:model_specific_evolution_setups}
\label{app:formal_evaluation_protocols}

\paragraph{Split design and data isolation.}
We distinguish the \emph{evolution substrate} from the \emph{formal
evaluation protocol}. The former is a deterministic, reduced-cost proxy that
makes hundreds of candidate evaluations feasible; the latter restores the
full model and benchmark-scale data after the terminal source has been frozen.
The benchmark datasets and formal split conventions follow the corresponding
model literature: the 1,000/200 Navier--Stokes convention used by FNO-family
studies~\citep{li2020fourier,tran2021factorized}, the official CIFAR-100
train/test partition used by SNGP~\citep{krizhevsky2009learning,liu2020simple},
the seven-dataset, ten-split GCNII protocol~\citep{chen2020simple,pei2020geom},
standard NARMA and Mackey--Glass reservoir-computing
benchmarks~\citep{verstraeten2007experimental,ceni2024residual,mackey1977oscillation},
and full-image Kodak evaluation used in learned image
compression~\citep{liu2023learned}. In contrast, the exact search-time counts,
such as the 200/48/48 SpecB--FNO allocation and the 2,048-patch TCM--Lite
panels, are fixed computational allocations introduced for evolution; they
are not claimed to be benchmark splits prescribed by the cited studies.

For every search seed, all search-time data roles, sample indices, random
seeds, and derived Probe assets are fixed before they are used for candidate
selection. When separate data are required for nuisance-parameter fitting,
Probe construction, qualification, or audit, their roles are fixed as
specified below. In particular, the SpecB--FNO spectral mask and SNGP
temperature are estimated on panels disjoint from keep-validation. The GCNII
online perturbation graphs instead use keep-validation receivers, but are
constructed once at the Stage-I branch point without test nodes or
ground-truth labels and are then shared unchanged by the matched Stage-II
arms. Audit and formal-test data are unavailable to both the coding agent and
the online retention controller. Within a search seed, matched Stage-II arms
reuse the same frozen data manifest.

\paragraph{Source transfer and reporting.}
Formal evaluation begins only after the terminal source hashes have been
frozen. Only source code and architecture are transferred; proxy weights,
optimizer states, fitted calibration parameters, Probe thresholds, and
validation checkpoints are discarded. The reference and all compared terminal
sources are retrained from scratch using paired seeds or official splits.
Held-out data are never used for online retention or source selection.

SpecB--FNO, SNGP, and TCM--Lite use the checkpoint at the end of their fixed
training schedules. GCNII selects checkpoints using clean validation NLL,
whereas ESN fits a fresh closed-form readout for every paired configuration.
No test metric is used for checkpoint selection. Unless stated otherwise,
reported dispersions are sample standard deviations.

Table~\ref{tab:app_data_isolation} summarizes the data roles. ``Keep-val''
denotes the split used for online parent--candidate decisions; it is not a
formal test set.

\begin{table*}[t]
\centering
\caption{Search-time data isolation and post-search evaluation. Reduced
evolution substrates support search throughput, whereas formal results are
obtained only after source freezing and full-scale retraining.}
\label{tab:app_data_isolation}
\footnotesize
\setlength{\tabcolsep}{3.5pt}
\renewcommand{\arraystretch}{1.08}
\begin{tabularx}{\textwidth}{@{}
  l
  >{\raggedright\arraybackslash}p{0.27\textwidth}
  >{\raggedright\arraybackslash}p{0.22\textwidth}
  >{\raggedright\arraybackslash}X@{}}
\toprule
\textbf{Model} &
\textbf{Evolution training and auxiliary data} &
\textbf{Online keep-validation} &
\textbf{Post-search retraining and formal evaluation} \\
\midrule

SpecB--FNO &
Navier--Stokes official training pool: 200 training examples and 48 disjoint
mask-construction examples; the remaining 704 examples are unused online. &
48 fixed examples, disjoint from training and mask construction. &
\textit{Retrain:} 800/100/100 train/mask/validation partition of the official
training pool. \textit{Test:} 200 official examples (indices 1000--1199). \\
\addlinespace

SNGP &
CIFAR-100 training set, stratified per class into 400 training and 20
temperature-calibration images. A further 50 images/class are reserved for
search-inaccessible audit. &
30 fixed images/class, disjoint from training, calibration, and audit. &
\textit{Retrain:} all 50,000 CIFAR-100 training images. \textit{Test:}
CIFAR-100; balanced OOD comparisons with SVHN, and CIFAR-10. \\
\addlinespace

GCNII &
Chameleon public split~0: its 60\% per-class training partition; five frozen,
label-free disagreement graphs are constructed without test nodes. &
The corresponding 20\% validation partition; the 20\% test partition is
inaccessible online. &
\textit{Retrain/test:} the frozen source is trained independently on each of
ten public 60/20/20 splits of Cora, Citeseer, Pubmed, Chameleon, Cornell,
Texas, and Wisconsin. \\
\addlinespace

ESN &
One frozen NARMA-30 realization: 200 washout and 5,000 readout-fit samples,
evaluated with reservoir seeds 1103, 2207, and 3301. &
The subsequent 2,500 samples from the same realization. &
\textit{Refit/test:} independent NARMA-30 and Mackey--Glass sequences crossed
with ten independent reservoir seeds; a fresh ridge readout is fitted for
every pair. \\
\addlinespace

TCM--Lite &
An image-disjoint DIV2K+Flickr2K patch bank: 32,768 training patches and four
separate 2,048-patch panels for mechanism development, hidden audit,
qualification discovery, and qualification confirmation. &
2,048 fixed \(64\times64\) RGB patches from a fifth
source-image-disjoint panel. &
\textit{Retrain:} all 3,450 DIV2K+Flickr2K source images. \textit{Test:}
full-resolution Kodak images using actual arithmetic coding. \\
\bottomrule
\end{tabularx}
\end{table*}

\subsubsection{SpecB--FNO}
\label{app:exp_specb}

\paragraph{Evolution data and task.}
We use \texttt{NavierStokes\_V1e-5\_N1200\_T20.mat} at viscosity
\(\nu=10^{-5}\) and spatial resolution \(64\times64\). Each example provides
ten context frames followed by a ten-step autoregressive target rollout. The
dataset contains the 1,000 training and 200 held-out test trajectories used in
the FNO-family benchmark convention~\citep{li2020fourier,tran2021factorized}.
The official 200-example test set is never loaded during evolution.

Within the 1,000-example training pool, the proxy assigns 200 examples to
candidate fitting, 48 to construction of the frozen dominant/non-dominant
spectral partition, and 48 to keep-validation. The remaining 704 examples are
deliberately unused online. Reducing candidate fitting to 200 trajectories
makes repeated autoregressive training feasible under the matched proposal
budget. The two 48-example panels are equal-size, disjoint computational
allocations: one fixes the spectral mask without observing keep-validation,
and the other supports online retention. We do not attribute these exact
counts to the FNO literature or claim that they are an estimator-optimal
split. Channels are standardized for training and converted back to physical
scale before spectral metrics are evaluated. The scalar task metric is
ten-step keep-validation NRMSE.

\paragraph{Evolution model and optimization.}
The proxy has width 32, four Fourier layers, 32 retained modes, one residual
corrector, pointwise-MLP expansion 2, GELU activations, and coordinate input.
It is limited to 30M parameters and an 8-GiB device-memory envelope. Each
candidate is trained for 16 base epochs followed by 16 residual-refinement
epochs with batch size 16. We use AdamW with learning rate
\(5\times10^{-4}\), zero weight decay, cosine decay, gradient-norm clipping at
10, deterministic execution, and no mixed precision.

\paragraph{Editable boundary.}
The agent may edit the Fourier spectral transform and gates, the internal
residual-corrector architecture, and fusion between the spectral and spatial
branches. It may not alter the data manifest, rollout definition, objective,
trainer, evaluator, formal test, input/output interface, epoch budget, or
resource limits.

\paragraph{Formal evaluation.}
After source freezing, the first 1,000 trajectories are repartitioned into 800
training, 100 mask-construction, and 100 validation examples. This increases
training coverage while retaining a mask panel independent of formal
validation. The selected source is instantiated as the full model, with width
100, eight Fourier layers, 32 retained modes, and one residual corrector. It is
trained for 100 base epochs followed by 100 residual-refinement epochs with
batch size 40, AdamW at \(2\times10^{-4}\), zero weight decay, cosine decay,
gradient-norm clipping at 10, deterministic execution, and no mixed precision.
Paired training seeds are 1, 3, and 4.

The untouched 200-example official test set reports full-field NRMSE and
non-dominant-mode NMSE (ND-NMSE). Dominant-mode error is retained only as a
spectral guard or diagnostic when reported; it is not substituted for the
task metric. Parameter count is descriptive and is not used for model
selection. Results are reported as mean \(\pm\) sample standard deviation over
the three paired training seeds.

\subsubsection{SNGP}
\label{app:exp_sngp}

\paragraph{Evolution data and task.}
CIFAR-100 provides 50,000 official training and 10,000 official test
images~\citep{krizhevsky2009learning}. We split the training set within each
class into 400/20/30/50 images for candidate training, temperature calibration,
keep-validation, and audit, respectively. Per-class stratification preserves
the original class balance in every search-time panel. The 2,000-image
temperature panel is separate because fitting temperature on keep-validation
would adapt a nuisance parameter to the same examples used to retain edits;
this separation follows the role of held-out calibration in temperature
scaling~\citep{guo2017calibration}. The 5,000-image audit panel is unavailable
to proposal generation and retention and is reserved only for protocol-level
checks; it is neither used to select the online winner nor reported as a
formal test set. The exact 400/20/30/50 allocation is our search-throughput
design, not an SNGP benchmark split.

Training augmentation uses four-pixel padding, a random \(32\times32\) crop,
horizontal flipping, and fixed CIFAR-100 normalization. Calibration,
keep-validation, and audit use deterministic transforms. The scalar metric is
temperature-scaled keep-validation NLL. Temperature is fitted only on the
calibration panel, constrained to \([0.05,20]\), and optimized by L-BFGS for
at most 50 steps with a strong-Wolfe line search.

\paragraph{Evolution model and optimization.}
The proxy is WRN-28-2 with channels 16/32/64/128, four residual blocks per
group, 128-dimensional pre-GP features, and dropout 0.1. It is trained for 40
epochs with batch size 256 using SGD, base learning rate 0.08, momentum 0.9,
Nesterov acceleration, and weight decay \(3\times10^{-4}\). Epoch~0 uses
\(0.1\) times the base rate; learning-rate multipliers at 30\%, 60\%, and
80\% of training are 0.2, 0.04, and 0.008. After training, the GP precision
matrix is rebuilt using a deterministic loader before temperature fitting and
keep-validation.

\paragraph{Editable boundary.}
The agent may edit \texttt{SpectralConv2d}, random-Fourier-feature (RFF)
construction, normalization, scale and multi-bank structure, the
\texttt{RandomFeatureGP} head, precision/covariance and predictive-variance
calculation, and mean-field correction. The WRN backbone and adaptor remain
byte-identical. RFF width is restricted to 512--1024, the complete model must
remain below 10M parameters, and audit or test data cannot be loaded.

\paragraph{Formal evaluation.}
The frozen SNGP source is transplanted into WRN-28-10, with depth 28, widen
factor 10, 640-dimensional pre-GP features, and dropout 0.1. RFF-bank widths
are scaled by the fivefold feature-width ratio; proxy weights, the GP precision
matrix, and fitted temperature are not transferred. Each source is retrained
on all 50,000 CIFAR-100 training images for 250 epochs with batch size 256,
SGD at 0.08, momentum 0.9, Nesterov acceleration, and weight decay
\(3\times10^{-4}\). Paired training seeds are 101, 202, and 303. The GP
precision matrix is rebuilt deterministically after training, and no
post-hoc temperature is fitted on the clean test set.

We report clean NLL on CIFAR-100. For each of SVHN, and CIFAR-10,
the evaluator forms a deterministic, size-balanced comparison with CIFAR-100
test images using seed 20260814. CIFAR-100 examples form the negative class
and OOD examples form the positive class. The three OOD datasets are scored
separately and are not pooled.

The primary OOD score is the implemented Dempster--Shafer inverse-evidence
score
\begin{equation}
s_{\mathrm{DS}}(\mathbf u)
=
\frac{
C
}{
\displaystyle
\sum_{c=1}^{C}
\exp\!\left(
\operatorname{clip}
\bigl(\ell_c(\mathbf u),-30,30\bigr)
\right)
},
\end{equation}
rather than maximum softmax probability. Clean NLL is the formal task metric,
whereas OOD AUPR is a held-out conflict-related outcome and is not identical
to either online SNGP Probe. This protocol evaluates the distance-aware
uncertainty objective of SNGP~\citep{liu2020simple} while remaining isolated
from the in-distribution representation Probes used during evolution. Results
are reported as mean \(\pm\) sample standard deviation over the three paired
training seeds.

\subsubsection{GCNII}
\label{app:exp_gcnii}

\paragraph{Evolution data and task.}
The GCNII full-supervision protocol evaluates Cora, Citeseer, Pubmed,
Chameleon, Cornell, Texas, and Wisconsin using ten per-class 60\%/20\%/20\%
train/validation/test splits~\citep{chen2020simple,pei2020geom}. To keep
hundreds of online graph retrainings tractable, evolution uses only public
Chameleon split~0. Chameleon is a useful search substrate because it retains
the heterophilous hyperlink setting targeted by the conflict while providing
substantially more receivers and edges than the small Cornell, Texas, and
Wisconsin WebKB graphs. This choice is a computational search design; it does
not make Chameleon the formal test distribution or permit selection on the
other six datasets.

The 60\% partition trains each candidate, the 20\% validation partition
provides clean NLL for retention, and the 20\% test partition is unavailable
online. Clean accuracy is recorded as an auxiliary task statistic. At the
Stage-I branch point, the final label-free disagreement graphs are constructed
and frozen. They replace \(\rho=0.3\) of incoming non-self edges for validation
receivers, restrict sources to train+validation non-neighbors, degree-match a
32-node candidate pool, and select the source with maximum Jensen--Shannon
disagreement under the frozen parent. Five graphs use seeds 7301--7305.
Neither test nodes nor ground-truth labels participate in online graph
construction.

\begin{table}[h]
\centering
\caption{GCNII configurations used for formal evaluation.}
\label{tab:app_gcnii_formal_configs}
\small
\setlength{\tabcolsep}{3pt}
\renewcommand{\arraystretch}{1.05}
\begin{tabular}{@{}lrrrrrr@{}}
\toprule
Dataset & Layers & Hidden & Dropout & \(\alpha\) & \(\lambda\) & Weight decay \\
\midrule
Chameleon & 8  & 64 & 0.5 & 0.2 & 1.5 & \(5\times10^{-4}\) \\
Cornell   & 16 & 64 & 0.5 & 0.5 & 1.0 & \(10^{-3}\) \\
Texas     & 32 & 64 & 0.5 & 0.5 & 1.5 & \(10^{-4}\) \\
Wisconsin & 16 & 64 & 0.5 & 0.5 & 1.0 & \(5\times10^{-4}\) \\
Cora      & 64 & 64 & 0.5 & 0.2 & 0.5 & \(10^{-4}\) \\
Citeseer  & 64 & 64 & 0.5 & 0.5 & 0.5 & \(5\times10^{-6}\) \\
Pubmed    & 64 & 64 & 0.5 & 0.1 & 0.5 & \(5\times10^{-6}\) \\
\bottomrule
\end{tabular}
\end{table}

\paragraph{Evolution model and optimization.}
The Chameleon proxy has eight GCNII layers, hidden width 64, dropout 0.5,
\(\alpha=0.2\), \(\lambda=1.5\), non-variant identity mapping, and no
additional residual flag, matching the scale of the published Chameleon GCNII
configuration~\citep{chen2020simple}. Training uses Adam with learning rate
0.01, weight decay \(5\times10^{-4}\), at most 500 epochs,
validation-NLL patience 50, seed 42, and deterministic execution.

\paragraph{Editable boundary.}
Only \texttt{PropagationMixer} is editable. At layer \(\ell\), it may use the
current representation \(h^{(\ell)}\), the propagated representation
\(\widehat A h^{(\ell)}\), their difference, the initial representation
\(h^{(0)}\), and the layer index. The baseline mixer is
\begin{equation}
u^{(\ell)}
=
(1-\alpha)\widehat A h^{(\ell)}
+
\alpha h^{(0)},
\end{equation}
followed by the frozen identity-map coefficient
\(\theta_\ell=\log(\lambda/(\ell+1)+1)\). A proposal may add at most 50\% to
the reference parameter count, and its source must remain below 64~KiB. The
graph splits, trainer, layer count, evaluator, and formal perturbation code are
protected.

\paragraph{Formal evaluation.}
The same frozen source is instantiated and trained independently on all ten
public splits of each of the seven datasets; no dataset-specific source edit
or post-search selection is allowed. Every model is retrained from scratch
using Adam at 0.01 for at most 1,500 epochs, with clean-validation-NLL patience
100, seed 42, and deterministic execution. Checkpoint selection uses only
clean validation NLL. Dataset-specific depth, hidden width, dropout,
\(\alpha\), \(\lambda\), and weight decay follow the frozen configurations in
Table~\ref{tab:app_gcnii_formal_configs}.

Formal contamination replaces \(\rho=0.3\) of incoming non-self edges using
frozen seeds 7301--7305. Validation and test receivers are evaluated
separately, and sources from the full graph are allowed. Ground-truth
incompatibility may be used in this stress test because it is constructed only
after training and cannot affect checkpoint or source selection. We report
clean accuracy, clean NLL, contaminated NLL, and the contamination increase
\begin{equation}
\Delta\mathrm{NLL}_{\mathrm{contam}}
=
\mathrm{NLL}_{\mathrm{contam}}
-
\mathrm{NLL}_{\mathrm{clean}}.
\end{equation}
Contaminated NLL is first averaged over the five frozen perturbation graphs
within each split, after which the mean and sample standard deviation are
computed over the ten official splits. If Wisconsin is displayed separately
in a compact main-text table, it is treated as a representative case rather
than an additional model-selection target; the cross-dataset claim is
supported by the complete seven-dataset protocol.

\subsubsection{ESN}
\label{app:exp_esn}

\paragraph{Evolution data and task.}
NARMA-30 jointly requires nonlinear processing and information retention over
long temporal dependencies and is therefore a standard reservoir-computing
stress test~\citep{verstraeten2007experimental,ceni2024residual}. Evolution
uses the frozen recurrence
\begin{equation}
y_{t+1}
=
0.2y_t
+
0.04y_t\sum_{i=0}^{29}y_{t-i}
+
1.5u_{t-29}u_t
+
0.001,
\qquad
u_t\sim\mathcal U(0,0.5),
\end{equation}
with data seed 20260827, 200 washout samples, 5,000 readout-fit samples, and
the subsequent 2,500 samples for keep-validation. A single sequence removes
between-sequence noise from parent--candidate comparisons and keeps every
proposal inexpensive; it is an evolution proxy rather than evidence of
sequence-level generalization. That limitation is addressed using independent
formal sequences.

The scalar task metric is variance-normalized NRMSE. The fixed reservoir has
100 states, spectral radius 0.9, recurrent density 0.1, input scale 0.5, leak
rate 1, tanh activation, reservoir seeds 1103/2207/3301, and float64
arithmetic.

\paragraph{Readout fitting.}
For the bias-augmented design matrix \(\Phi=[\mathbf 1,H]\), the ridge readout
is
\begin{equation}
\widehat W
=
(\Phi^\top\Phi+\lambda R)^{-1}\Phi^\top Y,
\qquad
R=\operatorname{diag}(0,1,\ldots,1),
\qquad
\lambda=10^{-6}.
\end{equation}
The bias is unregularized, and a fresh readout is fitted for every candidate
and every formal sequence--reservoir pair.

\paragraph{Editable boundary.}
Only the parameter-free \texttt{ReservoirUpdateBlock}, which maps
\((h_{t-1},W_{\rm res}h_{t-1},W_{\rm in}u_t)\) to \(h_t\), may change. It may
not introduce trainable parameters or buffers, must remain within twice the
reference per-step operator count, and is limited to 49,152 source bytes.
Reservoir matrices, data generators, readout fitting, Probes, and evaluation
are protected. Static and runtime guards reject target leakage, future-input
access, non-determinism, non-finite states, or removal of dependence on either
fixed projection.

\paragraph{Formal evaluation.}
For in-domain evaluation, we cross 20 independent NARMA-30 sequence seeds
31001--31020 with ten reservoir seeds 61001--61010, producing 200 paired
configurations per source. Every configuration uses 200 washout, 5,000 fresh
readout-fit, and 5,000 held-out samples and refits the ridge readout. Probe
diagnostics use 64 anchors and all delays 1--30. This protocol tests whether a
source found on one search sequence remains effective under both new driving
sequences and new reservoir realizations.

Cross-task evaluation uses the Mackey--Glass delayed dynamical system
\begin{equation}
\frac{dx}{dt}
=
\frac{0.2x(t-17)}{1+x(t-17)^{10}}
-
0.1x(t),
\end{equation}
integrated with step 0.1, subsampled every ten integration steps, and preceded
by 1,000 sampled burn-in steps from an independently drawn
\(\mathcal U(0,1)\) history. Every replicate uses 200 washout, 5,000
readout-fit, and 5,000 test samples. Sequence seeds 41001--41020 are crossed
with the same ten reservoir seeds, and separate direct readouts are fitted for
prediction horizons \(h=1\) and \(h=84\).

We report NRMSE, MSE, and \(R^2\) as task metrics, together with
\(Z_{\mathrm{mem}}\) and \(Z_{\mathrm{nl}}\) as structural diagnostics.
Primary summaries are mean \(\pm\) sample standard deviation over all 200
sequence--reservoir pairs. When reported, supplementary 95\% intervals are
computed over the 20 sequence-level means using
\(t_{0.975,19}=2.093024\).

\subsubsection{TCM--Lite}
\label{app:exp_tcm}

\paragraph{Evolution data and task.}
The source pool combines the 800 DIV2K training images with 2,650 Flickr2K
images, giving 3,450 natural images~\citep{agustsson2017ntire,lim2017enhanced}.
Using split seed 20260910, source images---rather than merely extracted
patches---are assigned to six disjoint roles before cropping: candidate
training, mechanism development, evolution validation, hidden audit,
qualification discovery, and qualification confirmation. Consequently,
patches from a source image cannot occur in two roles. Patches are stored as
contiguous uint8 NHWC arrays, converted to RGB values in \([0,1]\), and
receive no online augmentation.

Candidate training uses 32,768 \(64\times64\) patches. Each other role contains
2,048 patches. Mechanism development is used to instantiate and sanity-check
the rate/detail pathway; qualification discovery fixes the qualification
statistics, and the disjoint confirmation panel checks them without reuse.
The hidden panel is reserved for audit and is unavailable to the agent and
retention controller. Evolution validation is the only panel used for online
keep decisions. Patch training and equal-size auxiliary panels bound
per-proposal runtime and keep the costs of auxiliary comparisons comparable.
These panel sizes define our fixed search proxy and are not the training split
of the original TCM paper. The scalar task metric is estimated
rate--distortion loss at \(\lambda=0.013\).

\paragraph{Evolution model and optimization.}
The model has 96 main-latent channels, 48 hyper-latent channels, four channel
slices, at most two previous slices per context, total downsampling factor 8,
attention window 4, and MLP ratio 2. Each candidate is trained for 1,500 steps
with batch size 128 using Adam at \(10^{-4}\), betas \((0.9,0.999)\), zero
weight decay, an auxiliary optimizer at \(10^{-3}\), training-sequence seed
314159, deterministic execution, and no mixed precision. The reduced patch
size and training schedule are used only to make repeated evolution
affordable; they are replaced by the full formal protocol below.

\paragraph{Editable boundary.}
The agent may modify the \texttt{RateDetailCodingBlock}, including
hyper/context fusion, slice prediction, latent refinement, and local detail
compensation. Arithmetic coding, entropy-likelihood accounting, the bitstream
interface, external tensor shapes, data, and evaluator are protected. A
candidate is limited to 8M parameters, 1.5 times the reference parameter count
and training-step runtime, and 98,304 source bytes. It may not bypass entropy
likelihoods or condition on validation/test identity.

\paragraph{Formal evaluation.}
After source freezing, proxy weights and patch banks are discarded. Each
source is retrained on all 3,450 DIV2K+Flickr2K images using
\(128\times128\) crops, batch size 64, and 30,000 Adam steps. The learning rate
is \(10^{-4}\) and drops to \(10^{-5}\) for the final 20\% of steps. The
primary operating point is \(\lambda=0.013\) with paired seeds 8101, 8209, and
8311. A secondary rate--distortion curve uses
\(\lambda\in\{0.0035,0.0067,0.013,0.025\}\) with seed 8101. Each seed uses a
separate fixed crop bank containing 32 patches per source image; runtime
sampling uses replacement and horizontal flipping.

Formal testing follows the full-image compression setting used by
TCM~\citep{liu2023learned}. Each Kodak image is compressed and decompressed
using the actual arithmetic coder. Images are padded to a multiple of 32 for
coding and cropped back to their original dimensions before scoring. The
actual rate is
\begin{equation}
\operatorname{bpp}_{\mathrm{actual}}
=
\frac{8N_{\mathrm{bytes}}}{HW}.
\end{equation}
We additionally record estimated bpp, RGB PSNR, RGB MS-SSIM, and the held-out
detail metric. The primary operating-point summary reports actual bpp,
MS-SSIM, and the detail metric as mean \(\pm\) sample standard deviation over
the three paired training seeds. No Kodak outcome is used for source
selection.

\subsection{Proposal validation and retention}
\label{app:selection_rules}

\paragraph{Common validity checks.}

Before metric-based comparison, a proposal must parse and import
successfully, satisfy the editable-file and API contracts, preserve the
required tensor shapes, dtypes, and devices, and produce finite outputs and
gradients. It must also satisfy the model-specific parameter-count,
source-size, runtime, and memory limits. Failure of any hard check causes
rejection.

Parent and candidate metrics are compared under the same evaluator manifest,
data split, training seeds, perturbations, and resource limits. A cached
parent result may be reused only when it is bound to the exact parent-source
hash and the same immutable evaluator manifest.

\paragraph{Task gain and scalar-only retention.}

Let \(p_t=x_t\) denote the incumbent parent and \(c_t=x_t^{e_t}\) the
candidate proposed at iteration \(t\). We use the direction-aligned task gain
\begin{equation}
G_t^S
=
S(p_t)-S(c_t),
\label{eq:app_task_gain}
\end{equation}
because all five search-time task metrics are defined so that lower values are
better. Thus, \(G_t^S>0\) denotes task improvement,
\(G_t^S=0\) denotes a task tie, and \(G_t^S<0\) denotes task degradation.

Stage~I and the AutoResearch Stage-II continuation use the scalar-only rule
\begin{equation}
\operatorname{Keep}_{\mathrm{AR}}(e_t)
=
\mathbb{I}\!\left[G_t^S>0\right].
\label{eq:app_scalar_keep}
\end{equation}
They therefore have no substantive-gain threshold: every valid proposal with
a strictly positive task gain is retained.

\paragraph{ConflictGuide retention.}

For model--conflict setting \(c\), let \(\tau_{S,c}\) separate clear task
improvements from the marginal task-gain region, let
\(\mathcal M_c\subseteq[0,\tau_{S,c}]\) denote the model-specific Probe-route
band, and let \(Q_c(e_t)\) denote its Probe and guard conditions. The common
ConflictGuide rule is
\begin{equation}
\operatorname{Keep}_{\mathrm{CG}}(e_t)
=
\mathbb{I}\!\left[
G_t^S>\tau_{S,c}
\;\lor\;
\left(
G_t^S\in\mathcal M_c
\;\land\;
Q_c(e_t)
\right)
\right].
\label{eq:app_conflict_keep}
\end{equation}
The first route retains a clear task improvement directly. The second route
retains a marginal or tied task result only when the model-specific conflict
condition is satisfied. No proposal with \(G_t^S<0\) is admissible under
either route.

Table~\ref{tab:app_keep_conditions} specifies the instantiated rules. In the
table, \(p\), \(c\), and \(a\) denote the current parent, proposed candidate,
and frozen Stage-I anchor, respectively. Every Probe gain is oriented so that
a positive value denotes improvement.

\begin{table*}[t]
\centering
\caption{Model-specific ConflictGuide retention conditions. The task column
shows the direct-retention condition followed by the Probe-route task band.
No retained proposal may have negative task gain.}
\label{tab:app_keep_conditions}
\footnotesize
\setlength{\tabcolsep}{3.2pt}
\renewcommand{\arraystretch}{1.15}
\begin{tabularx}{\textwidth}{@{}
l
p{0.3\textwidth}
>{\raggedright\arraybackslash}X
p{0.22\textwidth}
@{}}
\toprule
\textbf{Model}
&
\textbf{Task-gain condition}
&
\textbf{Required Probe improvement}
&
\textbf{Guard condition}
\\
\midrule

SpecB--FNO
&
Direct: \(G_t^S>5\times10^{-4}\).

Probe route:
\(0<G_t^S\le 5\times10^{-4}\).
&
\[
\frac{
Z_{\mathrm{ND}}(p)-Z_{\mathrm{ND}}(c)
}{
Z_{\mathrm{ND}}(p)+\epsilon
}
\ge 10^{-2}.
\]
&
\(
Z_{\mathrm D}(c)\le1.01Z_{\mathrm D}(a)
\)
and
\(
Z_{\mathrm{late}}(c)\le1.02Z_{\mathrm{late}}(a).
\)
\\
\addlinespace

SNGP
&
Direct: \(G_t^S>10^{-3}\).

Probe route:
\(0<G_t^S\le10^{-3}\).
&
\(
Z_{\mathrm{mar}}(c)-Z_{\mathrm{mar}}(a)\ge10^{-2}
\)
and
\(
Z_{\mathrm{dist}}(a)-Z_{\mathrm{dist}}(c)
\ge5\times10^{-3}.
\)
&
\(
\operatorname{Acc}(c)
\ge
\operatorname{Acc}(a)-10^{-2}.
\)
\\
\addlinespace

GCNII
&
Direct: \(G_t^S>\epsilon_S\).

Probe route:
\(0\le G_t^S\le\epsilon_S\),
where
\(\epsilon_S=1.01\times10^{-3}\).
&
\[
Z_{\mathrm{dis}}(p)-Z_{\mathrm{dis}}(c)
>
\tau_P.
\]
&
No additional model-specific guard; the non-negative task-gain requirement
prevents clean-NLL regression.
\\
\addlinespace


ESN
&
Direct: \(G_t^S>\epsilon_S\).

Probe route:
\(0\le G_t^S\le\epsilon_S\), where
\(\epsilon_S=4.18\times10^{-2}\).
&
Either
\[
G_{\mathrm{mem}}>\tau_{\mathrm{mem}},
\quad
G_{\mathrm{nl}}\ge-\tau_{\mathrm{nl}},
\]
or
\[
G_{\mathrm{nl}}>\tau_{\mathrm{nl}},
\quad
G_{\mathrm{mem}}\ge-\tau_{\mathrm{mem}}.
\]
&
\(\tau_{\mathrm{mem}}=8.85\times10^{-3}\) and
\(\tau_{\mathrm{nl}}=2.63\times10^{-4}\), frozen from the 95th percentiles
of paired no-edit variation.
\\
\addlinespace

TCM--Lite
&
Direct: \(G_t^S>0\).

Probe route:
\(G_t^S=0\).
&
Either
\[
G_{\mathrm{rate}}>\tau_R,
\;
G_{\mathrm{detail}}\ge-\tau_D,
\]
or
\[
G_{\mathrm{detail}}>\tau_D,
\;
G_{\mathrm{rate}}\ge-\tau_R.
\]
&
\(
\tau_R=5.89\times10^{-2}
\)
and
\(
\tau_D=4.44\times10^{-2}.
\)
\\

\bottomrule
\end{tabularx}
\end{table*}

\paragraph{Calibrated thresholds.}

All thresholds are fixed before the corresponding Stage-II evolution begins.
For the final label-free GCNII Probe, the calibrated values of \(\tau_P\) are
\(2.28\times10^{-2}\) and \(1.35\times10^{-2}\) for search seeds 1 and 2,
respectively. Table~\ref{tab:app_keep_conditions} reports numerical values to
three significant digits for readability; the executable evaluator manifests
retain the full-precision values used by the experiments.

\paragraph{Task non-regression invariant.}

Eq.~(\ref{eq:app_conflict_keep}) and all model-specific instantiations in
Table~\ref{tab:app_keep_conditions} exclude negative task gains. Therefore,
for every retained transition,
\begin{equation}
G_t^S\ge0,
\qquad\text{and hence}\qquad
S(x_{t+1})\le S(x_t).
\label{eq:app_task_nonregression}
\end{equation}
Numerical tolerances may be used only to identify equality at the evaluator's
recorded precision; they do not authorize a measurable task regression.
Evolution curves are generated from this retained-parent ledger, with the
incumbent carried forward after rejected or invalid proposals.

\subsection{Evolution budgets and protocol exceptions}
\label{app:budget_exceptions}

Table~\ref{tab:app_evolution_budgets} reports proposal opportunities rather
than only successful evaluations or retained edits. A budget of \(100+100\)
denotes 100 shared Stage-I proposals followed by 100 proposals in each matched
Stage-II continuation.

\begin{table*}[t]
\centering
\caption{Executed main-search budgets. ESN uses a shorter matched budget
because of its more restricted editable space.}
\label{tab:app_evolution_budgets}
\footnotesize
\setlength{\tabcolsep}{4pt}
\renewcommand{\arraystretch}{1.08}
\begin{tabular}{@{}lcccc@{}}
\toprule
Model & Search seeds & Stage I & Stage-II AutoResearch & Stage-II ConflictGuide \\
\midrule
SpecB--FNO & 0, 1, 2 & 100 & 100 & 100 \\
SNGP       & 0, 1, 2 & 100 & 100 & 100 \\
GCNII      & 0, 1, 2 & 100 & 100 & 100 \\
ESN        & 0, 1, 2 & 20  & 50  & 50  \\
TCM--Lite  & 0, 1, 2 & 100 & 100 & 100 \\
\bottomrule
\end{tabular}
\end{table*}

\paragraph{Standard setting.}
SpecB--FNO, SNGP, GCNII, and TCM--Lite use 100 task-only proposals to
construct the shared Stage-I incumbent, followed by 100 proposals in each
Stage-II continuation. The Stage-I trajectory is shared within a search round
and is therefore executed only once. Each Stage-II method consequently
receives the same 100 new proposal opportunities from the common branch point.
Search seeds alter the coding-agent trajectory but not the frozen data
manifest, evaluator, editable boundary, or per-candidate training budget.

\paragraph{ESN exception.}
ESN uses 20 Stage-I proposals followed by 50 proposals in each Stage-II
continuation. Its editable space is substantially narrower than those of the
other substrates: only the parameter-free \texttt{ReservoirUpdateBlock} may
change, while the reservoir matrices, readout fitting, data generator, and
evaluation procedure remain fixed. Candidate updates are additionally
constrained by a per-step operation budget and cannot introduce trainable
parameters or buffers. Under these restrictions, the executed trajectories
showed less proposal diversity and reached stable incumbents earlier, so the
sources admitted to held-out evaluation were frozen at the \(20+50\)
endpoints.

Both Stage-II arms start from the same Stage-I source and receive the same 50
proposal opportunities, preserving the within-model matched comparison.
Unexecuted proposals are not imputed or included in proposal-level statistics.
If a plot extends an ESN trajectory beyond its executed endpoint for visual
alignment, the final incumbent is carried forward as a constant value and the
extension is not counted as an evaluated proposal. Formal data, reservoir-seed
crosses, readout fitting, and held-out evaluation remain identical across the
reference and both terminal sources.

\subsection{Agent configuration, prompts, and history}
\label{app:agent_configuration}

\paragraph{Proposal agent.}
All main-table searches for SpecB--FNO, SNGP, GCNII, ESN, and
TCM--Lite use Claude Opus~4.6 as the proposal agent.  Each proposal is
generated in a fresh, nonpersistent session.  Consequently, the agent retains
no hidden conversational state across iterations; all cross-iteration
information available to it is supplied explicitly through the current source,
the task description, and the serialized structured history.  The experiment
controller does not override the model with an additional sampling-temperature
setting.

\paragraph{Prompt contents and tool boundary.}
At every iteration, the prompt provides the task definition, current retained
source, editable boundary, resource constraints, structured records of previous
proposal outcomes, and required output format.  Stage~I and the AutoResearch
continuation expose only scalar task feedback.  The ConflictGuide continuation
additionally exposes the registered conflict, Probe values and desirable
directions, relevant mechanism and model components, and the corresponding
keep or rejection reason.

The proposal agent may read only the registered program description, its
serialized history view, and the current \texttt{train.py}; it may modify only
\texttt{train.py}.  It cannot invoke the protected training or formal-evaluation
pipelines, inspect held-out data, modify controller code, or alter the retention
rule.

\paragraph{Complete ledger and agent-visible history.}
We distinguish the complete experimental ledger from the history rendered to
the proposal agent.  The complete ledger is append-only and records every
attempted proposal, including retained, rejected, invalid, crashed, non-finite,
out-of-memory, and timed-out candidates.  Each record is bound to the proposal
source hash, parent source hash, evaluator manifest, metric outputs, and
retention decision.

The agent-visible history is a deterministic structured projection of this
ledger.  Depending on the model-specific context limit, this projection is
either the complete structured history or a bounded context pack.  A bounded
pack preserves all retained records and deterministically selects recent,
high-performing, and otherwise relevant older records.  Text fields such as
\texttt{change\_summary} may be normalized and length-bounded outside the
proposal agent; the original uncompressed record remains in the complete
ledger.

\paragraph{History matching across comparison arms.}
History realization is allowed to differ across model families because their
proposal records, evaluation costs, and context requirements differ.  It is
never allowed to differ between paired comparison arms within a model and
search seed.  Before Stage~II, the AutoResearch and ConflictGuide arms use the
same branch-point source and the same preregistered history scope: either both
inherit the same ordered Stage-I prefix, or both start a new history at the
branch point.  They also share the same serialization schema, record ordering,
context limit, deterministic record-selection procedure, summary compression,
and tie-breaking rules.

After branching, the two visible histories naturally diverge because the
methods propose and retain different edits.  This post-branch divergence is a
consequence of the assigned feedback and retention policy and is therefore
part of the method effect, rather than a pre-existing history confound.  The
paired comparison requires equality of the pre-branch history and of the
history-construction policy, not equality of the method-dependent records
generated after branching.

For each paired run, the branch-point source hash, ordered pre-branch history
hash, history-policy identifier, prompt-template hash, and evaluator-manifest
hash are stored for audit.  Our cross-model conclusions aggregate paired
within-model effects; they do not assume that proposal records from different
model families are exchangeable.

\paragraph{Recovery and retries.}
The controller archives the exact rendered agent input before each call,
validates the returned source in an isolated child process, terminates the
complete process group on timeout, and restores the incumbent after failure.
Provider failures occurring before usable candidate source is returned follow
the frozen retry policy and do not consume a proposal opportunity.  Once
usable candidate source has been produced, the iteration is recorded even if
subsequent validation or evaluation fails.

\begin{table*}[!th]
\centering
\caption{Relationship between the upstream AutoResearch implementation and
our controlled multi-model evolution harness.  The modifications isolate the
effect of competing-behavior feedback while supporting heterogeneous training
substrates.}
\label{tab:autoresearch_differences}
\footnotesize
\setlength{\tabcolsep}{4pt}
\renewcommand{\arraystretch}{1.12}
\begin{tabularx}{\textwidth}{@{}
p{0.15\textwidth}
p{0.25\textwidth}
p{0.29\textwidth}
X@{}}
\toprule
\textbf{Aspect}
& \textbf{Upstream AutoResearch}
& \textbf{Our controlled harness}
& \textbf{Reason for the modification} \\
\midrule

Loop ownership
& The coding agent edits the program, invokes training, interprets the output,
and performs Git-based keep or reset operations.
& The agent only proposes source edits.  A protected external controller owns
validation, training, evaluation, retention, rollback, and accounting.
& Prevents evaluator modification and information leakage, and applies the
same frozen decision rule to every candidate. \\

Session and history
& The agent follows an iterative experiment loop and records a compact
\texttt{results.tsv}.
& Every proposal uses a fresh session supplied with a deterministic,
source-hash-bound structured history view; a separate complete ledger archives
all attempts.
& Removes unrecorded conversational state and permits exact reconstruction of
the information available at each iteration. \\

Task substrate
& The public implementation optimizes one language-model training program
under a fixed wall-clock budget and a scalar validation metric.
& We instantiate the loop for five model families with frozen model-specific
training budgets and scalar task metrics $S(x)$.
& A single training duration and metric are not meaningful across PDE,
classification, graph, reservoir, and compression tasks. \\

Feedback
& Each experiment is summarized primarily by the scalar task metric.
& The scalar-only arm receives only $S(x)$, whereas ConflictGuide additionally
receives registered competing-behavior Probes and Probe-aware decision
information.
& This is the experimental intervention whose effect we study. \\

Search design
& The public loop follows a single sequential trajectory.
& Each search seed runs its own Stage-I scalar trajectory, after which matched
AutoResearch and ConflictGuide continuations begin from the same branch point.
& Creates a paired comparison that controls initialization and all
pre-branch search outcomes. \\

Editable boundary
& \texttt{train.py} is editable while supporting infrastructure is fixed.
& We retain the one-file editable boundary, but impose model-specific API,
shape, determinism, runtime, memory, and parameter constraints.
& Preserves evaluator semantics and prevents edits from changing the
experimental task or resource budget. \\

Selection and rollback
& The agent manages experiment commits and resets through Git.
& A protected controller applies the frozen retention predicate and identifies
sources through immutable hashes and append-only ledgers.
& Makes acceptance reproducible and prevents the proposal agent from changing
or bypassing selection. \\

Probe computation
& No competing-behavior Probe layer is present.
& Probe computations are external to the editable program.  When needed for
cost matching, they are also computed privately in the scalar-only arm but are
not exposed or used for retention.
& Separates visibility and selection effects from evaluator workload. \\

Held-out evaluation
& The public loop focuses on the online validation metric.
& Formal evaluation uses frozen terminal source hashes and held-out data that
are unavailable during evolution.
& Prevents winner selection on reported test metrics and separates search-time
feedback from final evaluation. \\

Failure accounting
& Crash recovery and continuation are largely handled within the agent-driven
loop.
& Validation failures, crashes, timeouts, and non-finite outputs follow frozen
accounting and rollback rules.
& Ensures that both comparison arms consume proposal opportunities under the
same failure policy. \\

\bottomrule
\end{tabularx}
\end{table*}

\subsection{Relationship to the upstream AutoResearch implementation}
\label{app:autoresearch_relationship}

Our scalar-only baseline preserves the central AutoResearch abstraction: a
coding agent repeatedly edits a designated training program, receives a scalar
task result, and retains task-improving source code.  However, our experimental
system is a controlled, multi-model implementation of this abstraction rather
than a literal execution of the upstream repository
\citep{karpathy2026autoresearch}.  Table~\ref{tab:autoresearch_differences}
summarizes the deliberate differences.

Accordingly, ``AutoResearch'' in our figures and tables denotes the
scalar-only arm of this controlled harness.  It shares the proposal agent,
editable source boundary, training and evaluation substrate, structured-history
policy, branch-point initialization, and proposal budget with ConflictGuide;
it differs in the preregistered feedback and retention policy described above.

\subsection{Compute and runtime controls}
\label{app:compute_runtime}

\begin{table}[t]
\centering
\caption{Administrative watchdogs used during online evolution.
These limits determine failure handling and do not replace the fixed
model-specific training budgets.}
\label{tab:app_runtime_caps}
\footnotesize
\setlength{\tabcolsep}{3.5pt}
\renewcommand{\arraystretch}{1.10}
\begin{tabular}{@{}lcc@{}}
\toprule
Model
& Proposal generation
& Candidate evaluation \\
\midrule
SpecB--FNO & 1,800 s & 1,500 s \\
SNGP       & 1,800 s & 1,024 s \\
GCNII      & 1,800 s & 168 s \\
ESN        & 1,800 s & 1,800 s \\
TCM--Lite  & 660 s   & 300 s \\
\bottomrule
\end{tabular}
\end{table}

All search and formal-evaluation jobs were executed on shared
CUDA-enabled multi-GPU servers with explicit device assignment.  Concurrent
jobs were launched only when sufficient GPU memory and host CPU capacity were
available.  Scheduling decisions affected only resource allocation and were
not used for proposal evaluation or winner selection.  The assigned device,
software environment, source hash, and protocol hash are retained in the
corresponding run manifests.

Online evolution uses separate watchdogs for proposal generation and candidate
evaluation.  These watchdogs are administrative failure limits rather than
optimization budgets: the number of training epochs, steps, data examples, and
proposal opportunities is fixed independently by the model-specific protocol.
A candidate that completes within its watchdog is evaluated solely by the
registered task and Probe metrics; elapsed runtime is not part of the
retention objective.

Table~\ref{tab:app_runtime_caps} reports all watchdog values used in the
main experiments.  Multiple evaluator values indicate preregistered or
recorded run-specific amendments and are detailed below the table.

Where a watchdog was amended, the amendment changed only the maximum time
allowed for the corresponding process to finish; it did not change the
training data, epoch or step budget, proposal budget, metric definitions, or
retention rule.  Timeout events are recorded as failed proposal attempts under
the common failure-accounting policy.

External model calls are used only to propose source edits.  Candidate
validation, training, metric computation, retention, formal retraining, and
result aggregation are executed by frozen local code.  Search cost is measured
in proposal opportunities, whereas formal-evaluation cost is measured using
the paired training seeds, official data splits, or crossed
sequence--reservoir configurations specified in the corresponding evaluation
protocols.

\paragraph{Incremental cost of Probe feedback.}

The watchdogs in Table~\ref{tab:app_runtime_caps} are failure ceilings and
should not be interpreted as typical runtimes. We therefore separately
account for the evaluator-side computation associated with Probe feedback.
Proposal-generation latency was not isolated consistently across all five
model families because the available operational records can include provider
latency, queueing, lock waiting, and retries. We consequently make no
quantitative claim about the incremental LLM latency caused by the additional
Probe fields in the prompt.

Table~\ref{tab:app_probe_runtime_overhead} summarizes the evaluator-side
runtime records. For SpecB--FNO, ESN, and TCM--Lite, the scalar-only and
ConflictGuide arms executed the same metric workload: Probes were computed in
both arms but were hidden from the scalar-only agent and were not used for its
retention decisions. Their realized evaluator-time difference is therefore
zero. The separately reported Probe-pass time indicates the computational
cost that would be removed by an evaluator that omitted these hidden
measurements.

\begin{table*}[t]
\centering
\caption{Evaluator-side cost of Probe feedback. Times are wall-clock seconds.
A zero realized difference means that the scalar-only arm computed the same
hidden Probe quantities under the matched evaluator workload. Probe-pass
times isolate the computation attributable to those measurements.}
\label{tab:app_probe_runtime_overhead}
\footnotesize
\setlength{\tabcolsep}{4pt}
\renewcommand{\arraystretch}{1.12}
\begin{tabularx}{\textwidth}{@{}
l
p{0.22\textwidth}
p{0.22\textwidth}
X
@{}}
\toprule
\textbf{Model}
&
\textbf{Scalar-only evaluator}
&
\textbf{ConflictGuide evaluator}
&
\textbf{Incremental-cost accounting}
\\
\midrule

SpecB--FNO
&
Task metric and hidden spectral Probes
&
Identical computation, with Probes exposed to the agent
&
The realized difference is \(0.0\)~s. Across eight matched calibration
evaluations, the spectral Probe pass required a median of \(3.60\)~s,
corresponding to \(0.88\%\) of the complete candidate-evaluation time.
\\
\addlinespace

SNGP
&
Training, GP-state reconstruction, temperature fitting, and task metrics
&
The same computation plus representation collection and geometry Probes
&
Across three matched no-edit branch anchors, the median paired increase was
\(7.68\)~s, or \(2.57\%\) of the scalar-only evaluation time. This is a
descriptive wall-clock estimate because the paired jobs were not executed
simultaneously.
\\
\addlinespace

GCNII
&
Median candidate-evaluation time of \(10.4\)~s
&
Probe-normalized median candidate-evaluation time of \(11.7\)~s
&
The estimated incremental cost is \(1.3\)~s per candidate, corresponding to
a \(12.8\%\) increase over the scalar-only evaluator.
\\
\addlinespace

ESN
&
Median total candidate-evaluation time of \(82.4\)~s, including hidden
Jacobian Probes
&
The same median total time of \(82.4\)~s, with Probe measurements exposed
to the agent
&
The three reservoir-seed Jacobian
Probe passes together required approximately \(77.9\)~s, corresponding to
\(94.5\%\) of the complete candidate-evaluation time.
\\
\addlinespace

TCM--Lite
&
Median total candidate-evaluation time of \(258.5\)~s, including hidden
rate and detail measurements
&
The same median total time of \(258.5\)~s, with the measurements exposed
to the agent
&
The separately timed multiscale-detail
pass required approximately \(0.04\)~s, or less than \(0.1\%\) of the total.
The rate measurement adds no separate pass because it is already required
by the rate--distortion task metric.
\\

\bottomrule
\end{tabularx}
\end{table*}

The primary search budget is therefore defined by proposal opportunities
rather than aggregate wall-clock time. SpecB--FNO, ESN, and TCM--Lite use
matched evaluator workloads across the two Stage-II arms. SNGP incurs a small
additional representation-statistics pass, while GCNII incurs a modest
additional cost for evaluating the frozen disagreement interventions. In all
cases, elapsed runtime is recorded for accounting purposes and is not part of
the retention objective.

\section{Feedback Baselines}
\label{app:feedback_controls}

This appendix describes the two feedback baselines compared with
ConflictGuide in Table~8: \emph{multi-metric feedback} and
\emph{trade-off-only prompting}. The comparison is conducted on SNGP and
GCNII using search round~2. Its purpose is to test whether conventional
additional metrics or a qualitative trade-off reminder can substitute for
model-specific, quantitatively measured conflict feedback.

Unless stated below, each condition follows the corresponding main-experiment
configuration in Appendix~\ref{app:experimental_details}, including the
dataset partition, model and training configuration, editable scope, proposal
agent, structured-history format, continuation budget, and formal-evaluation
protocol.

\subsection{Compared feedback conditions}
\label{app:feedback_conditions}

All three conditions expose the same model-specific scalar task feedback:
temperature-scaled keep-validation NLL and accuracy for SNGP, and clean
keep-validation NLL and accuracy for GCNII. They differ only in the
additional information available to the proposal agent and in whether that
information supports a conflict-specific retention route.

\begin{table*}[t]
\centering
\caption{Additional information supplied by the three feedback strategies.
Trade-off-only prompting supplies a qualitative reminder without
additional quantitative measurements.}
\label{tab:feedback_condition_summary}
\footnotesize
\setlength{\tabcolsep}{4pt}
\renewcommand{\arraystretch}{1.14}
\begin{tabularx}{\textwidth}{@{}
p{0.10\textwidth}
p{0.30\textwidth}
p{0.22\textwidth}
>{\raggedright\arraybackslash}X
@{}}
\toprule
\textbf{Model}
&
\textbf{Multi-metric feedback}
&
\textbf{Trade-off-only prompting}
&
\textbf{ConflictGuide}
\\
\midrule

SNGP
&
Fifteen-bin calibration error and development OOD
Dempster--Shafer AUPR.
&
A generic qualitative reminder, without additional metrics or
model-specific conflict information.
&
Low-quantile centroid margin \(Z_{\mathrm{mar}}\), input--feature distance
distortion \(Z_{\mathrm{dist}}\), their desirable directions, and
Probe-aware retention feedback.
\\
\addlinespace

GCNII
&
Mean NLL and accuracy under five frozen, randomly rewired
degree-matched graphs.
&
A generic qualitative reminder, without additional metrics or
model-specific conflict information.
&
Disagreement-contamination sensitivity \(Z_{\mathrm{dis}}\), its desirable
direction, and Probe-aware retention feedback.
\\

\bottomrule
\end{tabularx}
\end{table*}

Neither baseline receives the ConflictGuide taxonomy path, the
model-specific conflict instantiation, Probe definitions or values, Probe
thresholds, or Probe-aware rejection explanations.

\subsection{Multi-metric feedback}
\label{app:multi_metric_feedback}

The multi-metric baseline augments ordinary task feedback with conventional
outcome or robustness measurements. These quantities are shown to the
proposal agent and included in its structured history.
Unlike ConflictGuide's Probes, they are not constructed to measure the
instantiated competing behaviors separately.

\subsubsection{SNGP}

Let \(\widehat T_x\) be the temperature fitted for candidate \(x\) on the
frozen temperature-calibration split. Its calibrated keep-validation
probabilities are
\begin{equation}
\label{eq:feedback_sngp_probability}
\mathbf{p}_i^{(x)}
=
\mathrm{softmax}
\left(
\frac{\boldsymbol{\ell}_i^{(x)}}{\widehat T_x}
\right).
\end{equation}

In addition to calibrated keep-validation NLL and accuracy, the agent observes
the equal-width fifteen-bin expected calibration error
\begin{equation}
\label{eq:feedback_sngp_ece}
M_{\mathrm{ECE}}(x)
=
\sum_{b=1}^{15}
\frac{|\mathcal{I}_b|}{|\mathcal{V}_{\mathrm{keep}}|}
\left|
\mathrm{acc}(\mathcal{I}_b)
-
\mathrm{conf}(\mathcal{I}_b)
\right|,
\qquad
M_{\mathrm{ECE}}(x)\downarrow.
\end{equation}
Here, \(\mathcal{I}_b\) contains the keep-validation examples whose maximum
calibrated probability lies in bin \(b\);
\(\mathrm{acc}(\mathcal{I}_b)\) is their empirical accuracy and
\(\mathrm{conf}(\mathcal{I}_b)\) is their mean maximum probability.

The second additional metric is development OOD average precision. For
\(C=100\) classes, define the inverse-evidence score
\begin{equation}
\label{eq:feedback_sngp_dev_score}
s_{\mathrm{dev}}(\mathbf{u};x)
=
\frac{C}{
\displaystyle
\sum_{c=1}^{C}
\exp
\left(
\mathrm{clip}
\left(
\frac{\ell_c^{(x)}(\mathbf{u})}{\widehat T_x},
-30,
30
\right)
\right)
}.
\end{equation}
This score is a strictly increasing transformation of the corresponding
bounded Dempster--Shafer score and therefore gives the same AUPR ranking.

Let \(\mathcal{D}_{\mathrm{dev}}\) contain \(3{,}000\) frozen CIFAR-100
keep-validation examples and \(3{,}000\) frozen CIFAR-10 training examples.
Define
\begin{equation}
y_{\mathrm{OOD}}(\mathbf{u})
=
\begin{cases}
1, & \mathbf{u}\text{ is from CIFAR-10},\\
0, & \mathbf{u}\text{ is from CIFAR-100}.
\end{cases}
\end{equation}
The development OOD metric is
\begin{equation}
\label{eq:feedback_sngp_dev_aupr}
M_{\mathrm{dev\text{-}OOD}}(x)
=
\mathrm{AP}
\left(
\left\{
s_{\mathrm{dev}}(\mathbf{u};x)
\right\}_{\mathbf{u}\in\mathcal{D}_{\mathrm{dev}}},
\left\{
y_{\mathrm{OOD}}(\mathbf{u})
\right\}_{\mathbf{u}\in\mathcal{D}_{\mathrm{dev}}}
\right),
\qquad
M_{\mathrm{dev\text{-}OOD}}(x)\uparrow.
\end{equation}

These metrics measure aggregate calibration and downstream OOD ranking.
They do not separately measure vulnerable-tail class separation and
preservation of input-relative representation geometry, and hence differ
from ConflictGuide's \(Z_{\mathrm{mar}}\) and \(Z_{\mathrm{dist}}\).

\subsubsection{GCNII}

The GCNII multi-metric condition uses \(K=5\) frozen random graph
interventions,
\begin{equation}
A_{\mathrm{rnd}}^{(1)},\ldots,A_{\mathrm{rnd}}^{(K)}.
\end{equation}
In each realization, \(30\%\) of eligible incoming non-self messages to
keep-validation receivers are replaced. Each replacement source is sampled
uniformly from a degree-matched pool of train-or-validation non-neighbors.
Candidates are trained on the clean graph; the perturbed graphs are used only
for post-training forward evaluation.

The random-perturbation NLL is
\begin{equation}
\label{eq:feedback_gcnii_random_nll}
M_{\mathrm{rnd\text{-}NLL}}(x)
=
\frac{1}{K}
\sum_{k=1}^{K}
\mathcal{L}_{\mathrm{NLL}}
\left(
x;
A_{\mathrm{rnd}}^{(k)},
X,
\mathcal{V}_{\mathrm{keep}}
\right),
\qquad
M_{\mathrm{rnd\text{-}NLL}}(x)\downarrow.
\end{equation}

The corresponding random-perturbation accuracy is
\begin{equation}
\label{eq:feedback_gcnii_random_acc}
M_{\mathrm{rnd\text{-}Acc}}(x)
=
\frac{1}{K}
\sum_{k=1}^{K}
\mathrm{Acc}
\left(
x;
A_{\mathrm{rnd}}^{(k)},
X,
\mathcal{V}_{\mathrm{keep}}
\right),
\qquad
M_{\mathrm{rnd\text{-}Acc}}(x)\uparrow.
\end{equation}

These interventions measure generic robustness to random message replacement.
Their sources are not selected using labels, anchor-prediction disagreement,
or Jensen--Shannon divergence. They therefore differ from ConflictGuide's
targeted high-disagreement intervention and its contaminated-minus-clean
Probe \(Z_{\mathrm{dis}}\).

\subsection{Trade-off-only prompting}
\label{app:tradeoff_only_prompting}

The trade-off-only baseline tests whether a qualitative warning is sufficient
without quantitative competing-behavior measurements. Its condition-specific
addition to the proposal prompt is:

\begin{quote}
\ttfamily
Consider trade-offs when modifying model components; avoid improving one
behavior at the expense of another.
\end{quote}

This reminder does not name the two model-specific behaviors, identify the
components through which they interact, define a Probe, provide numerical
behavior measurements, or specify how the trade-off should be assessed.
Relative to ordinary task-only feedback, the quoted sentence is the only
additional information visible to the proposal agent.

\subsection{Matched comparison protocol}
\label{app:feedback_comparison_protocol}

For each model, all feedback conditions start from the terminal incumbent
obtained after the complete \(100\)-proposal Stage-I search in round~2. Thus,
the branch point is the final Stage-I winner, even when its last accepted
update occurred before the final Stage-I iteration.

Within each model, the conditions use the same branch source, no-edit anchor,
proposal agent, data partition, training recipe, editable-file boundary,
structured-history policy, resource limits, and continuation-budget
accounting. Prompt fields unrelated to the feedback condition are held fixed.
Rejected and invalid proposals leave the retained parent unchanged.

All conditions use task performance as the primary retention criterion.
The feedback baselines follow their corresponding task-based rules,
while ConflictGuide incorporates qualified Probe measurements into
auxiliary retention checks, as specified in
Appendix~\ref{app:selection_rules}.
Table~\ref{tab:feedback_baselines} evaluates these feedback configurations together with their
associated retention rules.

\subsection{Formal evaluation and interpretation}
\label{app:feedback_formal_evaluation}

The terminal source from each condition is evaluated using exactly the same
full-scale retraining, paired seeds, held-out datasets, checkpoint-selection
rules, and reported metrics as the corresponding SNGP or GCNII main
experiment. These shared procedures are specified in
Appendix~\ref{app:experimental_details} and are not repeated here.

Two distinctions are relevant when interpreting the results. First, the SNGP
multi-metric agent observes a frozen subset of CIFAR-10 \emph{training}
examples during evolution, whereas Table~8 evaluates OOD AUPR on CIFAR-10
\emph{test} examples. The samples are disjoint, but the OOD domain is the
same; the result is therefore held-out evaluation within an observed
development domain rather than unseen-domain generalization. The
trade-off-only and ConflictGuide agents do not observe CIFAR-10 measurements
during evolution.

Second, the GCNII multi-metric condition exposes generic random-perturbation
metrics, whereas formal evaluation uses the registered incompatible-message
stress test. The latter is applied only after source selection and is not
available to any proposal agent.

Multi-metric feedback therefore tests whether additional conventional
measurements are sufficient without explicitly measuring the instantiated
conflict. Trade-off-only prompting tests whether a qualitative description is
sufficient without quantitative evidence. ConflictGuide differs by making the
competing behaviors observable through model-specific Probes and using those
measurements in both proposal feedback and retention.

\section{Extension to Other Agents and Frameworks}
\label{app:transfer}

\begin{table}[t]
\centering
\caption{Agent and search configurations in the SpecB--FNO transfer
experiments.  Numerical candidate metrics are computed by the fixed evaluator
in all settings.  The post-execution review model in AIDE only identifies
implementation or runtime failures.}
\label{tab:transfer-agent-configurations}
\small
\setlength{\tabcolsep}{3pt}
\renewcommand{\arraystretch}{1.12}
\begin{tabularx}{\linewidth}{@{}
p{0.19\linewidth}
p{0.25\linewidth}
p{0.23\linewidth}
>{\raggedright\arraybackslash}X
@{}}
\toprule
\textbf{Setting}
&
\textbf{Code generation}
&
\textbf{Post-execution bug review}
&
\textbf{Search structure}
\\
\midrule

Claude Code
&
Claude Opus~4.6
&
Not used
&
Sequential retained-parent search
\\

Kimi Code
&
Kimi K2.7 Code
&
Not used
&
Sequential retained-parent search
\\

AIDE (OpenAI)
&
o4-mini
&
GPT-4.1-mini
&
Solution tree with improve and debug operations
\\

AIDE (GLM)
&
GLM-5.3 (low reasoning effort)
&
GLM-5.3-Flash
&
Solution tree with improve and debug operations
\\

\bottomrule
\end{tabularx}
\end{table}

The main experiments use Claude Code with Claude Opus~4.6 as the proposal
agent. We examine transfer on SpecB--FNO in two settings. First, we replace
Claude Code with Kimi K2.7 Code while retaining the sequential
proposal--evaluation--retention controller. Second, we integrate
competing-behavior feedback into AIDE's solution-tree search under two
code-generation and execution-review model configurations. Each experiment
compares scalar-only feedback with ConflictGuide within the same agent or
framework and under matched continuation budgets. The Kimi experiment uses
search seed~2; both AIDE configurations use search seed~0. Results across
settings are not intended as a controlled ranking of the underlying LLMs.

\subsection{Common SpecB--FNO setup}
\label{app:transfer_common_substrate}

All transfer experiments use the SpecB--FNO conflict described in
Appendix~\ref{app:probe_specb_fno}: dominant-mode predictive fidelity
versus effective utilization of non-dominant spectral modes. The scalar
search metric is validation NRMSE,
\[
S(x)=\operatorname{NRMSE}_{\mathrm{val}}(x),
\qquad S(x)\downarrow.
\]
The primary conflict probe is non-dominant relative spectral error
\(Z_{\mathrm{ND}}\downarrow\); dominant-mode error
\(Z_{\mathrm D}\downarrow\) and late-rollout error
\(Z_{\mathrm{late}}\downarrow\) serve as guards.

The experiments fix the Navier--Stokes data and split, optimizer, proxy
training budget, seed policy, candidate interface, and evaluator. Candidates
may modify only the registered spectral component, residual corrector, and
stage-fusion implementation. Data loading, training, validation, selection
code, and the formal evaluator are outside the editable surface. The
held-out test set is unavailable during search. Proxy evaluations use width
\(32\), four layers, and the fixed training configuration. Formal evaluation
transfers the selected architecture to the full-scale model, discards its
proxy-trained weights, and retrains it from scratch.

\subsection{Proposal-agent transfer: Kimi K2.7 Code}
\label{app:transfer_kimi}

The Kimi experiment changes the proposal backend while preserving the
sequential controller. At each iteration, Kimi receives the task and
editable-surface description, current retained source, structured history,
and feedback for its assigned condition, and returns a complete candidate
module. The unchanged controller validates the candidate, trains it,
computes metrics, applies the retention rule, and restores the retained
parent when necessary. Kimi does not receive additional evaluator files or
held-out data.

Kimi first performs \(100\) scalar-guided Stage-I proposals from the
registered initial source. The resulting source and search history form a
common anchor for two \(100\)-proposal continuations. The scalar-only
continuation receives task feedback and uses task-based retention; the
ConflictGuide continuation additionally receives the qualified
SpecB--FNO probe results and uses probe-informed retention. The two
continuations share the anchor, backend, task instructions outside the
feedback section, editable surface, data, training configuration,
evaluator, and proposal budget.

Each history record reports the proposal, evaluation status, scalar task
measurement, retention outcome, and source identity. In the scalar-only
continuation, probe results are withheld from the proposal agent. In the
ConflictGuide continuation, valid records additionally report
\(Z_{\mathrm{ND}}\), \(Z_{\mathrm D}\), and \(Z_{\mathrm{late}}\), together with
the identified conflict and the desirable direction of each measurement.
The instructions prioritize clear task improvements and use the probes to
assess edits in the calibrated marginal-gain region. Task and probe
measurements are not combined into a weighted scalar objective.

\subsection{Search-framework transfer: AIDE}
\label{app:transfer_aide}

AIDE maintains a journal and solution tree, expanding nodes through
improvement or debugging operations. We retain its node and journal
representations, improvement and debugging operations, scalar-best parent
selection in the scalar-only condition, debugging probability of \(0.5\),
maximum debugging depth of \(3\), and node-eligibility rules. We replace
AIDE's unrestricted initial-draft stage with the registered SpecB--FNO
source as a fixed root. The root is evaluated once before search and does
not consume an expansion. This gives both feedback conditions the same
initial model and evaluator.

An adapter connects AIDE's candidates to the fixed SpecB--FNO executor.
The code-generation model supplies a short design sketch and a complete
candidate module, with instructions to make one change to a model mechanism
per node within the editable surface. The executor validates and trains
each candidate and computes its metrics; neither LLM supplies metric
values. A separate execution-review model assesses implementation and
runtime failures without access to probe results in either condition.
Failed candidates remain in the tree and consume an expansion, so eligible
nodes can subsequently be selected for debugging.

We evaluate two model configurations. The first uses o4-mini for code
generation and GPT-4.1-mini for execution review. The second uses GLM-5.3
with low reasoning effort for code generation and GLM-5.3-Flash for
execution review. Within each configuration, the scalar-only and
ConflictGuide conditions use the same models and search settings.

\subsubsection{Matched continuations}

Each configuration begins with \(100\) shared scalar-guided AIDE
expansions from the fixed root. We then checkpoint the complete solution
tree, node metadata, candidate sources, and random-number-generator state.
Identical copies of this checkpoint initialize a scalar-only refinement
and a ConflictGuide refinement, each with \(100\) further expansions.
The search and executor seeds are \(0\), and the same candidate-evaluation
time limit applies to both refinements. Their interventions differ in the
feedback available to improvement proposals and the probe-informed
parent-selection rule.

\subsubsection{Probe feedback and parent selection}

The scalar-only refinement stores probe measurements for analysis but
excludes them from code-generation and execution-review prompts. It uses
AIDE's scalar-best rule for improvement-parent selection. In the
ConflictGuide refinement, an improvement prompt additionally describes
the dominant versus non-dominant spectral conflict and reports the selected
parent's validation NRMSE, non-dominant and dominant spectral errors, and
late-rollout error. It states the desirable direction of each measurement
and prioritizes clear task improvements; it does not prescribe an
architecture or a weighted objective. Execution-review prompts remain
probe-free.

For probe-informed parent selection, let \(\mathcal N_t\) be the valid
nodes available after expansion \(t\), and define
\[
S_t^\star=\min_{v\in\mathcal N_t}S(v).
\]
The scalar frontier is
\begin{equation}
\label{eq:aide_scalar_frontier}
\mathcal F_t=
\left\{
v\in\mathcal N_t:
S(v)-S_t^\star\leq\tau_S
\right\},
\end{equation}
where \(\tau_S\) is the fixed SpecB--FNO scalar-equivalence tolerance.
A node in \(\mathcal F_t\) is probe-qualified only if its edit is also
scalar-equivalent to its parent, provides the calibrated parent-relative
improvement in \(Z_{\mathrm{ND}}\), and satisfies the dominant-mode and
late-rollout guards relative to the refinement anchor. A qualified node
may be preferred within \(\mathcal F_t\); probe results cannot promote a
node outside the scalar frontier. Remaining ties are resolved by lower
validation NRMSE, stronger normalized probe relief, and then earlier node
step and identifier.

In the o4-mini/GPT-4.1-mini experiment, no refinement node met the full
probe-qualification rule. Thus, probe-informed parent selection did not
activate in that run; differences between its continuations arose from
the feedback available to code generation.

\subsection{Matched formal evaluation}
\label{app:transfer_formal_evaluation}

The selected source from each continuation is fixed before formal
evaluation. Proxy-trained weights are discarded, and each architecture
is retrained from scratch at width \(100\), with eight layers, \(32\)
retained Fourier modes, \(100\) base-training epochs, and \(100\)
residual-training epochs. The paired training seeds are \(1\), \(3\), and
\(4\). The held-out test set comprises examples \(1000\)--\(1199\) and
is accessed only after training; no proposal or architecture change
follows test evaluation. The formal task metric is test NRMSE, and the
conflict-related metric is non-dominant relative NMSE.

The o4-mini/GPT-4.1-mini selected architectures use
\(\operatorname{GroupNorm}(8,\cdot)\) at proxy width \(32\), but eight
groups are invalid at formal width \(100\). A construction check made
before formal training or test access identified this incompatibility.
Both continuations therefore use the same portability rule: select the
largest divisor of the channel width that does not exceed eight. The rule
retains eight groups at width \(32\) and uses five at width \(100\).
The original selected sources and the common transformation are retained
for inspection. The GLM-based AIDE architectures require no such change.

\newcommand{\AAEditBox}[3]{%
  \par\begingroup
  \setlength{\fboxsep}{6pt}%
  \noindent\fbox{%
    \begin{minipage}{\dimexpr\linewidth-2\fboxsep-2\fboxrule\relax}
    \textbf{#1}\par\smallskip
    #2
    \vspace{1pt}

    \small #3
    \end{minipage}}%
  \par\endgroup\vspace{0.6em}%
}

\section{Additional Results and Analyses}
\label{app:additional_analysis}

This section complements the aggregate results with proposal- and trajectory-level analyses. We first examine competing-behavior trajectories and the ambiguity of behavior outcomes under similar scalar gains. We then analyze how feedback shifts the distribution of code-edit mechanisms, before presenting representative edits and their connections to established design motifs.

\subsection{Competing-Behavior Trajectories and Scalar-Gain Ambiguity}
\label{app:behavior_trajectories}

\begin{figure*}[t]
\centering
\includegraphics[width=0.6\textwidth]{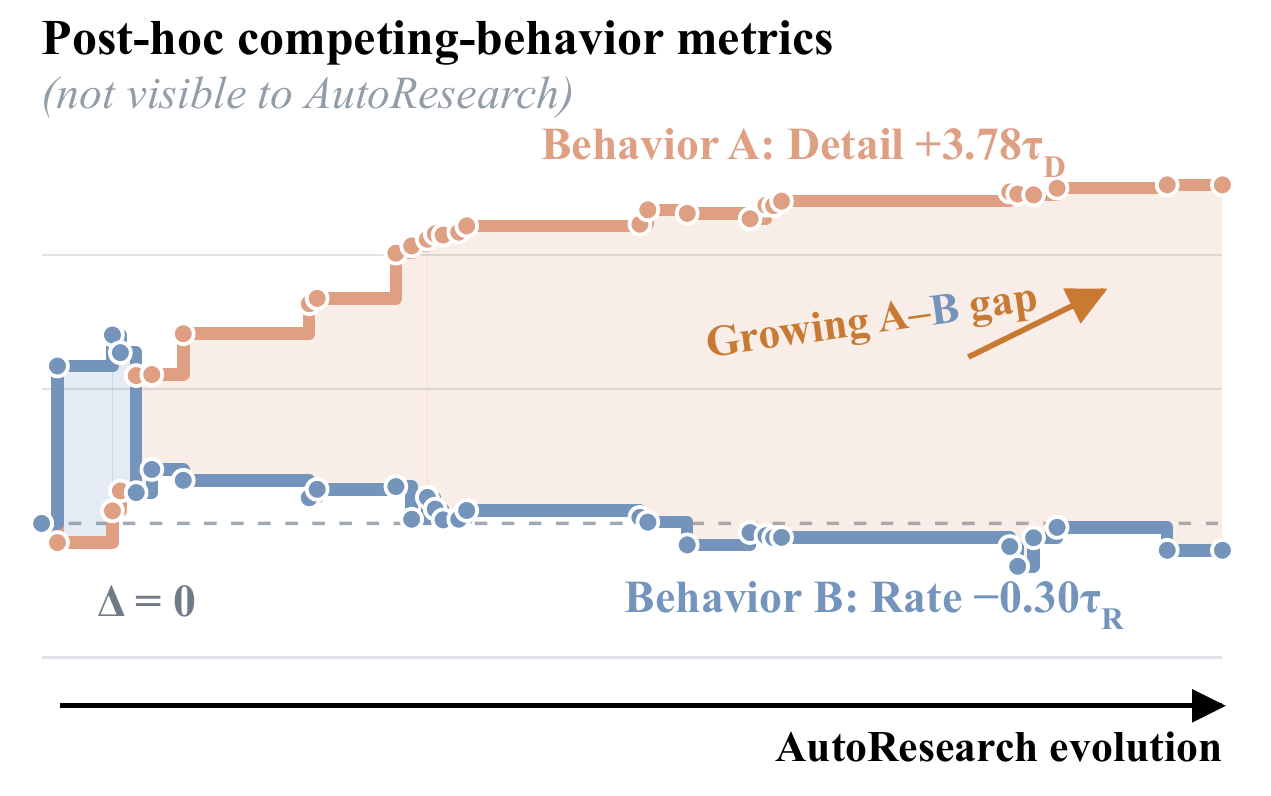}
\caption{Post-hoc competing-behavior trajectories for TCM--Lite during the task-only Stage~I of search round~1. The behavior metrics were hidden from the proposal agent. Changes are direction-aligned and normalized by the frozen Probe thresholds; positive values indicate improvement.}
\label{fig:app_tcm_behavior_trajectory}
\end{figure*}

Figure~\ref{fig:app_tcm_behavior_trajectory} makes the masking effect of an
aggregate task objective visible before any Probe feedback is introduced.
The retained Stage-I source improves detail by $3.78$ detail thresholds, while
the rate behavior ends $0.30$ rate thresholds below its initial value. Thus,
a monotonically improving rate--distortion objective can be driven primarily
by one component while the other moves in the opposite direction.

\begin{figure*}[ht]
\centering
\includegraphics[width=0.7\textwidth]{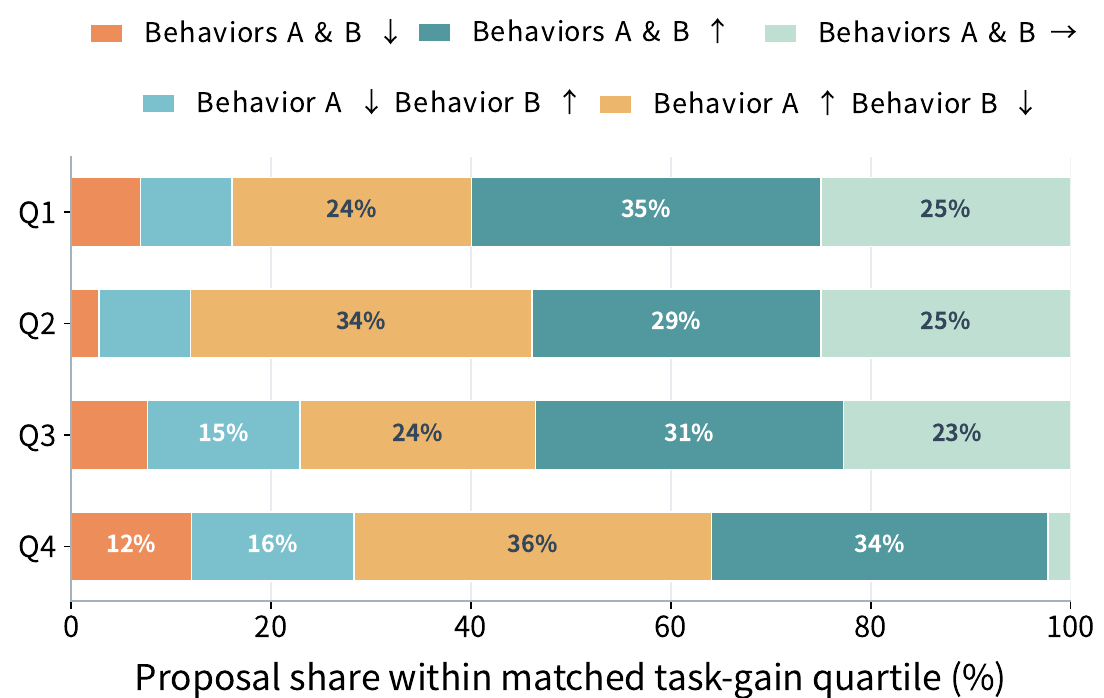}
\caption{Scalar ambiguity: similar task gains can correspond to different behavior outcomes. Positive-gain scalar-only proposals are grouped by within-model task-gain magnitude to assess whether scalar improvement identifies directional behavior outcomes. Multiple outcome categories appear in every group, showing that similar scalar gains remain compatible with materially different behavior changes.}
\label{fig:app_scalar_gain_ambiguity}
\end{figure*}

As shown in Figure~\ref{fig:app_scalar_gain_ambiguity}, conditioning on the relative size of the scalar improvement does not identify
the behavior outcome.  In every task-gain quartile, proposals occupy several
directional categories; the mean normalized outcome entropy is 0.892, where
one is the maximum, and no category exceeds 35.8\% in any quartile.  The same
conclusion holds with three or five within-model bins: the corresponding mean
entropies are 0.891 and 0.888.  The result supports a limited claim---similar
scalar gains are compatible with materially different behavior changes---not
the stronger claim that the scalar gain causally produces those changes.

\subsection{Code-Edit Distributions and Mechanism Examples}
\label{app:code_edit_distributions}

The tag audit shows that feedback changes where proposals concentrate. For
SNGP, kernel-scale edits account for 24.6\% of ConflictGuide proposals versus
12.7\% of AutoResearch proposals, whereas batch-normalization and
spectral-bound edits become less frequent. For TCM--Lite, pointwise-mixing
edits fall from 27.0\% to 7.4\%, while residual-refinement edits rise from
29.0\% to 34.1\%. The tags are non-exclusive: one proposal can instantiate
more than one mechanism. The boxes below make the most relevant labels
operational without reproducing long implementation diffs.

\AAEditBox{SNGP: learnable random-feature kernel scale}{%
\[
\begin{aligned}
s &\leftarrow \operatorname{softplus}(a)+\epsilon,\\
\Omega' &\leftarrow \Omega/s,\\
\Omega' &\leftarrow
\operatorname{clip}\!\left(\Omega',-\omega_{\max},\omega_{\max}\right),\\
\phi(z) &\leftarrow \sqrt{2/M}\,
\cos\!\left(z\Omega'+b\right),\\
\ell &\leftarrow W_{\mathrm{GP}}\phi(z).
\end{aligned}
\]
}{The edit learns the effective RFF length scale while retaining a frozen
frequency bound. It changes how strongly nearby pre-GP representations are
resolved by the GP head rather than adding an independent classifier.}

\AAEditBox{SNGP: stabilized ridge-precision update}{%
\[
\begin{aligned}
C &\leftarrow \sum_i \phi_i^{\top}W_i\phi_i,\\
\Lambda &\leftarrow \lambda I+C,\\
\Lambda &\leftarrow
Q\,\operatorname{diag}\!\bigl(\max\{\lambda_j,\epsilon\}\bigr)Q^{\top},\\
\Sigma &\leftarrow \Lambda^{-1}.
\end{aligned}
\]
}{Accumulation and eigendecomposition are carried out in higher precision.
Flooring small eigenvalues prevents poorly conditioned feature directions
from dominating posterior-variance and mean-field corrections.}

\AAEditBox{TCM--Lite: depthwise residual refinement}{%
\[
\begin{aligned}
q &\leftarrow \operatorname{DWConv}_{3\times3}(h),\\
r &\leftarrow
\operatorname{PWConv}_{1\times1}\!\left(\operatorname{GELU}(q)\right),\\
\eta &\leftarrow \sigma(a),\\
h' &\leftarrow h+\eta r.
\end{aligned}
\]
}{The depthwise filter supplies inexpensive local spatial correction, the
pointwise map mixes channels, and the learned gate controls how strongly the
new detail path changes the codec representation.}

\AAEditBox{TCM--Lite: pooled channel gate}{%
\[
\begin{aligned}
u_{\mathrm{avg}} &\leftarrow \operatorname{mean}_{HW}(h),\\
u_{\mathrm{max}} &\leftarrow \operatorname{max}_{HW}(h),\\
g &\leftarrow \sigma\!\left(
\operatorname{MLP}(u_{\mathrm{avg}}+u_{\mathrm{max}})\right),\\
h' &\leftarrow h\odot g_{[:,:,\mathrm{None},\mathrm{None}]}.
\end{aligned}
\]
}{Average pooling captures persistent channel activity and max pooling retains
localized responses. Their gate reallocates capacity without changing the
latent tensor shape.}

These distributions are descriptive, not estimates of a pure feedback effect: retained-edit frequencies reflect both proposal generation and retention, and the SNGP AutoResearch audit includes all 15 retained edits but only a nonrandom subset of rejected proposals.

\subsection{Representative Edits and Connections to Prior Work}
\label{app:representative_code_edits}

The examples below are arranged vertically rather than compressed into a
wide table. They are schematic summaries of recurring proposal mechanisms,
not verbatim winner implementations.

\paragraph{SpecB--FNO: retained-mode correction.}
\begin{equation*}
\begin{aligned}
X&\leftarrow\operatorname{rfft2}(x),
&Z_k&\leftarrow W_kX_k,\\
Z_k&\leftarrow g_kZ_k+s_kP(X)_k,
&y&\leftarrow\operatorname{irfft2}(Z),
\quad k\in\mathcal K.
\end{aligned}
\end{equation*}
The learned gain or gated bypass corrects selected retained modes without
replacing the full spectral operator. The connection to the Fourier-domain
parameterization of FNO~\citep{li2020fourier} is representational; the
per-mode correction and its placement are proposal-specific.

\paragraph{SNGP: conditioned GP state.}
\begin{equation*}
\begin{aligned}
\Phi&\leftarrow\operatorname{RFF}(z),
&C&\leftarrow\sum_i\Phi_i^{\top}W_i\Phi_i,\\
\Lambda&\leftarrow\lambda I+C,
&\widetilde\Lambda&\leftarrow
Q\operatorname{diag}(\max\{\lambda_j,\epsilon\})Q^{\top},\\
\Sigma&\leftarrow\widetilde\Lambda^{-1}.
\end{aligned}
\end{equation*}
The edit stabilizes the precision matrix before mean-field correction. It is
reported as a model-specific numerical mechanism, not as a new GP formulation.

\paragraph{GCNII: receiver-dependent message gate.}
\begin{equation*}
\begin{aligned}
m_i&\leftarrow\sum_j\widehat A_{ij}h_j,\\
\alpha_i&\leftarrow
\sigma\!\left(\operatorname{MLP}([h_i^{(0)},m_i])\right),\\
h_i'&\leftarrow(1-\alpha_i)h_i^{(0)}+\alpha_i m_i.
\end{aligned}
\end{equation*}
The gate varies self-retention and neighborhood mixing by receiver, allowing
the model to attenuate incompatible messages without removing graph
propagation globally.

\paragraph{ESN: state-dependent recurrent processing.}
\begin{equation*}
\begin{aligned}
g_t&\leftarrow\sigma(Uu_t+Vh_{t-1}),\\
r_t&\leftarrow W_{\mathrm{in}}u_t+g_t\odot Wh_{t-1},\\
\widetilde h_t&\leftarrow\tanh(r_t),\qquad
h_t\leftarrow(1-\alpha)h_{t-1}+\alpha\widetilde h_t.
\end{aligned}
\end{equation*}
The multiplicative gate makes local processing state--input dependent while
the explicit leaky path continues to carry long-range state information.

\paragraph{TCM--Lite: gated local correction.}
\begin{equation*}
\begin{aligned}
q&\leftarrow\operatorname{DWConv}_{3\times3}(h),\\
r&\leftarrow\operatorname{PWConv}_{1\times1}
\!\left(\operatorname{GELU}(q)\right),\\
h'&\leftarrow h+\sigma(a)r.
\end{aligned}
\end{equation*}
This edit combines depthwise--pointwise factorization with a gated residual
correction. Its closest general motifs are depthwise separable convolution
\citep{chollet2017xception} and residual learning~\citep{he2016deep}; neither
reference specifies this codec placement or gate.

Other audited proposals use mean/max pooled channel gates related at the
mechanism level to SE and CBAM~\citep{hu2018squeeze,woo2018cbam}, or spatially
shared channel masks related to SpatialDropout~\citep{tompson2015efficient}.
These are motif-level connections only: parameterization, initialization,
placement, and measured effects remain proposal specific.

\section{Limitations and Future Work}
\label{app:limitations}

\paragraph{Limitations.}
ConflictGuide applies when a conflict is supported by sufficient model-specific evidence and its competing behaviors can be measured with probes. ConflictGuide-Skill abstains when evidence is insufficient, and probes are qualified and fixed before evolution to avoid selecting metrics based on search outcomes. Controlled comparisons with taxonomy-free probe design and systematic evaluation of abstention remain future work. Our experiments examine distinct conflicts across five model families and show directional gains in the tested alternative code-agent settings; applicability to a wider range of settings remains to be established. We compare search outcomes under matched evolution budgets and separately report the additional time required to evaluate probes. The initial specification, implementation, and qualification of model-specific probes require setup effort that varies by model and is not included in a uniform end-to-end cost comparison. The transition to Stage II and the auxiliary retention rule follow prespecified criteria; adaptive timing remains to be studied.

\paragraph{Future Work.}
Future work could use search history to adjust the timing and strength of probe feedback and make probe construction and qualification more efficient. We also plan to study interacting conflicts, evaluate more model families and code agents, and systematically measure model-specific setup effort to better characterize applicability and total cost.

\end{document}